%% file: main.tex
\documentclass[]{style}
\usepackage[most]{tcolorbox}        
\usepackage{amsmath,amssymb}
\usepackage{lipsum}                 
\usepackage{xcolor}                 
\usepackage{fancyhdr}               
\usepackage{booktabs}
\usepackage[table]{xcolor}
\usepackage{url}
\usepackage{graphicx}   
\usepackage{subcaption} 
\usepackage{multirow}
\usepackage{tabularx}
\usepackage{bm}
\usepackage{float}
\usepackage{placeins}
\definecolor{lightpurple}{RGB}{153, 102, 204}  
\definecolor{lilac}{RGB}{182, 133, 210}        
\usepackage{hyperref}
\hypersetup{
    colorlinks=true,      
    linkcolor=purple,        
    citecolor=lilac,       
    urlcolor=magenta      
}
\usepackage{tabularx}
\newcolumntype{Y}{>{\raggedright\arraybackslash}X}

\definecolor{pwYellow}{RGB}{252,238,210}
\definecolor{pwPurple}{RGB}{239,220,247}
\definecolor{pwBlue}{RGB}{218,233,252}
\definecolor{pwGreen}{RGB}{218,241,234}

\definecolor{abstractpurple}{HTML}{9C27B0} 

\renewcommand{\maketitle}{\mymaketitle}

\usepackage{xcolor}
\usepackage{tcolorbox}
\tcbuselibrary{breakable}   
\newtcolorbox{notebox}[1][]{
    colback=lightpurple!20,
    colframe=lightpurple,
    fonttitle=\bfseries,
    title=#1,          
    boxrule=1pt,
    arc=4pt, 
    breakable,
    toptitle=4pt,      
    bottomtitle=4pt,   
    parbox=false,   
}

\usepackage{longtable}

\usepackage{booktabs}
\usepackage{tabularx}
\usepackage{array}
\usepackage{multirow}
\usepackage[table]{xcolor}
\usepackage{pifont}

\definecolor{AEIgreen}{HTML}{24965A}
\definecolor{AEIred}{HTML}{E05252}
\definecolor{AEIorange}{HTML}{E59A2F}

\newcommand{\cmark}{\textcolor{AEIgreen}{\ding{51}}}
\newcommand{\xmark}{\textcolor{AEIred}{\ding{55}}}
\newcommand{\pmark}{\textcolor{AEIorange}{\textbullet}}

\definecolor{AEIphysical}{HTML}{EEF3F8}
\definecolor{AEIworld}{HTML}{EFF5F1}
\definecolor{AEIscale}{HTML}{F8F3EB}
\definecolor{AEIvalid}{HTML}{F4F0F7}

\definecolor{AEIrow}{HTML}{F7F8FA}

\newcommand{\AEIgroupfont}{%
    \fontsize{6.8pt}{7.4pt}\selectfont\bfseries
}

\newcommand{\AEIsubfont}{%
    \fontsize{6.4pt}{7.0pt}\selectfont\bfseries
}

\newcommand{\AEIgroup}[1]{%
    {\AEIgroupfont #1}%
}

\newcommand{\AEIsub}[2]{%
    \shortstack[c]{%
        {\AEIsubfont #1}\\[-1.4pt]
        {\AEIsubfont #2}
    }%
}

\begin{document}

\title{Towards Agentic Embodied Intelligence}
\author{AEI Team 
}



\abstract{
Embodied Intelligence represents the ultimate frontier of bridging artificial intelligence with physical reality to achieve general-purpose physical agency. Recently, the infusion of multimodal foundation models has catalyzed an end-to-end sensorimotor paradigm, translating internet-scale semantic representations and human intents into motor actions.
However, the prevailing monolithic paradigm treats continuous physical interaction as an unconstrained, open-loop sequence generation task. 
Consequently, it suffers from catastrophic compounding covariate drift, temporal attention dispersion over long horizons, and a complete absence of in-situ error recovery, fundamentally confining current systems to fragile, human-dependent conditional autonomy.
We argue that transcending this plateau cannot be achieved by merely scaling feedforward policy weights; rather, the decisive breakthrough demands a fundamental paradigm shift toward \textbf{Agentic Embodied Intelligence} (AEI). 
In this survey, we examine embodied systems through an agentic lens, systematically characterizing the paradigm shift toward AEI across models, infrastructure, safety, data, environments, and evaluation. 
We formalize a cognitive closed loop: elucidating how physical grounding, deliberative planning, motor execution, and in-situ reflection dynamically interlock to bridge the chasm between abstract reasoning and continuous physical dynamics. 
Recognizing that the community remains in the nascent formative stage of building AEI, we critically dissect the prevailing open challenges across physical fidelity, real-time latency, and trustworthy deployment. 
We envision this work serving as a roadmap to guide future exploration toward resilient, general-purpose Physical AI in the open world.
}

\checkdata[
\raisebox{-0.2em}{\includegraphics[width=0.025\linewidth]{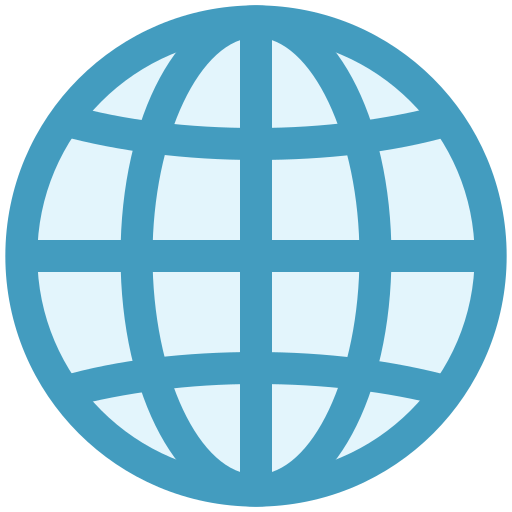}}~~Project Page]{\href{https://agenticei.github.io/AEI-Survey-page/}{\texttt{https://agenticei.github.io/AEI-Survey-page/}}
\\[-1.5ex]}


\maketitle

\begin{figure}[h]
  \centering
  \includegraphics[width=\linewidth]{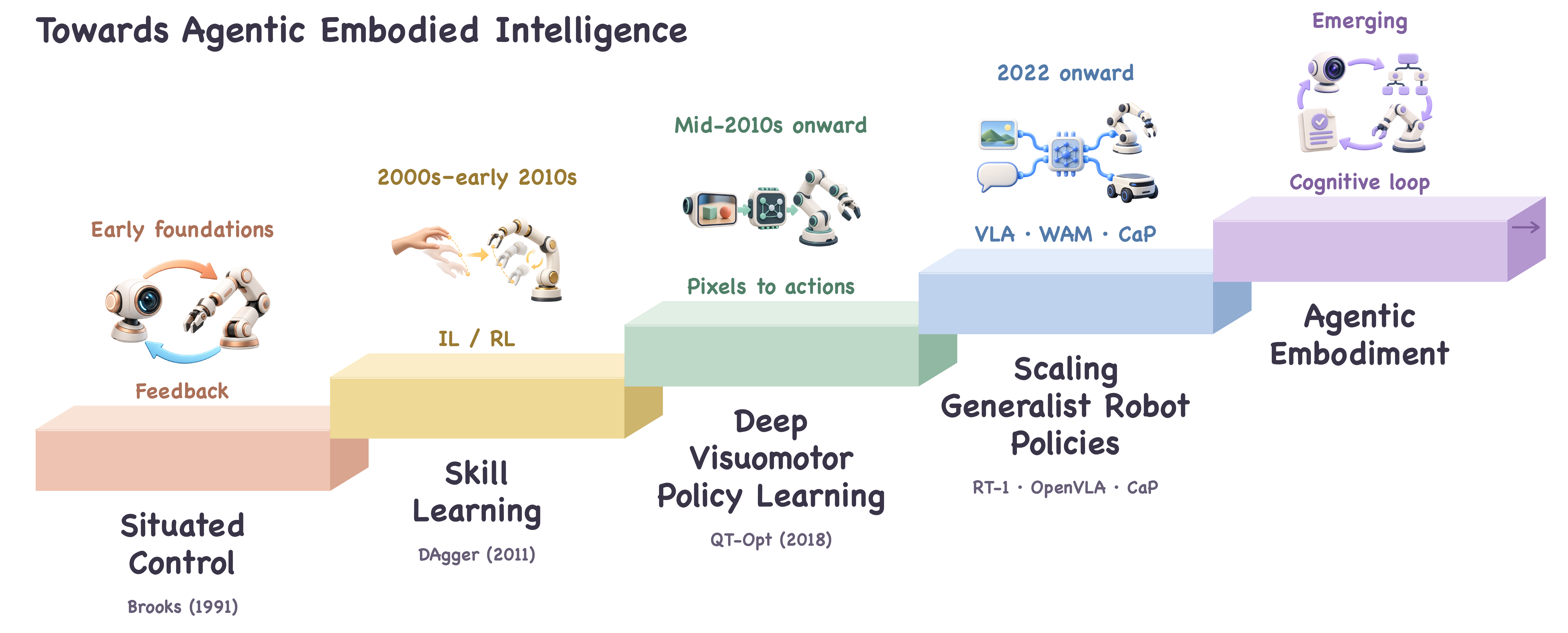}
\end{figure}

\vspace{1em}

\newpage
\tableofcontents   
\newpage

\section{Introduction}

\input{sec/1_intro}

\section{Conceptual Framework}
\label{sec:framework}
\input{sec/2_framework}

\section{Perception and Grounding}
\label{sec:perception}
\input{sec/3_perception}

\section{Reasoning and Planning}
\label{sec:planning}
\input{sec/4_planning}

\section{Action Execution}
\label{sec:action}
\input{sec/5_action}

\section{Closed-Loop Evolution}
\label{sec:loop}
\input{sec/6_evolution}

\section{Safety and Trustworthiness}
\label{sec:safety}
\input{sec/8_safety}

\section{Computing Infrastructure}
\label{sec:infra}
\input{sec/7_infra}

\section{Data Ecosystem}
\label{sec:data}
\input{sec/9_data}

\section{Simulators, Benchmarks and Evaluation Protocols}
\label{sec:bench}
\input{sec/10_bench}

\section{Open Challenges and Future Horizons}
\label{sec:future}
\input{sec/11_future}

\section{Conclusion}
\input{sec/12_conclusion}

\section*{Authors}
\addcontentsline{toc}{section}{Authors}
Contributors: Shengjun Zhang, Simin Huang, Chong Xia, Chensheng Dai, Cheng Nian, Xin Fei, Diankun Wu, Weichen Lyu, Min Chen, Mingyu Dong, Zhang Zhang, Yiwen Shen, Yueqi Duan

Institutions: Tsinghua University, National University of Singapore

\bibliographystyle{plain}
\bibliography{main}

\end{document}

%% file: sec/1_intro.tex
\begin{figure}
    \centering
    \includegraphics[width=\linewidth]{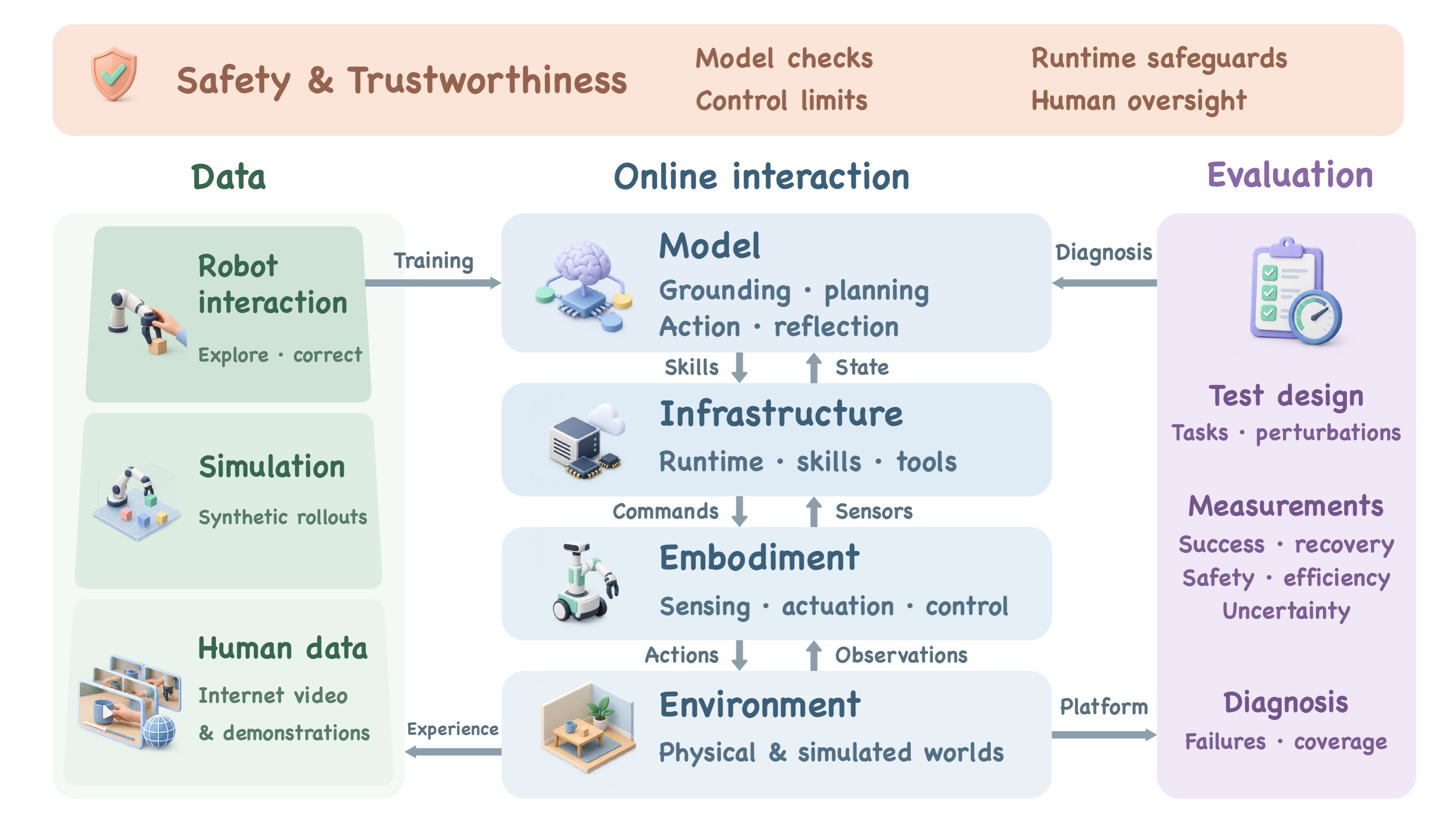}
    \caption{\textbf{System-level relationships in agentic embodied intelligence.}}
    \label{fig:system structure}
\end{figure}

The rapid advancement of large language models (LLMs) and multimodal foundation models has catalyzed a paradigm shift in artificial intelligence, moving beyond purely digital information processing toward agents that can perceive, reason, and act in the physical world, motivating the development of embodied intelligence. 
A prevalent but reductionist view frames this emerging field as simply \textit{installing large models onto robots}, yet the core principle of embodied intelligence is far more fundamental: it centers on the formation of a closed perception, cognition, and action loop embedded in the physical world, where intelligent behavior emerges from dynamic, reciprocal interactions between an agent, its body, and the environment, rather than from disembodied symbolic computation alone \cite{brooks1991intelligence, varela2017embodied, clark1997being, barsalou1999perceptual}. 
The notion of ``embodiment'' emphasizes that intelligence is inherently constrained by the agent's physical morphology, sensorimotor capabilities, actuator limits, and the laws of physics, and that knowledge and skills are acquired through active exploration, physical manipulation, and environmental feedback rather than passive data ingestion \cite{gibson2014ecological, lakoff1999review, moravec1988mind, ashby1957introduction}.

\subsection{Development of Embodied Systems}

The rise of embodied intelligence is often misframed as a sudden breakthrough spurred solely by recent large foundation models. In reality, its intellectual lineage spans several decades, evolving through successive paradigm shifts driven by the cross-pollination of cybernetics, cognitive science, robotics, and artificial intelligence. This long developmental trajectory traces a consistent intellectual arc: a steady departure from the classical view of intelligence as disembodied symbolic computation, toward a foundational consensus that intelligent behavior emerges from the closed-loop, reciprocal interactions among an agent’s physical body, its sensorimotor systems, and the constraints of the physical environment. Retracing this history is essential for contextualizing today’s rapid advances: it illuminates how long-standing theoretical insights have been gradually operationalized into engineering systems, how each technological wave has redefined the field’s core questions, and why the ongoing convergence of foundation models and robotics represents both a natural culmination of prior work and a qualitative leap toward general-purpose physical intelligence. 

\textbf{Situated Control.} 
Early embodied robotics challenged the idea that intelligent action must be produced by a single central model that first constructs a complete symbolic description of the world. Brooks's layered control system organized behavior around ongoing interaction with the environment, and his later ``Intelligence without Representation'' argued that robust competence could arise from situated perception--action coupling rather than elaborate internal world models \cite{brooks1991intelligence}. Developmental robotics made the sensorimotor loop itself an object of study: an agent's perceptual and action capabilities were treated as developing through interaction, rather than being supplied only as fixed modules \cite{lungarella2003developmental}. This early period is a conceptual and control-architecture foundation, not an embryonic version of today's pretrained embodied models. The systems were typically designed around particular robots, sensors, and behavioral requirements. Its lasting contribution is a framing principle: embodiment, action, and environmental feedback help determine what an agent can perceive and learn. Later data-driven methods changed how behaviors were acquired, but did not make physical interaction or feedback a new invention.

\textbf{Robot Skill Learning.} 
During the 2000s and early 2010s, robot learning consolidated demonstration-based learning~\cite{argall2009survey} and reinforcement learning~\cite{loop_kober2013rl} as systematic alternatives to specifying every behavior by hand. 
In physical robotics, apprenticeship learning enabled autonomous helicopter aerobatics from expert demonstrations, while policy search optimized robot motor primitives through interaction \cite{abbeel2010autonomous, kober2008policy}. DAgger~\cite{ross2011reduction} addressed the compounding errors that arise when an imitation policy visits states absent from its demonstrations by iteratively collecting expert labels on states induced by the learner. The learned unit was usually a task- or embodiment-specific skill: a policy for a defined manipulation, locomotion, or control problem, rather than a general-purpose model conditioned on open-ended instructions.

\textbf{Deep Visuomotor Policy Learning.}  
From roughly 2015, deep learning made it practical to learn visuomotor policies from pixels and robot experience. Previous works~\cite{levine2016end} demonstrated end-to-end training of deep visuomotor policies, and subsequent grasping work emphasized large-scale data collection and closed-loop, off-policy learning on physical robots~\cite{pinto2016supersizing,levine2018learning, kalashnikov2018scalable}. In parallel, sim-to-real transfer became an explicit research problem. For example, domain randomization~\cite{tobin2017domain} varied simulated appearance and dynamics to improve transfer of visual policies to real robots. 
These results expanded the scale and learning machinery for robot skills, but success remained tied to particular action spaces, tasks, data regimes, and robot platforms.
At the same time, a distinct embodied infrastructure~\cite{kolve2017ai2, shridhar2020alfred, savva2019habitat} emerged around interactive simulators and standardized tasks, providing simulation scenes in which agents could navigate and interact with environments. 
The physical-policy and simulator-benchmark lines advanced concurrently. Neither is a simple successor to the other. Together, they made embodied tasks more measurable and accelerated iteration, but benchmark performance in a simulator did not establish robust deployment on physical robots. The main change was that embodied intelligence increasingly became a reproducible learning-and-evaluation problem, not only a collection of robot-specific control demonstrations.

\textbf{Scaling Generalist Robot Policies.} 
While large scale pretraining demonstrates impressive generalizable capability in natural language processing,  researchers began to study whether robot skills could be learned as a shared policy from heterogeneous tasks, demonstrations, and embodiments, rather than as isolated controllers. 
Early instantiations~\cite{loop_driess2023palme, loop_brohan2022rt1} established the feasibility of transferring web-scale visual-linguistic priors into physical control. This trajectory crystallized into the Vision-Language-Action (VLA) paradigm~\cite{loop_brohan2023rt2, kim2024openvla, black2024pi_0} by transferring web-scale visual-semantic knowledge into robot action prediction. In parallel, world-action models (WAMs)~\cite{hu2024video, pai2025mimic, kim2026cosmos} add predictive structure to action learning. The motivation is to make temporal consequences available to action selection: instead of only asking which action matches the current input, the model can learn which action is expected to produce a useful next state.
Recently, with frontier models exemplified by next-generation agents~\cite{openai2026gpt, anthropic2026opus} demonstrating unprecedented software engineering proficiency, test-driven synthesis, and iterative traceback debugging, the Code-as-Policies (CaP)~\cite{loop_liang2022codepolicies} paradigm has commanded widespread renewed attention as a scalable bridge between abstract cognition and physical execution. Fundamentally, CaP reconceptualizes robot policy generation not as continuous motor torque regression or unstructured natural language dialogue, but as programmatic code synthesis. Under this formulation, a foundation model functions as a real-time software engineer. Given a complex, long-horizon instruction and real-time environmental context, it autonomously generates executable scripts that orchestrate a suite of perception tools and motor control primitives.

\textbf{Towards Open-World Agentic Embodiment.}  
A distinct, still-emerging direction treats the robot not merely as the endpoint of a policy call, but as an agent that repeatedly interprets a goal, gathers task-relevant evidence, selects among skills and tools, observes physical outcomes, and changes its plan when reality diverges from expectation. Current demonstrations generally operate over bounded task sets, finite skill libraries, short deployments, or controlled environments. They do not yet provide a general guarantee of calibrated perception, safe long-horizon execution, recovery from unanticipated failures, and continual improvement across arbitrary homes, workplaces, and robot bodies. The long-term goal is a system in which active perception selects what to inspect, a grounded planner chooses and revises subgoals, motor policies execute at the timescale of physical control, and outcome verification updates memory and future decisions. This organization is a plausible route toward autonomy in bounded operational domains and, eventually, toward the broader generality sought.

\input{tab/1_ladder}
\begin{figure}[t]
    \centering
    \includegraphics[width=\linewidth]{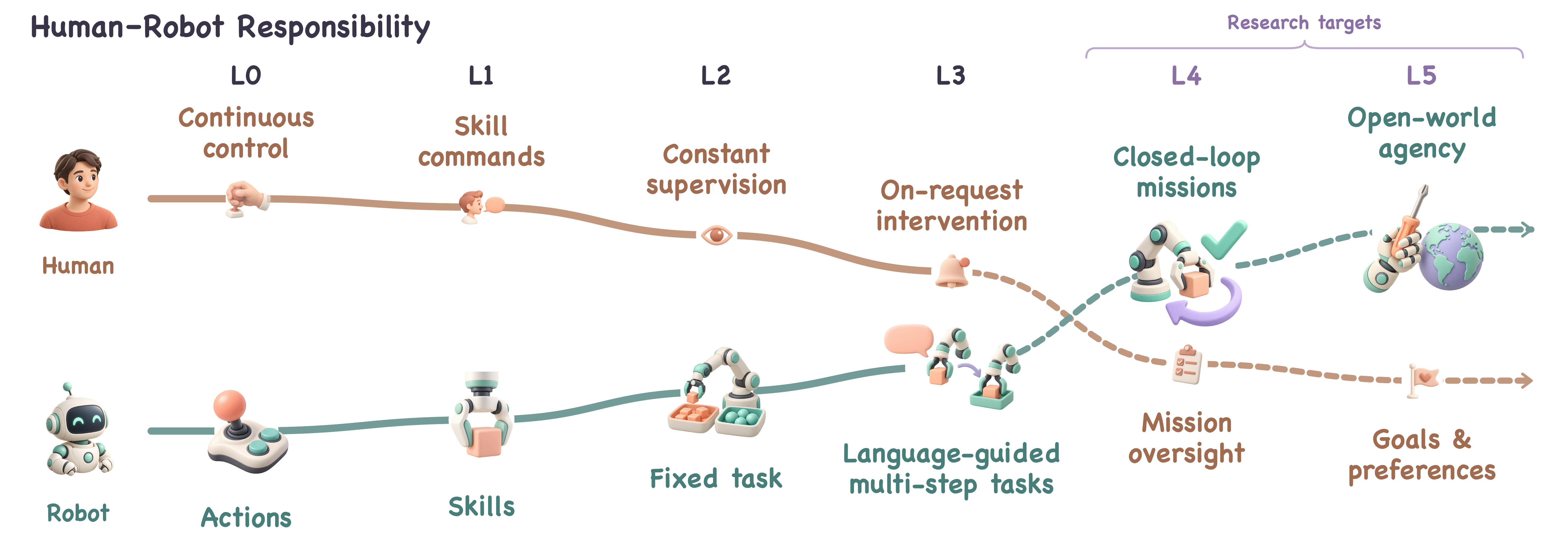}
    \caption{\textbf{Conceptual human–robot responsibility ladder from L0 to L5.} As robot capability advances from executing actions and skills to handling fixed tasks, language-guided multi-step tasks, closed-loop missions, and open-world agency, human involvement shifts from continuous control to skill-level commands, supervision, on-request intervention, mission oversight, and ultimately the specification of goals and preferences. The opposing trajectories illustrate this gradual transfer of responsibility.}
    \label{fig:ladder L0-L5}
\end{figure}

\subsection{Ladder of Embodied Autonomy}

To establish a rigorous taxonomy for robotic capability progression, we formalize the six levels of embodied autonomy in Figure~\ref{fig:ladder L0-L5}. This hierarchy systematically delineates the shifting boundaries of the control authority, and human-robot collaborative roles. Below, we dissect the state of maturity across these tiers, envision the frontier of L4/L5, and elucidate why \textit{Agentic Embodied Intelligence} serves as the foundational paradigm shift required to transcend the current plateau.

\textbf{L0 and L1.} Levels L0 and L1 represent the most mature baselines in robotics. L0 provides the sensory and kinematic backbone for safety-critical systems, notably surgical robotics and large-scale teleoperation data acquisition rigs~\cite{zhao2023learning}. 
L1 forms the core of modern industrial automation, where robots execute isolated, closed-loop atomic primitives, such as trajectory replay via teach pendants~\cite{ijspeert2013dynamical}, grasping primitives~\cite{mahler2017dex}, single-query path planning~\cite{kuffner2000rrt}. 
Currently, L0 and L1 play a pivotal role as the indispensable data-generation engines fueling the entire embodied learning pipeline across both physical and simulated domains.

\textbf{L2.} In Level L2, autonomous capability is fundamentally bounded by a closed-set task manifold, where deep robotic learning achieved near-perfect dynamic fidelity for isolated, predefined workflows in structured environments. This phase is epitomized by classical end-to-end visuomotor policies \cite{levine2016end}, scalable closed-loop bin-picking systems~\cite{kalashnikov2018scalable, fang2020graspnet}, and inductive spatial rearrangement architectures~\cite{zeng2021transporter}. 
Furthermore, modern high-precision imitation frameworks~\cite{chi2025diffusion, zhao2023learning}, alongside complex reinforcement learning specialists~\cite{akkaya2019solving}, demonstrated exquisite contact-rich motor dexterity. 
However, L2 controllers lack language comprehension and semantic reasoning. Their policy manifolds overfit strictly to known geometries and controlled lighting. Any out-of-distribution (OOD) perturbation immediately triggers execution failure, necessitating continuous, on-site human supervision.

\textbf{L3.} The progression from L2 to L3 marks the watershed transition from task-specific specialist controllers to open-vocabulary foundation policies. The integration of multimodal foundation models catalyzed the leap to L3, wherein physical actions are conditioned on open-vocabulary natural language instructions. Represented by landmark Vision-Language-Action (VLA) models~\cite{loop_brohan2022rt1, loop_brohan2023rt2, kim2024openvla, black2024pi_0} and World Action Models (WAM)~\cite{hu2024video, kim2026cosmos, pai2025mimic, li2026causal}, L3 generalist policies leverage internet-scale pretraining to zero-shot generalize across novel object instances, semantic synonyms, and diverse spatial layouts. 
Indeed, a capable, robust, and dexterous policy model remains an indispensable prerequisite and the physical bedrock upon which higher-tier L4 and L5 autonomy must be grounded.
Despite broad semantic flexibility, L3 remains fundamentally conditional. Because monolithic VLAs operate as open-loop trajectory rollouts without causal reasoning, active epistemic perception, or in-situ error diagnosis, they are intrinsically vulnerable to compounding covariate drift and physical deadlocks. 

\textbf{Vision for L4 and L5.}
The L4 target is a robot that can accept a mission, gather missing
information, execute and verify each step, recover from routine failures, and recognize when it must stop or raise an exception.
Unlike L3, which remains an open-loop execution regime, L4 demands system-level self-closure. 
Level L5 represents the ultimate aspiration of embodied intelligence: open-world, unbounded general-purpose physical agency. Operating in arbitrary, unstructured environments, an L5 system behaves as a fully competent physical partner. 
It requires zero specialized engineering oversight; non-expert end-users interact purely through abstract goals and subjective preferences.
Recently, no evidence in the surveyed literature establishes sustained, unattended L4 operation across a broad, explicitly validated domain.

\textbf{Why the agentic loop matters for L4 and L5.} 
The transition from L2/L3 to L4/L5 changes the unit of autonomy from an action or demonstrated task to a sustained mission under uncertainty. 
A mission requires the robot to ground ambiguous goals, choose actions, compose and dispatch skills, verify physical outcomes, recover or ask for help, and retain useful experience. These are the functions of the \textit{perception, planning, execution, reflection} loop developed in this survey. In this functional sense, agentic embodied intelligence is necessary for L4 and L5: a system must possess these closed-loop capabilities. Currently, with the development of coding agents~\cite{openai2026gpt, anthropic2026opus}, it provides the cognitive syntax, execution rigor, and self-debugging mechanics that make physical closed-loop self-healing mathematically and operationally feasible~\cite{zhang2026harness, ding2026zetta}.

\textbf{History of embodied intelligence.} 
The historical trajectory also has a loose correspondence with the autonomy levels: direct teleoperation is associated with L0; manually specified or learned atomic skills\cite{abbeel2010autonomous, kober2008policy} with L1; task-specific visuomotor policies~\cite{pinto2016supersizing,levine2018learning, kalashnikov2018scalable} and controlled simulation or robot learning with L2; and language-conditioned skill composition and generalist policy research~\cite{kim2024openvla, hu2024video, loop_liang2022codepolicies} with capabilities relevant to bounded L3 autonomy. This is not a one-to-one mapping: autonomy levels characterize system-level responsibility, operating domains, and human supervision, whereas historical stages describe evolving research paradigms. A detailed historical discussion is provided in Online Resource~1, Section~1.

\subsection{Scope and Contribution}

\textbf{Scope.} We study AEI from a system-level perspective. 
As shown in Figure~\ref{fig:system structure}, the central embodied core is the online interaction loop: the model grounds observations, plans, acts, and reflects; infrastructure translates its skills into commands for the robot; and the robot interacts with the physical or simulated environment, returning sensor data and state through the infrastructure to the model. The environment generates interaction data for training, and provides platforms for evaluating models through tasks and perturbations. Evaluation results and failure diagnoses then inform model development. Safety and trustworthiness cut across every part of this system, from model checks and control limits to runtime safeguards and human oversight.

\textbf{Framework.} Accordingly, Sections~\ref{sec:perception}--\ref{sec:loop} trace how this loop works in practice across perception, planning, execution and evolution. 
Section~\ref{sec:safety} treats safety and trustworthiness as constraints on the entire lifecycle. 
Section~\ref{sec:infra} examines the computing infrastructure, runtimes, interfaces, and hardware that coordinate model inference with robot execution across device, edge, and cloud resources. 
Section~\ref{sec:data} follows that experience from data sources and curation through training and the embodied data flywheel, while Section~\ref{sec:bench} considers simulators, benchmarks, and evaluation protocols as controlled environments for testing capabilities, diagnosing failures, and measuring whether progress transfers to physical deployment. 
Section~\ref{sec:future} discusses the cross-cutting open challenges and future directions emerging from this synthesis.

\textbf{Contributions.} In this survey, we present a system-level framework to analyze agentic embodied intelligence. We propose an L0--L5 autonomy ladder to relate robot capability to human responsibility and operational boundaries.
We organize the model core into four coupled functions: perception and grounding, reasoning and planning, action execution, and closed-loop evolution. We also connect the online path from model to infrastructure, embodiment, and environment with the feedback path that returns observations and outcomes. We synthesize the enabling system layers that connect these functions to deployment, including system-level safety, device-edge-cloud computation, and an embodied data and evaluation ecosystem.

%% file: tab/1_ladder.tex
\begin{table*}[t]
\centering
\small
\renewcommand{\arraystretch}{1.35}
\caption{\textbf{Taxonomy of Embodied Autonomy Levels.}}
\label{tab:embodied_autonomy_levels}
\begin{tabularx}{\textwidth}{c p{3cm} X X}
\toprule
\textbf{Level} & \textbf{Level Name} & \textbf{Capability Definition} & \textbf{Human Role} \\
\midrule
\rowcolor{AEIrow}
\textbf{L0} & \textbf{Direct\newline Teleoperation} & 
Direct mapping of continuous human motions to joint commands. & 
Physically controls every action frame in real-time. \\

\textbf{L1} & \textbf{Assisted\newline Primitives} & 
Executes single closed-loop atomic skills, while task decomposition and sequencing remain manual. & 
Decomposes tasks and sequentially issues atomic commands. \\

\rowcolor{AEIrow}
\textbf{L2} & \textbf{Closed-set\newline Autonomy} & 
Autonomously executes a complete, predefined single task in controlled settings. & 
Maintains constant visual line-of-sight; ready for immediate physical takeover. \\

\textbf{L3} & \textbf{Conditional\newline Autonomy} & 
Comprehends natural language; accomplishes multi-step composite tasks under moderate variations. & 
Not required for continuous monitoring, but must intervene upon request within a time budget. \\

\rowcolor{AEIrow}
\textbf{L4} & \textbf{High Domain\newline Autonomy} & 
Fully autonomous closed-loop execution; features long-horizon multi-step workflows. & 
Zero on-site physical intervention; manages macro missions and handles asynchronous exception tickets. \\

\textbf{L5} & \textbf{Full General\newline Autonomy} & 
Operates unbounded open worlds with human-level commonsense reasoning, tool-use transfer, active perception, and lifelong self-evolution. & 
Requires zero technical expertise; interacts naturally via high-level goals and preferences. \\
\bottomrule
\end{tabularx}
\end{table*}

%% file: sec/2_framework.tex

\subsection{System Organization}

Agentic embodied intelligence is best understood as a system rather than as a language model connected directly to a robot. As shown in Figure~\ref{fig:system structure}, the model interprets goals, estimates task-relevant state, and proposes plans or actions; an execution infrastructure mediates these proposals through skill and tool interfaces, scheduling, and communication; the embodiment supplies sensors, actuators, and low-level control; and the environment returns observations and physical consequences. The resulting interaction is bidirectional: commands flow from model to infrastructure and robot, while observations, contact signals, and verified outcomes flow back to the model. This separation allows semantic reasoning to operate at a slower timescale than stabilization and contact control, while keeping both accountable to the same task state. 

The online stack is complemented by two offline-facing subsystems. The data ecosystem collects demonstrations, robot trajectories, and interaction outcomes, then curates them for pretraining, adaptation, or policy improvement; large-scale cross-embodiment resources make this learning loop increasingly important . Evaluation provides repeatable tasks, perturbations, and measurements of success, recovery, efficiency, uncertainty, and safety events. Its results should diagnose coverage gaps and failures, guide targeted data collection, and inform model or system updates, rather than serve only as a final leaderboard score. Safety and trustworthiness cut across both offline and online operation: model-level uncertainty checks, runtime validation, control limits, monitoring, and human takeover jointly constrain what the system may do. In short, the model proposes; infrastructure mediates; the embodied controller acts; the world responds; and data, evaluation, and safety determine how the system learns and where its operation must stop.

\subsection{Model Cognitive Loop}

\begin{figure}[t]
    \centering
    \includegraphics[width=\linewidth]{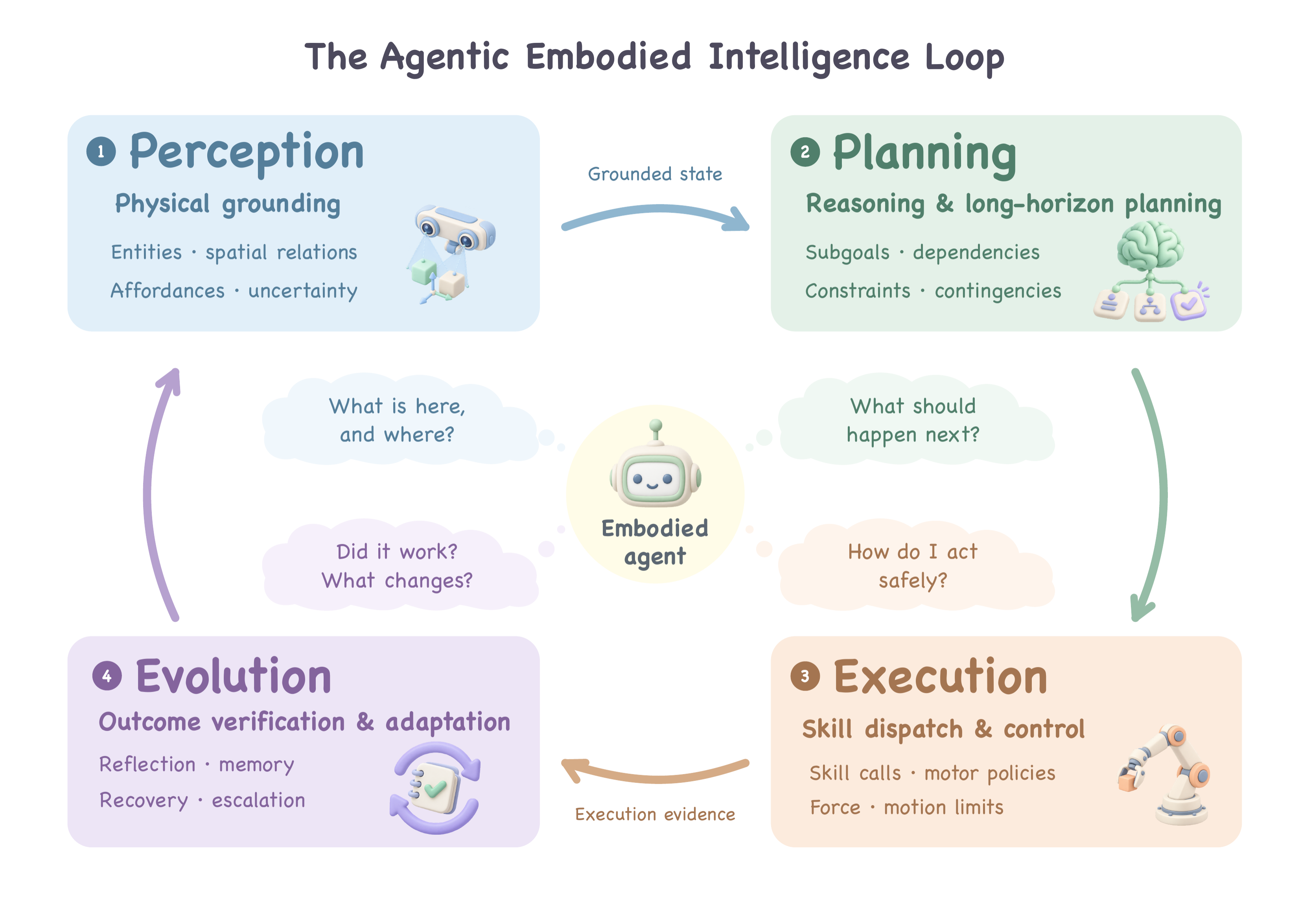}
    \caption{\textbf{The model cognitive loop of agentic embodied intelligence.}}
    \label{fig:model stage}
\end{figure}

The model core comprises four interdependent functions, not four isolated stages. Perception and physical grounding transform multimodal observations and task context into robot-relevant entities, spatial relations, affordances, and a state estimate that preserves uncertainty. Cognitive reasoning and long-horizon planning use this grounded state to interpret the goal, decompose it into subgoals, reason about dependencies and constraints, and decide whether more information is needed before acting. Action execution translates a selected plan into a skill call, motion target, or motor policy and carries it out through the appropriate controllers. Closed-loop evolution compares expected and observed outcomes, updates memory and state beliefs, diagnoses failure, and chooses among continuing, replanning, retrying, gathering more evidence, or handing control to a person. The forward path and return path are summarized in Figure~\ref{fig:model stage}.

The interfaces between these functions are central to agency. Perception must provide more than a fluent scene description: it should expose grounded references, spatial frames, and uncertainty that planning can use to test preconditions. Planning, in turn, should issue executable and monitorable subgoals rather than an unstructured narrative. Execution must return observations and action status, including failures and partial completion, instead of reporting only that a command was issued. Finally, closed-loop evolution must treat physical evidence as authoritative: it updates the working belief and memory, then routes the result to perception when the scene is ambiguous, to planning when the current plan is invalid, or to execution when a bounded retry is justified. Prior work on feasibility-aware language planning and feedback-conditioned replanning provides early examples of these interfaces; learned visuomotor and diffusion policies provide complementary action mechanisms but still require outcome monitoring.

This organization is a functional decomposition, not a claim that every system must contain four separately trained modules. A unified VLA model may internalize parts of perception, reasoning, and action, while external infrastructure supplies memory, monitoring, and safety checks. What makes the system agentic is that the functions remain causally connected: physical observations can change beliefs, beliefs can change plans, plans can dispatch bounded actions, and measured outcomes can change what happens next. Without this return path, an apparent multi-step capability is only open-loop action generation; with it, the agent can detect divergence between intended and actual state and adapt its behavior accordingly.

%% file: sec/3_perception.tex
An embodied agent relies on perception to supply task-relevant evidence about its environment to plan and execute actions. We distinguish three coupled dimensions of the estimated state: semantic grounding to identify objects, spatial grounding to determine their poses, and physical grounding to assess interaction feasibility.
Yet, this information is incomplete, where occlusion hides relevant objects and states, sensor noise limits estimation accuracy, and both the agent's actions and external events change the environment. An agentic system therefore maintains its estimates throughout the task and recognizes when additional evidence is needed via active perception.

Accordingly, this section is organized around the content of task-relevant state estimates and the mechanisms that keep them useful. Sections~\ref{sec:perception-semantic}, \ref{sec:perception-spatial}, and~\ref{sec:perception-physical} review semantic, spatial, and physical grounding, respectively. Section~\ref{sec:perception-update} then examines passive belief maintenance and active and interactive perception across these three dimensions. Together, these capabilities supply the state estimates, feasibility evidence, and outcome observations used by downstream planning, action execution, and closed-loop reflection.

\input{tab/3_taxonomy}
\input{fig/3_taxonomy}

\subsection{Semantic Grounding}
\label{sec:perception-semantic}

\textbf{Appearance Grounding.}
Category names and visual attributes provide a direct link between language and image content. Vision and language pretraining, exemplified by CLIP~\cite{clip}, extends recognition beyond a fixed set of labels. Open vocabulary detection then carries this semantic alignment from whole images to specific regions. Early approaches established this capability through visual concept transfer or text prompted detection, while later systems brought richer phrases into localization through joint training on detection and grounding data~\cite{vild,owlvit,glip,groundingdino}. The shared goal is to identify the image region that best matches a language description.
Recognizing individual objects, however, does not guarantee that their attributes and relations are interpreted correctly. A model may detect cups, bowls, and colors yet select the wrong object for the phrase ``the red cup.'' ARO~\cite{aro} exposes this gap in attribute binding and relational understanding. Grounding also needs to reach beyond whole objects. Identifying ``the drawer handle'' requires locating the correct part within the correct object. Image and point cloud methods extend semantic recognition to this finer level, providing candidate referents whose geometry must then be resolved by spatial grounding~\cite{vlpart,partslip}.

\textbf{Contextual Grounding.}
Some expressions can only be resolved by considering the surrounding scene. In ``the cup next to the kettle,'' the identity of the target depends on both objects and their relation. Methods differ in how they connect language to candidate regions. Some integrate the expression into detection, some associate generated text with localized regions, and others expose candidate regions through visual identifiers for a multimodal model to select~\cite{mdetr,kosmos2,som}. Despite these differences, all must determine whether the selected objects satisfy the relation in the instruction. Recognizing the cup and kettle separately is not enough.
Other references depend on information beyond the current image. ``My cup'' requires knowledge of a user specific object, while ``the cup I used earlier'' also requires access to interaction history. Reference images and learned user concepts can help resolve personal references, but historical references additionally require the agent to maintain object identity across observations~\cite{myvlm,bringmycup}. 

\textbf{Function Grounding.}
A task may describe an object's intended use rather than its name. For example, ``find something to cut the tape with'' requires the agent to identify a suitable object from the scene. One route uses the task description to guide object or region selection. Another relies on implicit reasoning about the purpose expressed by the instruction~\cite{li2022toist,lisa}. Both shift grounding from matching explicit object names toward identifying entities that may serve the user's goal.
Observed interactions provide a further source of functional evidence. They can link object regions to the ways people use them, helping an agent identify a relevant tool or part even when the object is unfamiliar~\cite{roboabc}. Functional relevance does not establish physical feasibility. Scissors may match the intended purpose but be unsuitable for the material or impossible for the robot to operate. These questions belong to physical grounding in Section~\ref{sec:perception-physical}.


\subsection{Spatial Grounding}
\label{sec:perception-spatial}

Spatial grounding connects observed entities and task referents to the geometry of the environment. It combines geometric estimation, which recovers shape, scale, position, and orientation, with spatial reasoning, which interprets relations between entities from a specified perspective. For example, estimating the geometry of a cup and a drawer provides evidence for judging whether the cup is inside the drawer. The two capabilities are interdependent. Geometric estimates support relational judgments, while the task determines which quantities and relations matter.

\textbf{Geometric Estimation}
Depth and scale provide the geometric basis for locating surfaces and measuring spatial extent. Large scale visual training has improved the generality of monocular depth estimation, while metric depth and stereo methods address the need to recover physical scale from visual evidence~\cite{depthanything,depthanythingv2,metric3dv2,unidepth,foundationstereo}. These approaches rely on different sources of scale information. Monocular predictions depend more heavily on learned priors, whereas stereo estimates depend on camera calibration. Their outputs therefore carry different assumptions and uncertainty.
Multiple views provide constraints on surfaces that are only partly visible in one image. Combining views requires matching corresponding regions, estimating camera poses, and aligning observations in a shared scene. Recent reconstruction systems increasingly learn these steps together, while other methods incorporate available calibration, pose, or depth information to improve geometric consistency~\cite{dust3r,mast3r,vggt,mapanything}. Once views are aligned, semantic features can be attached to locations in maps, radiance fields, or Gaussian representations. These representations allow an agent to locate task relevant entities beyond the current image~\cite{vlmaps,conceptfusion,lerf,langsplat}.
Object pose estimation locates an entity and determines its orientation within the scene. Existing methods vary in the object knowledge they require. Some estimate poses for known instances, others generalize across instances of a category, and reference based approaches extend estimation to unfamiliar objects. When no reference model is available, reconstruction can be combined with tracking to recover shape and pose from observation~\cite{posecnn,nocs,megapose,foundationpose,bundlesdf}. Occlusion and symmetry may still leave multiple poses consistent with the evidence. Moreover, an estimate in the camera frame must be aligned with the robot frame through calibration and, when needed, robot kinematics. This alignment is essential for turning an observed object or part into a target for planning and control.

\textbf{Spatial Reasoning}
Spatial reasoning determines what the geometry implies. It supports judgments such as which object is nearer, whether one region contains another, and how observations fit into a shared layout. These tasks require more than identifying the objects involved. Existing evaluations reveal a progression from interpreting visible spatial relations to comparing depth and integrating evidence across video and multiple views~\cite{whatsup,blink,vsibench}. As the evidence becomes distributed across viewpoints and time, the agent must maintain consistent spatial beliefs rather than answer from isolated images.
Spatial relations also depend on perspective. A change in viewpoint can alter the correct description even when the scene itself is unchanged. Situated spatial reasoning therefore requires the agent to establish the observer's position and orientation before interpreting a relation. Benchmarks and methods that make viewpoint explicit show that this information must be part of the reasoning process, while robot focused spatial supervision distinguishes egocentric, object centered, and world centered descriptions~\cite{sqa3d,sig3d,robospatial}.
Addressing these difficulties requires training signals and inputs that preserve spatial information. Geometrically grounded supervision strengthens predictions about directions, distances, and regions, while depth and three dimensional features provide information that may not be recoverable from appearance alone~\cite{spatialvlm,spatialrgpt,3dllm}. Verifiable spatial tasks can also provide a training signal through outcomes that are checked against image structure or depth~\cite{spatialssrl}. More reasoning is not always better, however. Evidence that extended deliberation can improve some reasoning tasks while weakening low level perception suggests that training should reward correct spatial judgments, not reasoning length alone~\cite{spatialtree}.
For questions involving several entities or views, explicit representations help organize the evidence used in reasoning. Scene graphs and cognitive maps make objects, places, and their relations available for later queries, while executable visual procedures allow spatial computation to be combined with perception tools~\cite{hydra,gu2023conceptgraphs,mindcube,vadar}. These approaches make intermediate information easier to inspect, but errors in detection, data association, or geometry can still propagate into the final judgment.

\subsection{Physical Grounding}
\label{sec:perception-physical}

Physical grounding connects task relevant objects and candidate interactions to physical properties, states, and constraints. It provides evidence about what an action is likely to change and whether the change is feasible. Lifting a mug, for example, requires more than locating its handle. The load and contact conditions also affect whether the mug may slip and whether the robot can maintain a stable grasp. We organize physical grounding around three related questions: what physical conditions hold, what consequences an interaction may produce, and whether those consequences satisfy the task within the robot's capabilities.


\textbf{Estimating Physical Conditions.}
Physical conditions include persistent properties such as mass and stiffness, as well as changing states such as loading and contact. No single observation source reveals all of them. Appearance can suggest material and physical attributes, while observed motion provides evidence about hidden properties through their effects. Interaction can reveal mechanical constraints, such as whether a part moves or articulates~\cite{physobjects,wu2015galileo,ditto}. These sources provide complementary evidence rather than interchangeable estimates.
Touch adds local information that vision may miss. Tactile observations can reveal contact geometry, force, and slip, while tactile and language representations can support descriptions of object properties. As interaction continues, visual priors and sensor evidence can be combined with estimates of uncertainty~\cite{gelsight,octopi,phys2real}. The resulting evidence may describe an attribute, estimate a physical parameter, or report a contact state. Each constrains different aspects of the next interaction.



\textbf{Predicting Interaction Outcomes.}
Given the estimated conditions, physical prediction analyzes how an action changes an object or scene. Explicit dynamics models represent changes through interacting objects, relations, or particles. This makes the modeled entities and interactions accessible, but the prediction depends on accurate state estimates. Latent world models take a different approach by predicting future representations conditioned on an action without recovering every physical quantity explicitly~\cite{interactionnet,dpinet,adaptigraph,vjepa2}. The approaches therefore differ in whether they expose physical variables or prioritize prediction in a learned representation.
Another key question is how an outcome would change under a different initial condition. This counterfactual question differs from forecasting the scene under its observed conditions. Methods that infer latent physical factors from an observed event can reuse those factors to evaluate altered configurations~\cite{baradel2020cophy}. For an embodied task, the relevant prediction is the expected change in object state or contact under a proposed action and uncertain physical conditions.



\subsection{From Passive Perception to Active Perception}
\label{sec:perception-update}

The environment is not static when the robot explores or conducts missions.
Therefore, perceptual updates must both incorporate new evidence and identify when the available evidence is insufficient. \emph{Passive belief maintenance} updates the estimate from incoming observations, while \emph{active and interactive perception} selects actions to obtain missing evidence.

\subsubsection{Passive Perception}
\label{sec:perception-maintenance}

Passive perception continuously processes incoming sensory streams to convert raw observations into a persistent and consistent representation of the world without actively perturbing the environment. This process centers on two coupled challenges: establishing spatio-temporal consistency and managing dynamic state validity over time. 

First, correspondence, fusion, and representation resolve object identity across varying viewpoints, sensor noise, and partial occlusions by tracking point correspondences, temporal segmentation masks, and 6D poses~\citep{sam2,cotracker,foundationpose,bundlesdf}, integrating tactile contact cues when the end-effector occludes visual features during manipulation~\citep{neuralfeels,vhop}. Across wider environments and longer trajectories, SLAM backends and multi-view association reconcile accumulated drift and consolidate disparate views into unified spatial representations, structured scene graphs, or hierarchical episodic memories~\citep{orbslam3,hydra,gu2023conceptgraphs,onemap,remembr,3dmem,embodiedrag}. These structured representations compress history into queryable, persistent states~\citep{cut3r}, allowing selective retrieval of fine-grained spatiotemporal evidence for immediate control without storing redundant sensory streams.

Second, change detection and temporal validity handle environmental non-stationarity by disentangling static background structures from movable or dynamic objects across different timescales~\citep{panopticmtsdf,khronos,dovsg,dynamem}. Crucially, this mechanism accounts for memory staleness and object volatility~\citep{ltmem}, explicitly distinguishing verified physical state changes from temporary occlusions or unobserved absences to ensure that downstream planning modules do not mistake stale historical evidence for the current state of the world.

\subsubsection{Active Perception}
\label{sec:perception-active}

When passive observations leave critical decision variables unresolved, active perception treats sensing as a control problem~\citep{bajcsy1988,aloimonos1988,bajcsy2018}, and leverages physical action to elicit diagnostic sensory feedback~\citep{bohg2017interactive}. Rather than maximizing spatial coverage or accumulating generic training experience, the central objective is to resolve uncertainty that directly impedes a pending downstream decision. This proactive evidence-gathering operates through three complementary mechanisms targeting different sources of ambiguity.

First, to overcome line-of-sight limitations, occlusions, and uninformative viewing angles, the agent performs viewpoint selection and targeted search. Candidate viewpoints and spatial frontiers are prioritized by balancing motion costs against the expected information gain for the specific unresolved variable—utilizing geometric reconstruction metrics~\citep{connolly1985,isler2016}, semantic maps and vision--language guidance~\citep{semexp,vlfm}, or task-conditioned viewing policies~\citep{lookaround,avaloha}, rather than treating all unexplored areas as equally valuable.

However, an optimal viewpoint cannot uncover internal material properties or occluded interiors. This necessitates physical probing and scene intervention, where the agent couples physical actions with resulting sensory changes to reveal compliance, articulation, and hidden contents~\citep{bohg2017interactive,ditto,agrawal2016learning}. Such interventions are orchestrated via belief-space planning under occlusion~\citep{msbp} and action-conditioned scene representations~\citep{roboexp}, requiring a balance between diagnostic value and the risks of irreversible scene disruption or physical harm.

Finally, when ambiguity arises from subjective human intent rather than unobserved physical states, the agent resorts to human clarification. Interactive grounding and belief-driven query frameworks formulate discriminating, scene-grounded questions to resolve candidate referents~\citep{ingress,invigorate}, while conformal prediction sets identify when multiple hypotheses warrant requesting assistance~\citep{ren2023robotsask}. Because human attention is a finite operational resource, clarification is triggered when a targeted question can resolve the ambiguity at a lower cost than autonomous sensing or probing.

Maintenance cannot resolve ambiguities that the available observations do not distinguish. Active perception treats sensing as a control problem~\citep{bajcsy1988,aloimonos1988,bajcsy2018}, and interactive perception uses physical action to generate informative sensory responses~\citep{bohg2017interactive}.
Active information gathering requires a calibrated stopping rule~\citep{ren2024exploreconfident}. Verification actively probes the expected consequences of an action, utilizing multimodal success feedback and spatio-temporal execution constraints to monitor progress~\citep{loop_huang2022innermonologue,codeasmonitor}. By matching the timing and modality of evidence to critical failure modes, the agent differentiates verified success, detected failure, and persistent uncertainty, determining whether to advance execution, initiate further probing, or trigger upstream reflection.

%% file: fig/3_taxonomy.tex
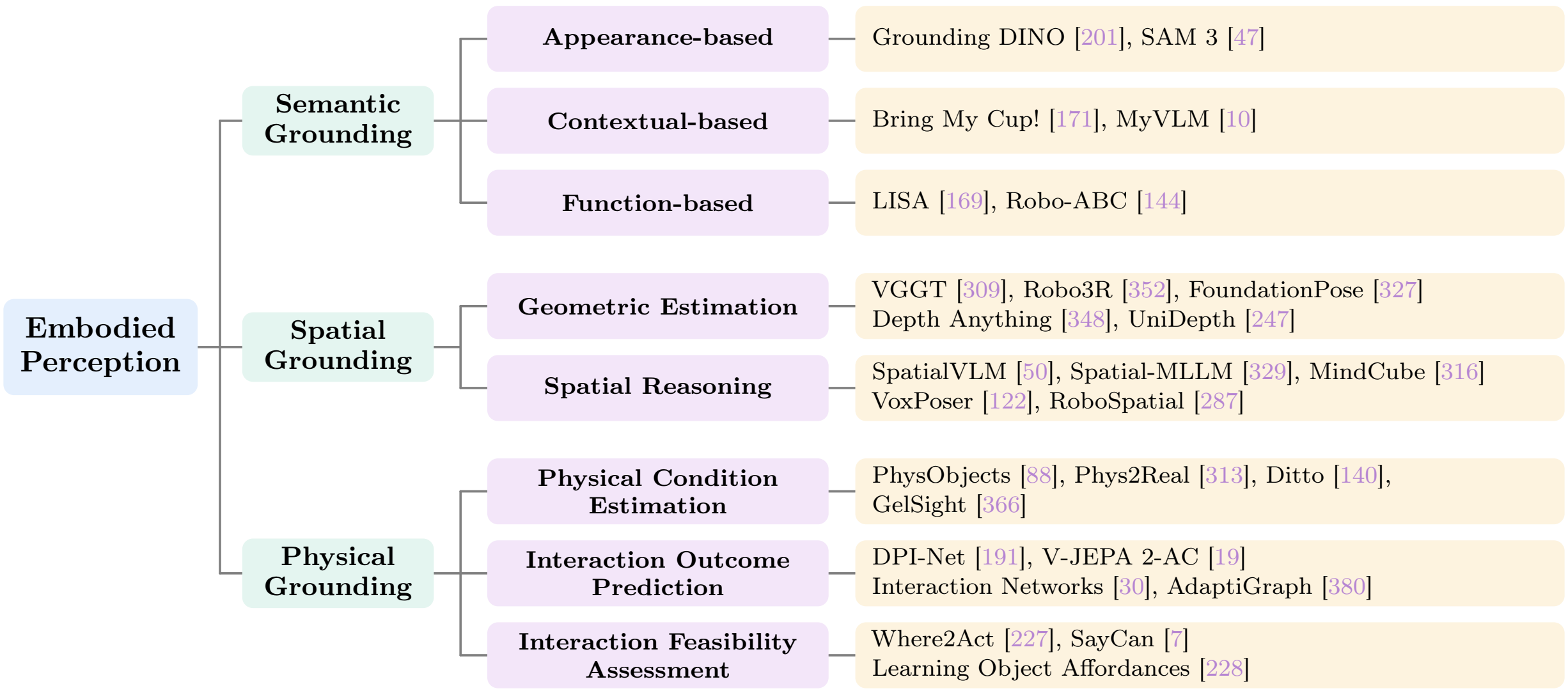
\begin{figure*}[t]
\centering
\resizebox{\linewidth}{!}{%
\begin{tikzpicture}[
  x=0.30pt, y=-0.30pt,
  figfourroot/.style={draw=none,fill=pwBlue!72!white,rounded corners=3.6pt,
    minimum width=65.4pt,minimum height=32.4pt,align=center,inner sep=0pt,
    font=\fontsize{10.2pt}{11.5pt}\selectfont\bfseries},
  figfourgroup/.style={draw=none,fill=pwGreen!72!white,rounded corners=3.6pt,
    minimum width=64.5pt,minimum height=23.4pt,align=center,inner sep=0pt,
    font=\fontsize{9pt}{10.1pt}\selectfont\bfseries},
  figfourtopic/.style={draw=none,fill=pwPurple!72!white,rounded corners=3.6pt,
    minimum width=114.9pt,minimum height=22.2pt,align=center,inner sep=0pt,
    font=\fontsize{8.4pt}{9.4pt}\selectfont\bfseries},
  figfourwork/.style={draw=none,fill=pwYellow!72!white,rounded corners=3.6pt,
    text width=227.4pt,minimum width=238.8pt,minimum height=22.2pt,
    align=left,inner sep=0pt,font=\fontsize{8.4pt}{9.9pt}\selectfont},
  figfourlink/.style={draw=black!50,line width=.78pt,line cap=round,line join=round}
]
\path[use as bounding box] (0,0) rectangle (1800,814);
\draw[figfourlink] (242,407.0) -- (292,407.0);
\draw[figfourlink] (267,153.0) -- (267,661.0);
\draw[figfourlink] (267,153.0) -- (292,153.0);
\draw[figfourlink] (267,661.0) -- (292,661.0);
\draw[figfourlink] (507,153.0) -- (537,153.0);
\draw[figfourlink] (537,61) -- (537,245);
\draw[figfourlink] (537,61) -- (567,61);
\draw[figfourlink] (950,61) -- (980,61);
\draw[figfourlink] (537,153) -- (567,153);
\draw[figfourlink] (950,153) -- (980,153);
\draw[figfourlink] (537,245) -- (567,245);
\draw[figfourlink] (950,245) -- (980,245);
\draw[figfourlink] (507,407.0) -- (537,407.0);
\draw[figfourlink] (537,361) -- (537,453);
\draw[figfourlink] (537,361) -- (567,361);
\draw[figfourlink] (950,361) -- (980,361);
\draw[figfourlink] (537,453) -- (567,453);
\draw[figfourlink] (950,453) -- (980,453);
\draw[figfourlink] (507,661.0) -- (537,661.0);
\draw[figfourlink] (537,569) -- (537,753);
\draw[figfourlink] (537,569) -- (567,569);
\draw[figfourlink] (950,569) -- (980,569);
\draw[figfourlink] (537,661) -- (567,661);
\draw[figfourlink] (950,661) -- (980,661);
\draw[figfourlink] (537,753) -- (567,753);
\draw[figfourlink] (950,753) -- (980,753);
\node[figfourroot] (perception) at (133.0,407.0) {Embodied\\Perception};
\node[figfourgroup] (semantic-grounding) at (399.5,153.0) {Semantic\\Grounding};
\node[figfourtopic] (approach-1) at (758.5,61) {Appearance-based};
\node[figfourwork] (works-1) at (1378.0,61) {\mbox{Grounding DINO~\citep{groundingdino}}, \mbox{SAM 3~\citep{sam3}}};
\node[figfourtopic] (approach-2) at (758.5,153) {Contextual-based};
\node[figfourwork] (works-2) at (1378.0,153) {\mbox{Bring My Cup!~\citep{bringmycup}}, \mbox{MyVLM~\citep{myvlm}}};
\node[figfourtopic] (approach-3) at (758.5,245) {Function-based};
\node[figfourwork] (works-3) at (1378.0,245) {\mbox{LISA~\citep{lisa}}, \mbox{Robo-ABC~\citep{roboabc}}};
\node[figfourgroup] (spatial-grounding) at (399.5,407.0) {Spatial\\Grounding};
\node[figfourtopic] (approach-4) at (758.5,361) {Geometric Estimation};
\node[figfourwork] (works-4) at (1378.0,361) {\mbox{VGGT~\citep{vggt}}, \mbox{Robo3R~\citep{robo3r}}, \mbox{FoundationPose~\citep{foundationpose}}\\\mbox{Depth Anything~\citep{depthanything}}, \mbox{UniDepth~\citep{unidepth}}};
\node[figfourtopic] (approach-5) at (758.5,453) {Spatial Reasoning};
\node[figfourwork] (works-5) at (1378.0,453) {\mbox{SpatialVLM~\citep{spatialvlm}}, \mbox{Spatial-MLLM~\citep{spatialmllm}}, \mbox{MindCube~\citep{mindcube}}\\\mbox{VoxPoser~\citep{huang2023voxposer}}, \mbox{RoboSpatial~\citep{robospatial}}};
\node[figfourgroup] (physical-grounding) at (399.5,661.0) {Physical\\Grounding};
\node[figfourtopic] (approach-6) at (758.5,569) {Physical Condition\\Estimation};
\node[figfourwork] (works-6) at (1378.0,569) {\mbox{PhysObjects~\citep{physobjects}}, \mbox{Phys2Real~\citep{phys2real}}, \mbox{Ditto~\citep{ditto}}, \mbox{GelSight~\citep{gelsight}}};
\node[figfourtopic] (approach-7) at (758.5,661) {Interaction Outcome\\Prediction};
\node[figfourwork] (works-7) at (1378.0,661) {\mbox{DPI-Net~\citep{dpinet}}, \mbox{V-JEPA 2-AC~\citep{vjepa2}}\\\mbox{Interaction Networks~\citep{interactionnet}}, \mbox{AdaptiGraph~\citep{adaptigraph}}};
\node[figfourtopic] (approach-8) at (758.5,753) {Interaction Feasibility\\Assessment};
\node[figfourwork] (works-8) at (1378.0,753) {\mbox{Where2Act~\citep{where2act}}, \mbox{SayCan~\citep{ahn2022saycan}}\\\mbox{Learning Object Affordances~\citep{montesano2008affordances}}};
\end{tikzpicture}%
}
\caption{\textbf{A taxonomy of grounding in embodied perception.} Semantic grounding identifies the entities and functions referred to by a task, spatial grounding estimates geometry and supports reasoning about locations and relations, and physical grounding estimates relevant physical conditions, predicts interaction outcomes, and assesses action feasibility.}
\label{fig:perception-grounding-taxonomy}
\end{figure*}

%% file: sec/4_planning.tex
\label{sec:cognitive-planning}

\begin{figure}[t]
    \centering
    \includegraphics[width=\linewidth]{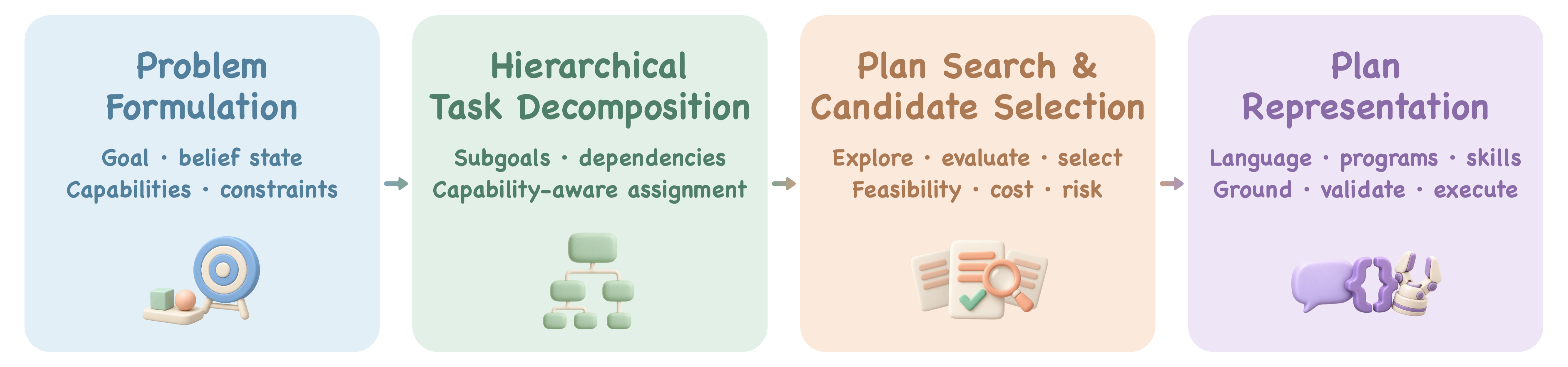}
    \caption{\textbf{Conceptual organization of cognitive reasoning and long-horizon embodied planning.} Problem formulation establishes the goal, belief state, available capabilities, and constraints. Hierarchical decomposition constructs candidate subgoals and their dependencies; search compares alternative branches and anticipated outcomes; plan representation grounds the selected plan in skill interfaces, programs, or constraints for validation and execution. Solid arrows indicate the main information flow, while dashed arrows indicate revision and replanning driven by constraint checks or observed outcomes. These functions may be interleaved in practical systems.}
    \label{fig:planning_overview}
\end{figure}

Embodied reasoning and planning enable agents to organize their interactions with the physical world in pursuit of high-level goals. They connect grounded understanding of the environment with purposeful action by determining what should be achieved, how a task should be structured, and which course of action should be pursued. This section reviews how embodied agents formulate planning problems, construct and evaluate plans, and express their decisions in forms that support physical execution.
As illustrated in Figure~\ref{fig:planning_overview}, this chapter is organized around four interconnected components. Section~\ref{sec:lh-embodied-problem} establishes the goals, state information, available capabilities, and constraints that define long-horizon embodied planning. Building on this formulation, Section~\ref{sec:hierarchical-decomposition} examines how high-level goals are structured into subgoals and their dependencies. Section~\ref{sec:plan-search} reviews how planners explore alternative courses of action, evaluate their anticipated consequences, and select among candidates based on task progress, feasibility, cost, and risk. Section~\ref{sec:plan-representation}  discusses how planning decisions are expressed through language, programs, constraints, or callable skills to support grounding, validation, and execution. Together, these components connect task intent to execution.

\subsection{Problem Formulation}
\label{sec:lh-embodied-problem}

Embodied planning determines a sequence of physically executable interactions to achieve a high-level task goal. For long-horizon tasks, a plan is typically abstracted into a progression of grounded skill invocations:
\begin{equation}
\Pi = \big[(g_1,\sigma_1,\theta_1),\ldots,(g_n,\sigma_n,\theta_n)\big],
\label{eq:hierarchical-plan}
\end{equation}
where $g_i$ specifies an intermediate goal, $\sigma_i \in \mathcal{S}$ is a selected skill or policy, and $\theta_i$ contains its arguments and grounding information.
Crucially, decomposing an instruction into locally valid subgoals does not guarantee task success. Omitted preconditions, conflicting dependencies, and continuous execution constraints can render discrete sequences physically infeasible. Addressing this gap requires task-and-motion planning (TAMP)~\citep{garrett2021integrated} principles to jointly reason over discrete subgoals and continuous physical reachability under uncertainty. 
To construct a coherent, resource-aware, and executable course of action, this chapter is organized around three complementary dimensions: \emph{hierarchical task decomposition} to establish subgoals and their causal orderings; \emph{plan search and candidate selection} to explore and evaluate feasible alternatives; and \emph{plan representation} to express these decisions in formats suitable for execution.

\subsection{Hierarchical Task Decomposition}
\label{sec:hierarchical-decomposition}

Hierarchical task decomposition connects an overall objective to intermediate units of reasoning and execution. Its organization reflects two complementary considerations: the structure of the task itself and the capabilities of the components that carry it out. The first determines which intermediate outcomes contribute to completion and how they depend on one another; the second determines how much work can be assigned to an executor without further subdivision. We examine these considerations first in general LLM-based systems and then in embodied systems, where task structure and execution capability are additionally connected through the physical state of the world.

\textbf{Decomposition in LLM Systems}
In textual domains, decomposition reduces the scope of individual reasoning problems while retaining their contribution to the overall objective. Its value lies in exposing intermediate results that later steps can use and in making the relationships between those results explicit. Sequential subproblems are appropriate when each answer supports the next, whereas nested subtasks allow an intermediate objective to be elaborated at a finer level. Least-to-Most Prompting~\citep{zhou2022least, wang2023plan} exemplifies the broader principle of organizing intermediate reasoning before completing the overall solution. The central issue is whether the chosen structure preserves the dependencies needed for completion without introducing unnecessary intermediate steps.
From the perspective of execution, a task has no universally appropriate decomposition depth. A subproblem that one component can solve directly may require several operations for another, and a useful division of work should reflect this difference. Specialization therefore affects both where subtask boundaries are placed and how their outputs are combined. Decomposed Prompting~\cite{khot2022decomposed} and HuggingGPT~\cite{shen2023hugginggpt} illustrate decomposition around specialized computational components, while ADaPT~\cite{prasad2024adapt} illustrates refinement according to executor outcomes. Modular delegation and adaptive refinement address different aspects of executor compatibility: the former allocates responsibility, while the latter adjusts the amount of work assigned as a unit. They can coexist within the same hierarchy.

\textbf{Decomposition in Embodied Systems.}
In embodied systems, task decomposition must connect the intended outcome to intermediate objectives that reflect the current environment. The same instruction may require different subtasks depending on object locations, existing relations, and the conditions under which actions can be performed. These conditions determine which preparatory steps are necessary and how their completion supports subsequent objectives. 
Unlike textual prompts, embodied decomposition must reflect physical preconditions, reachability, and scene geometry to ground intermediate goals~\citep{song2023llmplanner,wu2024mldt,rana2023sayplan,liu2025delta}.
At the execution level, setting subgoal granularity requires matching task demands to policy capabilities: overly coarse goals overwhelm low-level controllers, while overly granular splits introduce coordination overhead~\citep{ajay2023hip,shi2025hi,geminirobotics15}. Crucially, subtask boundaries must ensure that the termination state of one skill satisfies the initiation preconditions of the next~\citep{huang2026roboharness}.
With multiple robots, the same relationship extends to responsibility allocation and coordination. 
The task structure therefore accommodates both goal dependencies and the conditions of joint execution. Collaborative and hierarchical systems~\cite{mandi2024roco, zeng2026emboteam, su2026imr} provide examples of this broader setting. Their relevance here is the coupling of task division, executor assignment, and coordination, which extends the executor-matching problem from individual capabilities to team organization.


\subsection{Plan Search and Candidate Selection}
\label{sec:plan-search}


Because a task decomposition admits multiple pathways to its subgoals, \emph{search} explores the branching space of alternatives, while \emph{candidate selection} determines which trajectories to retain, refine, or commit to. How an agent allocates its computational budget across these alternatives marks a fundamental transition from digital to physical domains.

\textbf{Search in LLM Systems.}
Exploring alternatives requires deciding where to introduce variation and when to evaluate it. Complete-solution sampling generates multiple attempts for subsequent selection, whereas intermediate branching compares different continuations before a reasoning trajectory is finished. Frameworks like AlphaCode~\citep{li2022alphacode} generate and filter full candidates, whereas Tree of Thoughts~\citep{yao2023tree} evaluates intermediate reasoning trajectories before final completion. 
The central choice concerns how computation is distributed across alternatives: exploring additional attempts, extending a promising direction, or revisiting an earlier decision. Intermediate evaluation can redirect this effort, but requires estimating the value of a candidate before its eventual outcome is known.
Candidate selection depends on the evidence available for making these judgments. Model-generated assessments can estimate a candidate's promise, while execution results~\cite{chen2022codet, li2025codetree} reveal whether particular behavioral requirements are satisfied. These mechanisms connect evaluation to different decisions within search. Their reliability remains bounded by the coverage and correctness of the tests and by the quality of model judgments. Passing available tests supports a candidate without establishing that it satisfies the full task specification.
For interactive agents, evaluation also shapes the actions through which a solution is developed. Gathering information, modifying an artifact, and testing a hypothesis can change which continuations are worth pursuing. LATS~\cite{zhou2023lats} incorporates environment feedback and reflection into search across reasoning and interactive tasks, while SWE-Search~\cite{antoniades2025swe} explores repository-level software-engineering actions using execution information and model-based evaluation. Feedback therefore influences both the preferred candidate and the direction of further exploration. 
However, exploration in software domains benefits from cheap checkpointing and reversible trials, while recovery becomes expensive in the physical world.

\textbf{Search in Embodied Systems.}
Embodied search must evaluate whether a candidate advances the task, is currently executable, and leaves suitable conditions for subsequent actions. These considerations connect immediate selection to longer-horizon completion: an individually feasible action may still lead to an unsuccessful continuation. Skill-level selection in SayCan~\cite{ahn2022saycan} and sequence-level feasibility planning in Text2Motion~\cite{lin2023text2motion} illustrate these different scopes of evaluation.
Accounting for downstream consequences motivates predicting or simulating alternatives before execution. Prospective outcomes allow planners to compare possible continuations and revise choices before committing to physical interaction, as illustrated by VLP~\cite{du2023vlp} and Reflective Planning~\cite{feng2025reflective}. This comparison remains provisional because actual outcomes may differ from predictions. New observations must therefore inform subsequent decisions. Receding-horizon planning, exemplified by ParallelWorld~\cite{chen2026parallelworld}, connects prospective evaluation with execution feedback, allowing the preferred continuation to change as the environment evolves.

\subsection{Plan Representation}
\label{sec:plan-representation}

Plan representation defines the interface between high-level reasoning and robot execution. We distinguish three complementary families: language-based plans, programmatic or constraint-based representations, and skill- or policy-based plans. 

\begin{table}[t]
\centering
\caption{\textbf{Plan representations and their execution interfaces.}}

\label{tab:plan-representations}
\small
\begin{tabularx}{\linewidth}{>{\raggedright\arraybackslash}p{0.15\linewidth} >{\raggedright\arraybackslash}X >{\raggedright\arraybackslash}X >{\raggedright\arraybackslash}p{0.20\linewidth}}
\toprule
\textbf{Family} & \textbf{Variant} & \textbf{Plan interface} & \textbf{Representative work} \\
\midrule
\rowcolor{AEIrow}
Language-based & Language plans & Subgoals and action descriptions & LLM-Planner \cite{song2023llmplanner} \\
\rowcolor{AEIrow}
 & Symbolic compilation & Language--PDDL translation & LLM+P \cite{liu2023llmp} \\

Code-based & Program composition & Control flow, calls, checks & CaP \cite{loop_liang2022codepolicies} \\
 & Solver-based specifications & Task and geometric constraints & AutoTAMP \cite{chen2024autotamp} \\
 
\rowcolor{AEIrow}
Skill-based & Skill-library selection & Skill choice and grounding & SayCan \cite{ahn2022saycan} \\
\rowcolor{AEIrow}
 & Frozen VLA tools & VLA and analytic primitive calls & Harness VLA \cite{zhang2026harness} \\
 \rowcolor{AEIrow}
 & Heterogeneous policies & Capability routing and handoffs & RoboHarness \cite{huang2026roboharness} \\
\bottomrule
\end{tabularx}
\end{table}

Language plans express goals and subgoals in semantically meaningful terms, making them flexible to revise as observations change. LLM-Planner~\citep{song2023llmplanner} uses environment-grounded plans, while Inner Monologue~\citep{loop_huang2022innermonologue} incorporates execution feedback into language-mediated reasoning. Their flexibility, however, leaves object grounding, action interpretation, and outcome verification to downstream components. Language can also be compiled into a formal planning domain. LLM+P~\citep{liu2023llmp} translates task descriptions into PDDL under supplied action preconditions and effects. The resulting plan is checkable only to the extent that both the translation and the domain model match the physical environment.

Programs make execution structure explicit through variables, function calls, sequencing, and conditional control. Code as Policies~\citep{loop_liang2022codepolicies} and ProgPrompt~\citep{singh2022progprompt} use this structure to compose perception and action capabilities. A related but distinct approach specifies goals or constraints and delegates action synthesis to a planner or optimizer~\citep{chen2024autotamp, huang2023voxposer, huang2024rekep}. Programs prescribe how capabilities are orchestrated, while formal specifications describe conditions that the resulting behavior should satisfy. 

The plan selects callable capabilities and supplies their targets or arguments, while each skill or policy generates actions over multiple control steps. SayCan~\citep{ahn2022saycan} grounds language plans in available skills. Harness VLA~\citep{zhang2026harness} and RoboHarness~\citep{huang2026roboharness} illustrate composing learned policies with other robot capabilities. Reliable composition requires more than skill names: the planner needs information about applicability, expected effects, and termination or failure signals. For learned policies, such information may come from value estimates and execution experience rather than fixed symbolic rules.

As shown in Table~\ref{tab:plan-representations}, we summarize the three families described above. These representations can be compared by how faithfully they preserve task intent, expose execution conditions, and return evidence about outcomes. Decomposition proposes subgoals, search selects and orders them, and representation connects the selected plan to executable capabilities. Execution then supplies observations for verification and replanning, closing the planning loop.

%% file: sec/5_action.tex
Action execution serves as a critical bridge between high-level decision-making and physical robot interaction in agentic embodied systems. While high-level reasoning and planning determine task objectives and intermediate subgoals, action execution translates these decisions into concrete robot control commands. Unlike high-level planning, action execution requires continuously adapting robot actions according to sensory feedback and the current robot state. Recent advances in robot policy learning, ranging from visuomotor
policies~\cite{chi2025diffusion,loop_brohan2022rt1,loop_octo2024}
to large-scale vision-language-action (VLA)
models~\cite{loop_brohan2023rt2,kim2024openvla,black2024pi_0},
have enabled increasingly generalizable mappings from sensory
observations and task conditions to executable robot actions. These policies typically operate over short temporal horizons and can be formulated as:
\begin{equation}
    \pi_{\theta}:
    (\mathbf{o}_{t-k:t}, \mathbf{g}_t)
    \rightarrow
    \mathbf{a}_{t:t+H-1},
\end{equation}
where $\mathbf{o}_{t-k:t}$ denotes recent multimodal observations, $\mathbf{g}_t$ the task instruction or subgoal, and $\mathbf{a}_{t:t+H-1}$ a sequence of future actions. This formulation supports diverse action representations and generation paradigms, including autoregressive prediction, continuous regression, diffusion, and flow matching. Beyond action prediction, effective execution requires translating generated actions into low-level control commands to ensure reliable physical interaction. In the following, we review four key aspects of action execution: observation and task conditioning, action representation, action generation, and control execution.

\input{tab/5_representation}
\subsection{Observation and Task Conditioning}
\label{sec:observation_conditioning}

Short-horizon action policies leverage multimodal observations and task conditions to generate physically feasible actions aligned with task objectives. These inputs can be broadly categorized into two types: \textit{Observations}, which capture the surrounding environment and robot states, and \textit{Task Conditions}, which specify the intended behavior.

\textbf{Observation Modalities.}
Visual observations, particularly RGB images, serve as the primary sensory input for modern robot policies, capturing scene appearance, object configurations, and spatial relationships~\cite{loop_brohan2022rt1,kim2024openvla}. For geometry-aware manipulation, some approaches incorporate explicit 3D information through depth maps, point clouds, or voxel representations~\cite{peract,3dvla,spatialvla}. Proprioceptive observations, including joint positions, velocities, end-effector poses, and gripper states, provide information about the robot's current configuration and motion, and are widely incorporated into policies for action prediction~\cite{zhao2023learning,chi2025diffusion,loop_octo2024,black2024pi_0,nvidia2025groot}. Force and tactile observations provide direct feedback on physical contacts and interaction dynamics that are difficult to infer from vision alone, enabling more precise and adaptive action generation in contact-rich manipulation~\cite{fuse,forcevla,vlatouch,tavla}.

\textbf{Task Conditioning.}
Task conditions specify the desired behavior of robot policies and can take various forms, including natural-language instructions, goal images, and intermediate subgoals. \textit{Language conditioning} has become widely adopted in VLA models, enabling policies to interpret semantic instructions and generalize across diverse manipulation tasks~\cite{loop_brohan2022rt1,loop_brohan2023rt2,kim2024openvla,black2024pi_0,nvidia2025groot}. \textit{Goal-image conditioning} provides visual representations of desired task outcomes, either as direct policy inputs~\cite{loop_octo2024} or as generated goal states that guide downstream manipulation~\cite{chen2026goalvla}. Subgoal conditioning is commonly employed in hierarchical frameworks, where high-level planners decompose complex tasks into intermediate objectives or executable skills that guide low-level action policies~\cite{ahn2022saycan,huang2023voxposer,zawalski2024ecot}. Table~\ref{tab:observation_conditioning} summarizes the observation modalities and task conditions adopted by representative robot policies.

After selecting appropriate observations and task conditions, action policies integrate these heterogeneous inputs into a unified representation for action generation. Transformer-based architectures have become a widely adopted solution, leveraging modality-specific embeddings and attention mechanisms to capture cross-modal dependencies~\cite{loop_octo2024,black2024pi_0,nvidia2025groot}. Through such interactions, task information can guide the extraction of relevant sensory features, while robot proprioceptive state provides the physical context for action prediction. In this way, multimodal fusion bridges sensory perception and task objectives, providing a unified conditioning representation for subsequent action generation.

\subsection{Action Representation}

Action representation defines how robot motions are parameterized and organized for policy prediction. Existing short-horizon action policies differ in their control spaces, reference conventions, and temporal horizons, each affecting the flexibility, generalization, and execution efficiency of learned policies.

\textbf{Action Space.}
Robot actions are commonly represented in joint space or task space, depending on the control interface. \textit{Joint-space actions} specify joint positions, velocities, or torques, enabling explicit control of robot configurations and coordinated multi-joint motions, as demonstrated by ACT~\cite{zhao2023learning} and Mobile ALOHA~\cite{loop_fu2024mobilealoha}. While closely aligned with robot actuation, these representations are often embodiment-specific due to differences in joint structures and degrees of freedom. In contrast, \textit{task-space actions} describe end-effector poses, Cartesian displacements, or velocities, typically accompanied by gripper commands, as adopted by RT-1, RT-2, RT-X, and OpenVLA~\cite{loop_brohan2022rt1,loop_brohan2023rt2,o2024open,kim2024openvla}. Such representations provide a more transferable geometric interface across robot embodiments, but generally require inverse kinematics or low-level controllers to satisfy joint limits, reachability, and collision constraints. To accommodate heterogeneous robots and control requirements, recent generalist policies, such as Octo~\cite{loop_octo2024} and RDT-1B~\cite{liu2025rdt1b}, further support diverse action spaces, including joint-space and end-effector-based representations.

\textbf{Absolute vs. Relative Actions.}
Beyond the control space, robot actions can be represented as absolute targets or relative changes. \textit{Absolute actions} specify desired joint configurations or end-effector poses in a predefined reference frame, providing explicit control targets without accumulating errors from successive action integration~\cite{zhao2023learning,chi2025diffusion}. However, predicting absolute targets may require learning complex mappings from observations to global robot configurations. In contrast, \textit{relative actions} predict incremental joint displacements or Cartesian motions with respect to a reference state, facilitating local motion adjustments and closed-loop control~\cite{loop_brohan2022rt1,kim2024openvla,loop_octo2024}. While relative representations can simplify action prediction and adapt to varying initial configurations, accumulated prediction or tracking errors may cause drift during extended execution. Recent empirical studies further highlight the trade-offs between absolute and relative representations in terms of policy learnability and control stability~\cite{feng2026actionspace}. Ultimately, the choice depends on task precision, execution horizon, and the underlying control interface.

\textbf{Single-Step Prediction vs. Action Chunking.}
Beyond spatial parameterization, robot actions also differ in their temporal horizons. \textit{Single-step policies}, such as RT-1, RT-2, and OpenVLA~\cite{loop_brohan2022rt1,loop_brohan2023rt2,kim2024openvla}, predict an individual action $\mathbf{a}_t$ at each timestep, enabling frequent incorporation of new observations and rapid adaptation to environmental changes. However, repeated policy inference introduces computational overhead, while independently predicted actions may lead to temporally inconsistent motions. In contrast, \textit{action-chunking policies} jointly predict a sequence of future actions $\mathbf{a}_{t:t+H}$, as adopted by ACT~\cite{zhao2023learning}, Diffusion Policy~\cite{chi2025diffusion}, Octo~\cite{loop_octo2024}, $\pi_0$~\cite{black2024pi_0}, and GR00T N1~\cite{nvidia2025groot}. By modeling temporal dependencies across consecutive actions, chunking facilitates smoother and more coordinated motions while reducing inference frequency when multiple actions are executed per prediction. Nevertheless, executing longer action chunks without intermediate feedback can limit responsiveness and increase sensitivity to unexpected disturbances. In practice, the prediction and execution horizons can be decoupled through receding-horizon control, allowing policies to balance temporal consistency, computational efficiency, and closed-loop adaptability.

\subsection{Action Generation}
\label{sec:action_generation}

Given multimodal observations and task conditions, action generation models learn to predict executable robot actions according to the chosen action representation. Existing approaches differ primarily in how they model the conditional action distribution, ranging from discrete token prediction and direct continuous regression to generative diffusion and flow-matching models. These approaches exhibit different trade-offs in action expressiveness, computational efficiency, and the ability to model multimodal action distributions. Table~\ref{tab:action_generation} summarizes the representative methods, generation mechanisms, and key advantages and limitations of these paradigms.

\textbf{Discrete Action Generation.}
Discrete action generation converts continuous robot actions into discrete tokens, enabling action prediction through transformer-based sequence modeling. Early approaches, such as Gato~\cite{reed2022gato} and RT-1~\cite{loop_brohan2022rt1}, demonstrate this paradigm by discretizing action dimensions into predefined bins. Building on pretrained vision-language models, RT-2~\cite{loop_brohan2023rt2} and OpenVLA~\cite{kim2024openvla} further formulate action generation as autoregressive token prediction, unifying language understanding and robot control within a shared framework. However, conventional discretization can introduce quantization errors, while autoregressive decoding incurs increasing computational costs for high-dimensional actions and long action sequences. To overcome these limitations, SpatialVLA~\cite{spatialvla} introduces adaptive spatial action grids to improve the representation of 3D motions, whereas FAST~\cite{pertsch2025fast} employs frequency-based tokenization to compress continuous action sequences into compact discrete representations, reducing autoregressive decoding overhead.

\textbf{Continuous Action Generation.}
Direct continuous-action policies predict real-valued robot actions without discrete tokenization or iterative sampling. Deterministic regression approaches typically optimize supervised objectives such as $\ell_1$ or $\ell_2$ losses. Early methods, such as BC-Z~\cite{jang2022bcz}, learn continuous control commands through behavior cloning, while GR-1~\cite{wu2024gr1} and RoboFlamingo~\cite{li2024roboflamingo} incorporate pretrained visual or vision-language representations for continuous action prediction. OpenVLA-OFT~\cite{kim2025fine} further enables parallel action regression, improving inference efficiency compared with autoregressive token decoding. However, when multiple valid actions exist for the same observation, deterministic regression may average different behaviors, potentially producing ineffective motions. To address this limitation, latent-variable approaches, such as Play-LMP~\cite{lynch2020latent}, HULC~\cite{mees2022hulc}, and ACT~\cite{zhao2023learning}, introduce learned latent representations to capture variations in demonstrated behaviors. Overall, these methods enable efficient, quantization-free action prediction without iterative sampling, but their ability to capture diverse valid behaviors remains limited, particularly for deterministic regression-based action policies.

\textbf{Diffusion-Based Action Generation.}
Diffusion-based policies model conditional action distributions by progressively denoising randomly initialized action sequences into continuous robot trajectories. Given an action sequence $\mathbf{a}$ and conditioning information $\mathbf{c}$, the model is typically trained to predict the noise added at different diffusion steps, using the following objective:
\begin{equation}
    \mathcal{L}_{\mathrm{diff}}
    =
    \mathbb{E}_{\mathbf{a},\boldsymbol{\epsilon},s}
    \left[
    \left\|
    \boldsymbol{\epsilon}
    -
    \boldsymbol{\epsilon}_{\theta}
    (\mathbf{x}_s,s,\mathbf{c})
    \right\|_2^2
    \right],
\end{equation}
where $\mathbf{x}_s=\sqrt{\bar{\alpha}_s}\mathbf{a}+\sqrt{1-\bar{\alpha}_s}\boldsymbol{\epsilon}$ denotes the noisy action at diffusion step $s$, and $\boldsymbol{\epsilon}\sim\mathcal{N}(\mathbf{0},\mathbf{I})$. Unlike deterministic regression, diffusion models can capture multiple valid action modes and jointly generate temporally coordinated action sequences, making them suitable for complex manipulation tasks~\cite{chi2025diffusion}. Building on Diffusion Policy~\cite{chi2025diffusion}, DP3~\cite{ze2024dp3} incorporates 3D geometric representations to improve visuomotor generalization, while MDT~\cite{reuss2024mdt} extends diffusion transformers to multimodal goal conditioning. At a larger scale, Octo~\cite{loop_octo2024} enables generalist action generation across heterogeneous robot datasets, and RDT-1B~\cite{liu2025rdt1b} scales diffusion transformers for bimanual manipulation with unified action representations. Despite their expressiveness, diffusion policies typically require multiple denoising steps, introducing computational overhead and a trade-off between sampling quality and inference latency.

\textbf{Flow-Matching Action Generation.}
Closely related to diffusion models, flow matching provides an alternative generative framework that learns a time-dependent velocity field to transform random noise into continuous action trajectories. Given an action sequence $\mathbf{a}$ and conditioning information $\mathbf{c}$, a common conditional flow-matching objective is defined as:
\begin{equation}
    \mathcal{L}_{\mathrm{FM}}
    =
    \mathbb{E}_{\mathbf{a},\boldsymbol{\epsilon},s}
    \left[
    \left\|
    \mathbf{v}_{\theta}
    (\mathbf{x}_s,s,\mathbf{c})
    -
    (\mathbf{a}-\boldsymbol{\epsilon})
    \right\|_2^2
    \right],
\end{equation}
where $\mathbf{x}_s=(1-s)\boldsymbol{\epsilon}+s\mathbf{a}$ linearly interpolates between Gaussian noise $\boldsymbol{\epsilon}\sim\mathcal{N}(\mathbf{0},\mathbf{I})$ and target actions $\mathbf{a}$, with $s\in[0,1]$. During inference, action trajectories are generated by numerically integrating the learned velocity field from noise to the target action distribution. This paradigm has been adopted by generalist VLA models, including $\pi_0$~\cite{black2024pi_0} and $\pi_{0.5}$~\cite{black2025pi05}, which integrate flow-matching action experts with pretrained vision-language representations. GR00T N1~\cite{nvidia2025groot} extends this framework to humanoid manipulation, while SmolVLA~\cite{shukor2025smolvla} and X-VLA~\cite{zheng2025xvla} emphasize efficient deployment and cross-embodiment generalization, respectively. Like diffusion models, flow matching can capture complex, multimodal action distributions, while potentially enabling efficient generation with fewer integration steps. However, inference still requires numerical integration, introducing a trade-off between action quality and computational efficiency that depends on the integration scheme and model architecture.
\input{tab/5_generation}

\subsection{Control Execution}
\label{sec:control_execution}

Beyond action generation, reliable robot manipulation requires translating predicted actions into physical execution while maintaining responsiveness to environmental changes. Closed-loop execution continuously updates actions based on new observations, while multi-rate control coordinates the different operating frequencies of learned policies and robot controllers. Asynchronous execution further reduces inference-induced delays by overlapping action prediction with ongoing robot motion, whereas low-level motor controllers convert predicted actions into executable robot commands. Together, these mechanisms bridge learned action policies and real-world robot execution, enabling responsive and stable manipulation.

\noindent\textbf{Closed-Loop Action Execution.}
Closed-loop execution enables robot policies to continuously adapt their actions based on new observations, correcting deviations caused by prediction errors or environmental disturbances. A common strategy is receding-horizon control, which executes only part of a predicted action chunk before replanning with updated observations~\cite{chi2025diffusion,loop_octo2024}. This strategy balances temporal consistency with responsiveness to environmental changes. However, frequent replanning may produce inconsistent predictions across successive action chunks. To address this issue, ACT~\cite{zhao2023learning} introduces temporal ensembling, which averages overlapping action predictions from successive chunks to improve motion smoothness.

\noindent\textbf{Multi-Rate Control.}
Real-world robot systems operate at multiple temporal frequencies, with learned policies typically running at lower rates than low-level motor controllers~\cite{loop_brohan2022rt1,chi2025diffusion,black2024pi_0}. This mismatch is particularly pronounced in large VLA models, where inference latency may exceed the robot control period~\cite{kim2024openvla,kim2025fine}. To bridge this gap, action-chunking policies, such as $\pi_0$~\cite{black2024pi_0} and OpenVLA-OFT~\cite{kim2025fine}, generate multiple future actions in a single inference step, allowing commands to be buffered and executed at higher frequencies. When necessary, interpolation or resampling further aligns predicted actions with the low-level control frequency. By decoupling policy inference from high-frequency motor control, multi-rate execution enables computationally expensive policies to provide continuous control references without requiring inference at every control step.

\noindent\textbf{Asynchronous Action Execution.}
Despite action chunking, synchronous execution may introduce pauses when the robot must wait for the next action prediction, particularly for computationally expensive VLA models. To address this limitation, asynchronous execution overlaps policy inference with ongoing robot motion, allowing subsequent action chunks to be generated before the current ones are exhausted. SmolVLA~\cite{shukor2025smolvla} implements this paradigm through asynchronous inference and action buffering to maintain continuous robot execution. However, independently generated action chunks may introduce abrupt motion transitions, while inference delays can cause discrepancies between the observed and actual robot states. To improve execution continuity, Real-Time Chunking (RTC)~\cite{black2025rtc} conditions new action predictions on previously committed commands through action inpainting. FutureRTC~\cite{jiang2026futurertc} further addresses temporal misalignment by predicting execution-time observations and robot states. Together, these approaches improve the consistency of asynchronous execution across successive action chunks and changing robot states.

%% file: tab/5_representation.tex
\begin{table}[t]
    \centering
    \small
    \renewcommand{\arraystretch}{1.25}
    \setlength{\tabcolsep}{3pt}

    \caption{\textbf{Observation modalities and task conditioning in representative robot manipulation policies.}}
    \label{tab:observation_conditioning}
    \resizebox{\textwidth}{!}{
    \begin{tabular}{lcccccccc}
        \toprule
        \multirow{2}{*}{\textbf{Method}}
        & \multicolumn{5}{c}{\textbf{Observation Modalities}}
        & \multicolumn{3}{c}{\textbf{Task Conditioning}} \\
        \cmidrule(lr){2-6}
        \cmidrule(lr){7-9}
        & \textbf{RGB}
        & \textbf{3D Geometry}
        & \textbf{Proprioception}
        & \textbf{Force/Tactile}
        & \textbf{Audio}
        & \textbf{Language}
        & \textbf{Goal Image}
        & \textbf{Subgoal} \\
        \midrule

        \rowcolor{AEIrow}
        PerAct~\cite{peract}
        & $\checkmark$ & $\checkmark$ & $\checkmark$ & --
        & -- & $\checkmark$ & -- & -- \\

        OpenVLA~\cite{kim2024openvla}
        & $\checkmark$ & -- & -- & --
        & -- & $\checkmark$ & -- & -- \\

        \rowcolor{AEIrow}
        Octo~\cite{loop_octo2024}
        & $\checkmark$ & -- & $\checkmark$ & --
        & -- & $\checkmark$ & $\checkmark$ & -- \\

        ForceVLA~\cite{forcevla}
        & $\checkmark$ & -- & $\checkmark$ & $\checkmark$
        & -- & $\checkmark$ & -- & -- \\

        \rowcolor{AEIrow}
        FuSe~\cite{fuse}
        & $\checkmark$ & -- & $\checkmark$ & $\checkmark$
        & $\checkmark$ & $\checkmark$ & -- & -- \\

        Goal-VLA~\cite{chen2026goalvla}
        & $\checkmark$ & $\checkmark$ & $\checkmark$ & --
        & -- & $\checkmark$ & $\checkmark$ & -- \\

        \rowcolor{AEIrow}
        CoT-VLA~\cite{zhao2025cotvla}
        & $\checkmark$ & -- & -- & --
        & -- & $\checkmark$ & $\checkmark$ & $\checkmark$ \\

        \textbf{$\boldsymbol{\pi}_{0.5}$}~\cite{black2025pi05}
        & $\checkmark$ & -- & $\checkmark$ & --
        & -- & $\checkmark$ & -- & $\checkmark$ \\

        \rowcolor{AEIrow}
        VPP2~\cite{guo2026vpp2}
        & $\checkmark$ & -- & -- & -- & -- & $\checkmark$ & -- & -- \\
        OpenWAM-$\alpha$~\citep{wang2026openwam}
        & $\checkmark$ & -- & $\checkmark$ & -- & -- & $\checkmark$ & -- & -- \\

        \bottomrule
    \end{tabular}
    }
\end{table}

%% file: tab/5_generation.tex
\begin{table*}[t]
    \centering
    \small
    \renewcommand{\arraystretch}{1.2}
    \setlength{\tabcolsep}{3pt}

    \newcolumntype{L}[1]{>{\raggedright\arraybackslash}p{#1}}

    \caption{\textbf{Comparison of action generation paradigms in representative robot policies.}}
    \label{tab:action_generation}

    \resizebox{\textwidth}{!}{
    \begin{tabular}{@{}
        L{3.3cm}
        L{4.5cm}
        L{4.0cm}
        L{4.0cm}
        @{}}
        \toprule
        \textbf{Category}
        & \textbf{Representative Methods}
        & \textbf{Advantages}
        & \textbf{Limitations} \\
        \midrule
        \rowcolor{AEIrow}

        \textbf{Discrete Action Generation}
        & RT-1~\citep{loop_brohan2022rt1},
          RT-2~\citep{loop_brohan2023rt2},
          OpenVLA~\citep{kim2024openvla},
          SpatialVLA~\citep{spatialvla},
        & Unified VLM-action modeling; leverages pretrained language models
        & Quantization errors; sequential decoding overhead \\

        \textbf{Continuous Action Generation}
        & BC-Z~\citep{jang2022bcz},
          GR-1~\citep{wu2024gr1},
          RoboFlamingo~\citep{li2024roboflamingo},
          ACT~\citep{zhao2023learning},
          OpenVLA-OFT~\citep{kim2025fine}
        & Quantization-free actions; efficient and parallel prediction
        & Mode averaging in deterministic regression; latent modeling complexity \\

        \rowcolor{AEIrow}
        
        \textbf{Diffusion-Based Action Generation}
        & Diffusion Policy~\citep{chi2025diffusion},
          DP3~\citep{ze2024dp3},
          MDT~\citep{reuss2024mdt},
          Octo~\citep{loop_octo2024},
          RDT-1B~\citep{liu2025rdt1b}
        & Multimodal action distributions; flexible trajectory generation
        & Multiple denoising steps; inference latency \\

        \textbf{Flow-Matching Action Generation}
        & $\pi_0$~\citep{black2024pi_0},
          $\pi_{0.5}$~\citep{black2025pi05},
          GR00T N1~\citep{nvidia2025groot},
          SmolVLA~\citep{shukor2025smolvla},
          X-VLA~\citep{zheng2025xvla}
        & Multimodal action distributions; potentially efficient ODE sampling
        & Numerical integration overhead; quality--step trade-off \\

        \bottomrule
    \end{tabular}
    }
\end{table*}

%% file: sec/6_evolution.tex

\begin{figure*}[t]
    \centering
    \includegraphics[width=\textwidth]{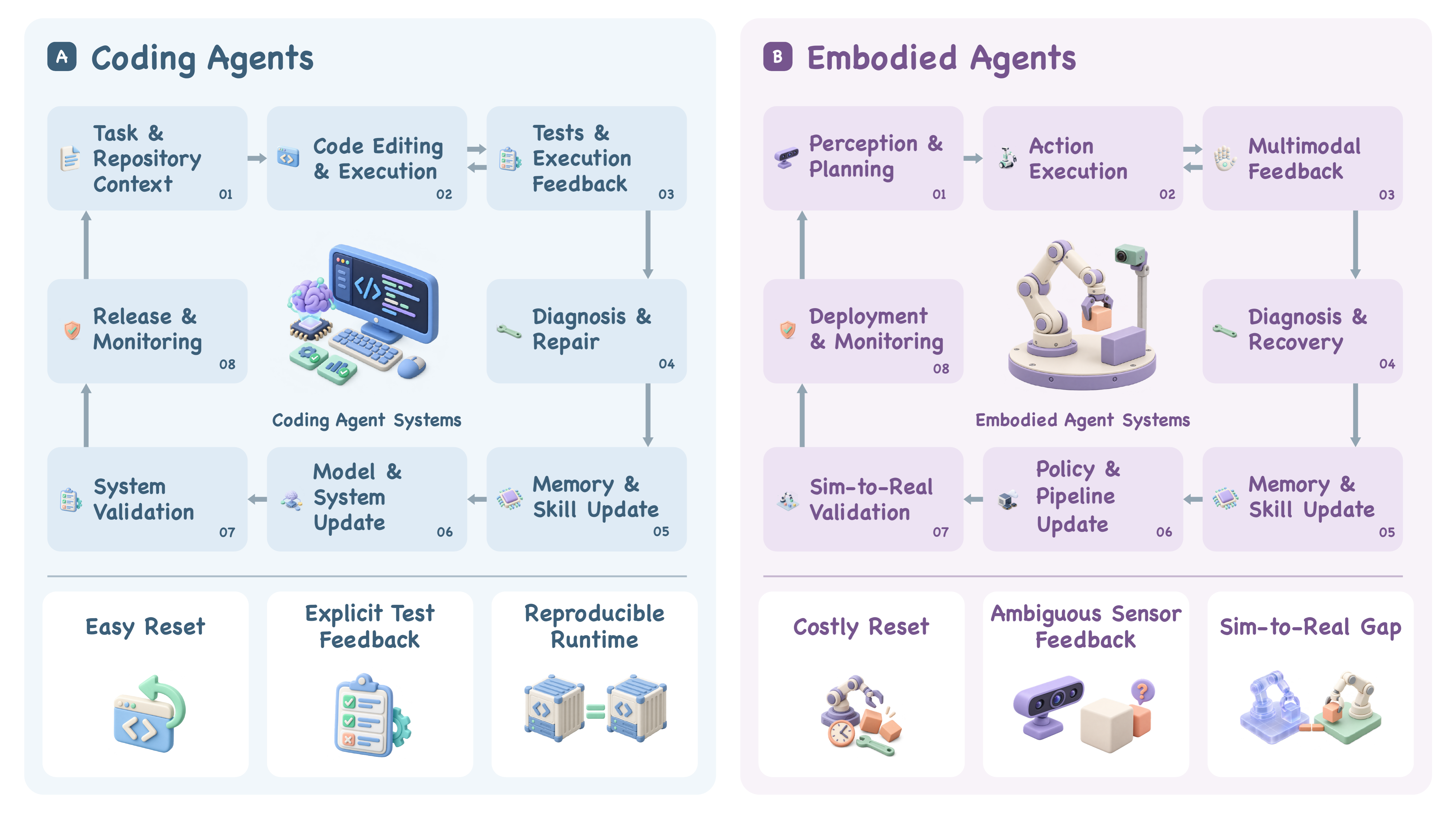}
    \caption{\textbf{Conceptual comparison of evolution loops in coding and embodied agent systems.} (A) Coding agents use execution and test feedback to diagnose failures, refine memory and skills, and inform model and system updates before validation and release. (B) Embodied agents use multimodal feedback to guide recovery and updates to memory, skills, policies, and pipelines, with sim-to-real validation and deployment monitoring. The bottom panels highlight three practical contrasts: reset cost, feedback verifiability, and the gap between validation and deployment. These contrasts describe typical operating conditions; the update stages may involve external orchestration or human oversight.}
    \label{fig:evolution_comparison}
\end{figure*}

The preceding sections decompose embodied intelligence into a perception, planning, and action loop. 
Yet continuous self-improvement cannot be realized within classical open-loop pipelines or monolithic end-to-end policies.
A vital conceptual blueprint for this physical agency emerges from the digital domain. 
Autonomous coding agents have established a mature paradigm of closed-loop iteration.
For embodied agents, this software engineering loop is mapped onto physical reality: sensory anomalies function as physical tracebacks, motor failures trigger in-situ programmatic debugging, and newly acquired manipulation primitives are registered as persistent tools.
As illustrated in Figure~\ref{fig:evolution_comparison}, coding and embodied agents share a feedback-driven evolution loop, but differ in reset cost, feedback verifiability, and the gap between validation and deployment. 
These differences make it harder to establish whether a successful recovery will generalize beyond the current trial. The comparison therefore motivates a distinction between correcting the current execution and validating a persistent update for future tasks.
Accordingly, Section~\ref{sec:loop-scope} explains five possible update targets. Section~\ref{sec:loop-digital} describes how coding agents currently execute, verify, revise, and reuse changes. Section~\ref{sec:loop-embodied} reviews the corresponding mechanisms in embodied systems. Section~\ref{sec:loop-pipeline} identifies missing components, proposes an integrated pipeline, and outlines the main evaluation and research challenges.

\subsection{Evolution Targets}
\label{sec:loop-scope}

We use five terms for the place where a change is made: \emph{control}, \emph{reflection}, \emph{memory}, \emph{models}, and \emph{system}. They describe persistence and evidence requirements rather than a universal maturity ladder:
\begin{enumerate}
    \item The \emph{control} target changes the next action or a short local trajectory. A grasp that is corrected after a measured slip is a control correction; the evidence is recovery within the current episode.
    \item The \emph{reflection} target changes the agent's explanation of the failure or its current plan. It should be counted only when the revised explanation leads to a better intervention under the same retry budget.
    \item The \emph{memory} target stores an experience, fact, or executable skill for later use, so the claim requires provenance and improvement on new instances after transient context has been removed.
    \item The \emph{models} target changes a policy, value function, or world model using new data; it requires held-out gains and checks that earlier abilities are retained.
    \item The \emph{system} target changes the agent's code, tools, curriculum, reward, or update procedure. This is the broadest claim and needs independent comparisons across versions.
\end{enumerate}
Control and reflection normally operate within one episode, while memory, models, and system changes persist across episodes. The agentic contribution is the selection of informative evidence, the choice of an appropriate update target, and the decision to accept or reject the resulting change.

\subsection{Coding Closed-Loop Improvement}
\label{sec:loop-digital}

Coding agents offer a useful reference because repository actions have executable consequences and inexpensive verifiers. The environment is a versioned workspace; observations include tests, compiler traces, logs, and diffs. This makes the loop cheap and reproducible, but does not by itself establish unrestricted self-improvement.

\textbf{Control and Reflection: Execute, Verify, and Replan.}
At the control and reflection targets, a coding agent proposes a bounded change, checks it against executable evidence, and revises it when the result falls short. Benchmarks and agent--computer interfaces make this loop observable through task outcomes, tests, logs, and workspace changes~\cite{loop_jimenez2024swebench,loop_yang2024sweagent,loop_wang2025openhands}. Reflection is useful only when it leads to a verifiable change: self-assessment or repeated sampling alone does not establish correction, while tool feedback and search can help identify better repairs~\cite{loop_yao2023react,loop_madaan2023selfrefine,loop_shinn2023reflexion,loop_gou2024critic,loop_wang2023selfconsistency,loop_huang2024selfcorrect,antoniades2025swe}. Any claim of lasting improvement therefore requires gains on later tasks, not only success on the current repair.

\textbf{Memory: Reuse a Validated Procedure Across Tasks.}
At the memory target, lessons from execution become retrievable assets for later tasks. Agent studies span both natural-language experience and executable skills, showing how stored feedback or procedures can guide subsequent attempts without necessarily changing model weights~\cite{loop_shinn2023reflexion,expel2024,loop_wang2023voyager}. For coding agents, this counts as improvement only if the memory remains useful after transient context is cleared and transfers to new tasks under an applicability check. Because repository changes can invalidate commands and visible tests can encourage overfitting, reusable memories also need provenance and conditions for expiry.

\textbf{Models and Systems: Adapt the Agent or Its Training Process.}
At the models target, interaction data can train or adapt an agent; at the system target, the update changes the agent's code, tools, or workflow. SWE-Gym demonstrates the former by using executable repository tasks to train software-engineering agents and verifiers, while automated agent-design work and the Darwin Gödel Machine explore the latter through benchmark-guided search over agent programs~\cite{swegym2025,hu2024adas,loop_zhang2025dgm}. These results support bounded adaptation under their reported data, benchmarks, and compute budgets, not unrestricted recursive self-improvement.
Coding-agent loops are comparatively mature because snapshots provide cheap resets, diagnostics are informative, specifications are executable, and versions are easy to compare. These are requirements for a credible embodied loop, not evidence that coding agents have solved self-improvement.

\begin{table}[t]
\centering
\small
\renewcommand{\arraystretch}{1.4} 
\caption{\textbf{Representative closed-loop mechanisms in digital and embodied agents.} Each row names the main update target and the capability shared by the two settings.}
\label{tab:coding-embodied-levels}
\begin{tabular}{p{0.08\linewidth}p{0.30\linewidth}p{0.30\linewidth}p{0.22\linewidth}}
\toprule
\textbf{Target} & \textbf{Digital-agent examples} & \textbf{Embodied-agent examples} & \textbf{Shared capability}\\
\midrule
\rowcolor{AEIrow}
Control & SWE-agent \cite{loop_yang2024sweagent}; OpenHands \cite{loop_wang2025openhands}. & SayCan \cite{ahn2022saycan}; Inner Monologue \cite{loop_huang2022innermonologue}; Code as Policies \cite{loop_liang2022codepolicies}. & Feedback-guided action correction.\\
Reflection & ReAct \cite{loop_yao2023react}; Self-Refine \cite{loop_madaan2023selfrefine}; Reflexion \cite{loop_shinn2023reflexion}; CRITIC \cite{loop_gou2024critic}. & REFLECT \cite{loop_liu2023reflect}. & Failure interpretation and replanning.\\
\rowcolor{AEIrow}
Memory & Reflexion \cite{loop_shinn2023reflexion}; Voyager \cite{loop_wang2023voyager}. & Voyager \cite{loop_wang2023voyager}; PhySGA \cite{lyu2026physga} & Reuse of prior experience.\\
Models & SWE-Search \cite{loop_jimenez2024swebench}; Darwin Godel Machine \cite{loop_zhang2025dgm}. & RoboCat \cite{loop_bousmalis2023robocat}; RT-1 \cite{loop_brohan2022rt1}; RT-2 \cite{loop_brohan2023rt2}; Open X-Embodiment \cite{loop_oxe2023}. & Data-driven model adaptation.\\
\rowcolor{AEIrow}
System & Darwin Godel Machine \cite{loop_zhang2025dgm}. & Eureka \cite{loop_ma2024eureka}; GenSim \cite{loop_wang2024gensim}; RoboGen \cite{loop_wang2024robogen}. & Search over update procedures.\\
\bottomrule
\end{tabular}
\end{table}

Table~\ref{tab:coding-embodied-levels} shows the common pattern: an execution signal is interpreted, a targeted change is proposed, and later trials decide whether it should be reused. Coding agents can usually reset to a known snapshot, inspect a precise trace, and compare versions cheaply. Embodied agents have counterparts for all five targets, but occlusion, contact uncertainty, irreversible actions, embodiment differences, and hardware drift make the same evidence much harder to obtain.

\begin{table}[t]
\centering
\small
\setlength{\tabcolsep}{2pt}
\caption{\textbf{Capability coverage in representative software and embodied closed-loop systems.} A checkmark means routine support; a triangle means partial or task-dependent support; a cross means that the capability is generally not established.}
\label{tab:coding-embodied-capabilities}
\begin{tabular}{p{0.40\linewidth}>{\centering\arraybackslash}p{0.10\linewidth}>{\centering\arraybackslash}p{0.10\linewidth}p{0.38\linewidth}}
\toprule
\textbf{Capability} & \textbf{Software} & \textbf{Embodied} & \textbf{Interpretive implication}\\
\midrule
\rowcolor{AEIrow}
Observable execution trace & $\checkmark$ & $\triangle$ & Sensors are incomplete under occlusion and contact.\\
Cheap reset and reproducible state & $\checkmark$ & $\times$ & Physical trials consume time, hardware, and sometimes objects.\\
\rowcolor{AEIrow}
Cheap rollback and state restoration & $\checkmark$ & $\triangle$ & A physical failure can damage the scene or the robot.\\
Failure diagnosis with external checks & $\checkmark$ & $\triangle$ & Physical symptoms often have several plausible causes.\\
\rowcolor{AEIrow}
Persistent memory or skill reuse & $\checkmark$ & $\triangle$ & Transfer needs explicit embodiment and context conditions.\\
Independent held-out validation & $\checkmark$ & $\triangle$ & Simulation and learned evaluators may share the candidate's blind spots.\\
\rowcolor{AEIrow}
Version tracking and independent promotion & $\checkmark$ & $\triangle$ & Calibration and hardware drift complicate safe release.\\
\bottomrule
\end{tabular}
\end{table}

\subsection{Embodied Closed-Loop Evolution}
\label{sec:loop-embodied}

Embodied systems contain counterparts to all five update targets described above, but the evidence and failure modes change when actions affect a physical, partially observed world. The following subsections review current approaches in increasing order of persistence, from within-episode recovery to policy and training-pipeline updates.

\textbf{Control Correction: Verify Actions and Recover.}
At the within-episode tier, the agent adjusts subsequent actions reactively upon sensing execution discrepancies without altering underlying policies. Foundational formulations integrate measured feedback into action selection~\citep{loop_kober2013rl}, coupling language reasoning with affordance feasibility~\citep{ahn2022saycan}, multimodal environmental critique~\citep{loop_huang2022innermonologue}, or executable API generation~\citep{loop_liang2022codepolicies}. These systems improve the connection between semantic intent and executable action, but a successful recovery does not by itself create a new skill.
Verification must combine task predicates with sensor evidence at an appropriate temporal resolution. For a pick-and-place operation, relevant evidence may include object identity before contact, gripper closure and load during lifting, object motion relative to the gripper, and the final placement relation. A final image alone may fail to distinguish a stable placement from an object that is about to fall. Conversely, low-level execution success may conceal a semantic error such as moving the wrong item. The monitor should retain the provenance and confidence of each predicate rather than collapse all feedback into a single textual success judgment.

\textbf{Task-Level Reflection: Diagnose Failure and Replan}
When simple recovery fails, task-level reflection translates multisensory anomalies into causal hypotheses to drive structured replanning. Systems like REFLECT~\citep{loop_liu2023reflect} aggregate hierarchical multimodal execution traces, prompting foundation models to diagnose underlying failure causes within physical benchmarks. This is a direct physical counterpart to tool-supported reflection, but the observations must first be grounded in contact, motion, and scene events.
A diagnostic record should separate symptom, hypothesis, evidence, and proposed test. Without these distinctions, an untested explanation can become a persistent rule. Hierarchical summaries reduce context length but may discard the transient contact event needed to distinguish two causes, so summaries should link back to timestamped raw observations. A language explanation remains a hypothesis unless recovery, intervention, or controlled perturbation supplies evidence for it.

\textbf{Memory and Skills: Reuse Experience Across Episodes.}
To prevent repeated mistakes across episodes, persistent memory turns verified trajectories into reusable procedural skills or contextual knowledge. While digital agents accumulate symbolic programs or textual reflections~\citep{loop_wang2023voyager,loop_shinn2023reflexion}, robot skill libraries~\citep{zhang2026harness, huang2026roboharness, zhang2026vista} demand geometric and hardware grounding. 
A useful embodied memory entry can be represented as
\begin{equation}
  m=(c,\xi,\kappa,\rho,v),
\end{equation}
where $c$ specifies the applicability context, $\xi$ links to supporting experience, $\kappa$ stores a fact or executable skill, $\rho$ records confidence and validation status, and $v$ identifies the source system and embodiment.
Effective retrieval continually resolves conflicting entries, verifies state validity against $c$, and retires stale skills when embodiment morphology or workspace calibration changes.
Memory quality is measured by downstream decisions: whether retrieval helps on new instances, whether irrelevant memories are rejected, and whether removing the memory reverses the claimed gain.

\textbf{Policy Adaptation: Learn from New Interaction Data}
At the policy-adaptation tier, empirical interaction data updates the underlying neural weights of policies and world models. Generalist architectures and large-scale cross-embodiment datasets establish scalable observation-to-action transfer~\citep{loop_brohan2022rt1,loop_brohan2023rt2,loop_driess2023palme,loop_oxe2023,loop_octo2024,loop_khazatsky2024droid}, while autonomous self-training frameworks iteratively expand their operational distributions~\citep{loop_bousmalis2023robocat}. However, repeatedly ingesting self-generated rollouts risks amplifying model bias and coverage gaps. Achieving a true continuous deployment-update loop therefore demands prioritized data selection targeting diagnostic weaknesses, paired with rigorous evaluation on held-out physical distributions to prevent policy collapse.

\textbf{System Evolution: Generate Rewards, Tasks, and Training Data}
The highest evolutionary tier automates the meta-pipeline itself, dynamically generating reward functions, simulation environments, and training curricula rather than adjusting individual policy checkpoints. Frameworks like Eureka~\citep{loop_ma2024eureka}, GenSim~\citep{loop_wang2024gensim}, and RoboGen~\citep{loop_wang2024robogen} orchestrate iterative propose--generate--learn loops that synthesize reward code and task suites from language feedback. Because simulated rewards are susceptible to reward hacking and physics-engine exploitation, generated assets require independent verification for physical fidelity, task diversity, and sim-to-real transferability. Ultimately, scaling this evolutionary tier balances automated simulation synthesis with targeted real-world human demonstrations and high-value teleoperation~\citep{loop_fu2024mobilealoha,loop_mandlekar2021demonstrations,loop_ahn2024autort}.

\subsection{From Component Loops to Embodied Evolution Pipeline}
\label{sec:loop-pipeline}

The preceding review identifies useful components, but most embodied results validate only one part of the outer loop at a time. This section turns those components into a concrete argument: first, what is already supported by evidence; second, what a complete pipeline must record and verify; and third, which open problems prevent that pipeline from operating reliably. 

Embodied systems~\cite{loop_huang2022innermonologue,loop_liu2023reflect,loop_wang2023voyager,loop_bousmalis2023robocat,loop_ma2024eureka,loop_wang2024gensim,loop_wang2024robogen} already provide several foundations for evolution. 
Proprioception, force sensing, geometric constraints, and local controllers expose physical execution. 
Language-conditioned planners can insert recovery actions.
Multimodal failure analysis can connect an outcome to scene and contact evidence. Executable skills and cross-embodiment datasets provide mechanisms for transfer. 
However, compared with coding agents, embodied systems still face five linked gaps: state and outcome observability; action reversibility; failure attribution; skills across embodiments; and independent promotion of a physical update. A robot therefore needs more than a reflection module: it needs a route from multimodal evidence to a narrow change and an independent release decision.

A complete evolution loop can be described as six connected functions:
\begin{enumerate}
  \item \textbf{Record.} Store synchronized observations, commands, measured execution, subgoal checks, interventions, and software, model, and hardware versions. Summaries should link back to timestamped raw evidence.
  \item \textbf{Check.} Combine terminal task checks with process signals and safety monitors. Keep uncertainty explicit and distinguish failure from an invalid or inconclusive evaluation.
  \item \textbf{Diagnose.} Separate candidate causes in perception, state estimation, planning, skill selection, control, and interfaces. If evidence is insufficient, seek new observations or a controlled diagnostic action.
  \item \textbf{Update.} Select the smallest suitable target: invalidate a memory, edit a skill, collect a demonstration, revise a reward or curriculum, or update a policy or world model. Record the supporting evidence and the conditions under which the change is expected to work.
  \item \textbf{Validate.} Use replay for historical evidence, simulation for counterfactuals under stated assumptions, and constrained physical trials for transfer. Shadow deployment can compare decisions without actuation, but it cannot replace a physical test of the action's outcome.
  \item \textbf{Promote.} Register provenance, calibration, models, tools, and evaluation results. Release a candidate only after regression and safety checks, retain a feasible recovery configuration, and feed new failures back into the log.
\end{enumerate}
These six steps connect the components reviewed above into a sequence that records evidence, selects a narrow update, tests it, and provides a way back if the update is unsafe. It is a reference workflow for organizing experiments, rather than a claim that current systems already implement every step.

%% file: sec/8_safety.tex
Agentic embodied systems connect perception, planning, execution, and feedback in a continuing loop. Beyond the action interface of a VLA policy, an agentic system can actively gather missing information, retain safety-relevant state, insert prerequisite actions, request assistance, and revise a plan after an execution error. These capabilities also introduce new failure modes: a planner proposes a risky instruction, a memory may preserve a stale or incorrect fact, a tool call may exceed the user's authority, and a policy may remain unsafe despite a convincing verbal explanation. This chapter therefore treats \emph{Safety and Trustworthiness} as system-level properties of AEI, rather than properties of the language model or VLA in isolation.
Section~\ref{sec:safety-scope} defines the protected entities, hazard types, temporal scope, and autonomy boundaries. 
Section~\ref{sec: safety on functions} analyzes the key points of safety on four functions in the closed-loop evolution.
Section~\ref{sec:safety-evaluation} asks what evidence is sufficient to support a safety claim and identifies the gaps that prevent current systems from providing end-to-end assurance.

\begin{figure}[t]
\centering
\includegraphics[width=\linewidth]{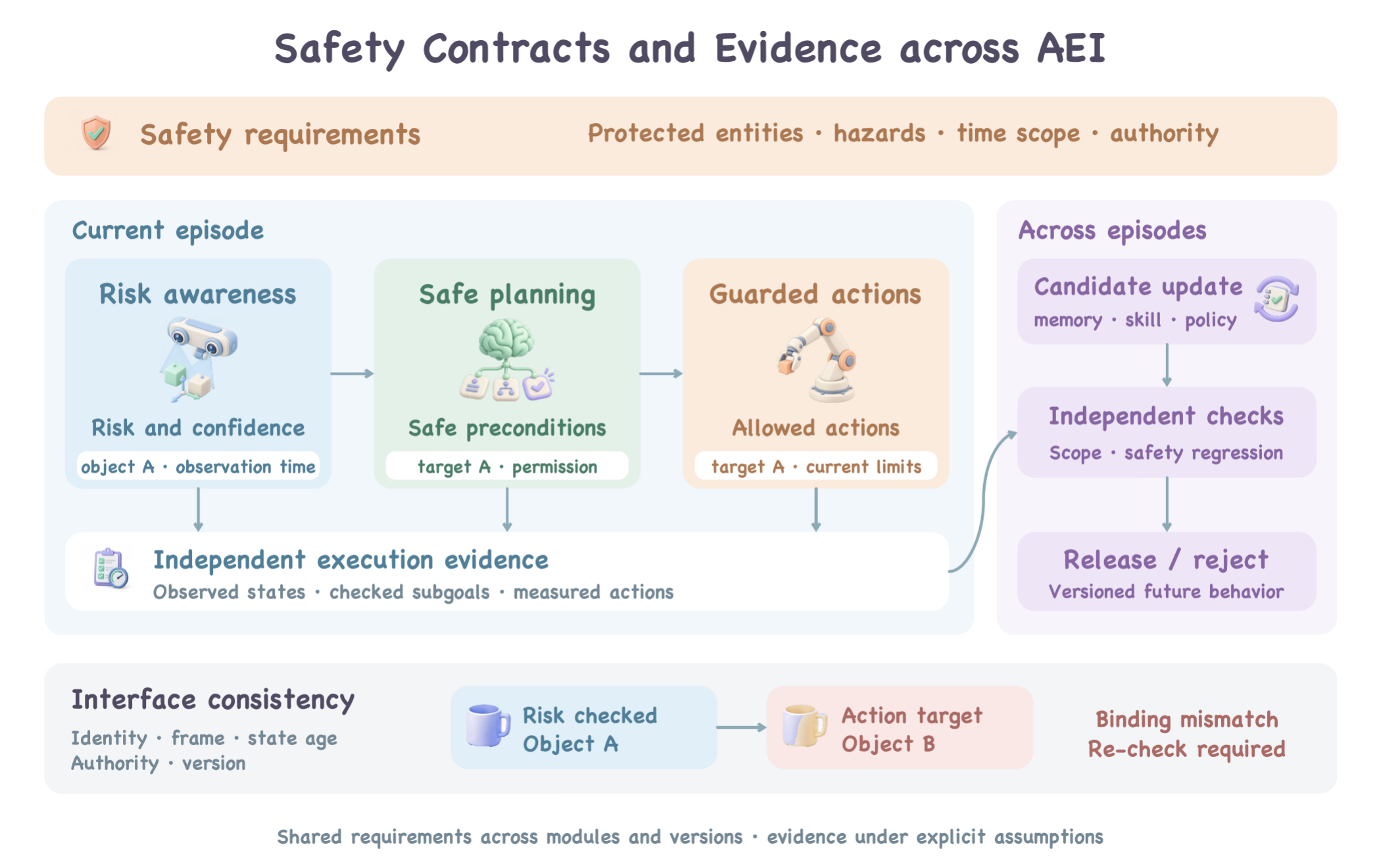}
\caption{\textbf{Safety contracts and evidence across agentic embodied intelligence.} Shared requirements apply within an episode and to persistent updates across episodes. Risk judgments, planned subgoals, and action constraints must preserve their referents and assumptions. Execution evidence supports independent checks before an update is accepted. The lower example shows a broken object binding: a risk check for object A cannot justify an action on object B. Identity, coordinate frame, state age, authority, and version are complementary consistency requirements. This is a conceptual synthesis, not an implemented architecture or an end-to-end safety guarantee.}
\label{fig:safety-contracts}
\end{figure}

\subsection{Scope of Safety}
\label{sec:safety-scope}

For each episode, safety must account for the observations, actions, latent state, and memory that evolve over time. Four overlapping risk situations are especially important:
\begin{enumerate}
  \item An adversary supplies a malicious instruction, scene perturbation, prompt injection, or poisoned memory;
  \item A benign task contains a contextual hazard or a missing precondition;
  \item The agent, tool, or controller deviates from the intended action;
  \item An unresolved hazard persists after the nominal task has succeeded.
\end{enumerate}
The fourth case is especially important for long-horizon agents: a task may satisfy its terminal predicate while leaving a person, appliance, or object in a dangerous state. Existing robot-safety surveys, human--robot collaboration studies, and safe reinforcement-learning surveys provide complementary taxonomies of these issues \cite{li2026embodiedsurvey,haddadin2009safe,villani2018hri,garcia2015safe,brunke2022safe}. For generalist robots, context-dependent and delayed consequences extend the safety problem beyond geometric constraints alone \cite{sinha2026rethinking}.

These situations must be translated into requirements that can be checked during operation. A safety specification should state what is protected, which situation counts as a hazard, when the condition must hold, and what authority the agent has. Examples include excessive contact force, entering a forbidden region, operating an appliance with an incompatible object, exposing private information, or executing an action without authorization. Safety must be checked throughout the episode and again at termination. The final check includes clean-up and hand-over obligations that a task-success label can miss, such as leaving an appliance in a safe state or returning a tool to an authorized person.

These requirements also delimit the agent's authority to act. A deployment should therefore specify which actions may be taken autonomously, which require confirmation, and which are prohibited. These boundaries should be enforced at the tool and controller interfaces rather than inferred from a generated explanation. Safety does not mean refusing every uncertain request. A useful system should complete benign tasks, mitigate contextual hazards when an authorized safe procedure exists, and refuse actions that are prohibited or cannot be executed within the declared limits.

\subsection{Safety on Four Coupled Functions}
\label{sec: safety on functions}
Earlier sections framed agentic embodied intelligence as a feedback loop. Perception grounds the current situation, reasoning and planning select what to do, action execution tests that decision in the physical world, and the observed outcome updates the agent's beliefs and future behavior. Safety is not an additional stage at the end of this loop. It constrains each transition and determines whether the agent may proceed, should gather more evidence, or must stop. We therefore examine safety through the same four coupled functions: hazard awareness in perception, constraint handling in planning, action enforcement during execution, and controlled updates through system evolution.

\textbf{Safety-Critical Perception and Risk Awareness.}
Safety begins with a task-relevant estimate of what may cause harm. This estimate must connect objects to their states, spatial relations, action history, and uncertainty. Detecting an object or generating a verbal warning is not enough, since risk often depends on context, such as whether food is inside a microwave or a person is within the path of a moving object. Recent work on contextual hazard recognition and relation-based evaluation reflects this shift from identifying risky objects to assessing risky situations~\cite{lu2026homeguard,yang2026saferelbench,son2025embodyguard}.
Memory can extend risk assessment beyond the current view by preserving previously observed hazards and failures. Its value depends on whether the relevant evidence was observed and remains applicable to the current robot and environment. Retrieved safety information should therefore retain its source, confidence, time, and conditions of use, rather than being treated as a universal rule~\cite{li2026safetask,sima2026jitmemory}. When neither current observations nor memory resolve a safety-critical uncertainty, the agent should seek a useful view, perform a bounded probe, or ask for clarification. These actions also carry time, physical, and human costs, so information gathering must itself be assessed for safety~\cite{ren2023robotsask}.

\textbf{Safe task decomposition and execution.}
Reasoning and planning make safety requirements operational by turning them into preconditions, ordering constraints, authorization checks, mitigation steps, and terminal obligations. This lets the planner account for hazards that arise between the initial state and the goal, rather than checking only the final action. Work on constraint-based safety planning and explicit robot principles illustrates how such requirements can be made available to deliberation~\cite{yang2024safetychip,sermanet2025constitutions}. Their effectiveness still depends on accurate specifications and grounding. A planner cannot enforce a constraint about a hazard that perception missed or bound to the wrong object.
The planner must then choose a response that respects those constraints. Depending on the evidence and whether the hazard can be mitigated, it may continue, change the action order, gather information, request confirmation, or stop. A task that is unsafe in its current context may become feasible after a verified mitigation, while an unauthorized action may require refusal. Help-seeking can resolve uncertainty that observation alone cannot settle, but its value should be judged together with response time and human burden~\cite{ren2023robotsask}. 

\textbf{Safe Policy and Runtime Enforcement.}
Planning cannot guarantee safe physical execution. A policy may produce an invalid command, its assumptions may become stale, or unexpected contact may change the situation. Safety therefore requires safeguards at multiple points: training can shape policy behavior, predictive checks can screen actions before commitment, and an independent runtime monitor can constrain or stop commands as they reach the robot. Research on constrained learning, predictive filtering, and runtime action enforcement supports this layered view~\cite{achiam2017cpo,zhang2025safevla,wabersich2021psf,althoff2014reachability,ames2019cbf,dalal2018safe,cheng2019barrier,hu2025vlsa,tabbara2025latent,seo2025uncertaintyfilter}.
These safeguards provide different levels of assurance and rely on explicit assumptions. Their protection can fail when the specification is incomplete, the estimated state is stale, the dynamics are inaccurate, or no safe action remains feasible. A runtime interface should therefore check action validity, physical limits, workspace and human-distance bounds, authorization, and timeouts, while preserving an independent emergency stop. Reported improvements should be distinguished from formal guarantees, and any guarantee should state the sensing, model, and feasibility assumptions on which it depends~\cite{haddadin2009safe,villani2018hri}.

\textbf{Safe Reflection, Memory, and Evolution.}
Closed-loop evolution allows an agent to use execution outcomes to improve later behavior, but a safety-oriented update must preserve protections across episodes. This is distinct from the capability-improvement loop discussed in Section~\ref{sec:loop}. Here, the central question is whether a lesson from one interaction can be reused without extending it beyond the evidence that supports it. Safety memories should retain their provenance, confidence, and conditions of applicability, and reflection should change a checkable prerequisite, skill, or monitor rather than only produce a post-hoc explanation~\cite{sima2026jitmemory,li2026safetask,roboosnext2025}.
Persistent changes should pass an independent release process. Incident evidence may motivate a candidate update, but promotion should follow replay, adversarial testing, and regression checks on both hazardous and benign tasks. Accepted changes should be versioned, monitored, and reversible, with a defined scope and rollback path. The mechanism that proposes an update should not be its sole evaluator, and software rollback cannot undo physical harm. 

\subsection{Evaluation and Assurance Gaps}
\label{sec:safety-evaluation}

We organize safety evidence as a claim--assumption--test triple, informed by the diagnostic emphasis of recent embodied-safety benchmarks and surveys \cite{li2026embodiedsurvey,son2025embodyguard,lu2026isbench,ying2025agentsafe}. The claim might be ``the agent removes a contextual hazard and finishes the task''; assumptions include the correctness of the scene representation, the action interface, and the hazard label; and the test must expose the assumptions through perturbations and independent observation. Table~\ref{tab:safety-benchmarks} summarizes the evidence and limits of representative benchmarks and safeguard evaluations.

\begin{table}[t]
\centering
\small
\caption{\textbf{Representative safety evaluation benchmarks.}}
\label{tab:safety-benchmarks}
\begin{tabularx}{\linewidth}{@{}p{0.24\linewidth}p{0.32\linewidth}X@{}}
\toprule
\textbf{Evaluation} & \textbf{Useful evidence} & \textbf{Remaining boundary}\\
\midrule
\rowcolor{AEIrow}
Robot constitutions and semantic cases \cite{sermanet2025constitutions} & Whether stated principles and semantic hazards are recognized and followed & Does not test noisy contact dynamics or long-horizon execution\\
AGENTSAFE \cite{ying2025agentsafe} & Perception--planning--execution behavior under hazardous instructions & Does not cover all benign tasks requiring active mitigation\\
\rowcolor{AEIrow}
IS-Bench \cite{lu2026isbench} & Interactive, prerequisite, and postcondition safety in household tasks & Safety remains relative to its task ontology and simulator/robot setup\\
SafeRelBench \cite{yang2026saferelbench} & Process risks caused by support, containment, and proximity relations & Limited coverage of hidden state, multi-robot interaction, and hardware faults\\
\rowcolor{AEIrow}
SafeVLA \cite{zhang2025safevla}, VLSA \cite{hu2025vlsa}, latent filters \cite{tabbara2025latent} & Policy safety costs and action-level constraint enforcement & Does not by itself establish privacy, authorization, or persistent-memory safety\\
\bottomrule
\end{tabularx}
\end{table}

\textbf{Structuring safety evidence.}
A safety result is meaningful only relative to a specific claim and the assumptions on which it depends. For example, a claim that an agent mitigates a contextual hazard and completes a task presupposes sufficiently accurate scene state, hazard labels, and action interfaces; evaluation should expose these assumptions through controlled perturbations and independent observations. This claim--assumption--test framing follows the diagnostic emphasis of embodied-safety benchmarks and surveys~\citep{li2026embodiedsurvey,son2025embodyguard,lu2026isbench,ying2025agentsafe}. 

\textbf{Separating safety outcomes.}
Evaluation should distinguish benign tasks, repairable contextual hazards, and prohibited or irreparable actions because they require different behavior. Report task success, safe completion, and erroneous refusal for benign tasks; for repairable hazards, assess mitigation, exposure, near misses, and residual risk at hand-over; and for prohibited actions, measure appropriate refusal or safe termination. Interactive benchmarks such as IS-Bench and hazardous-instruction evaluations such as AGENTSAFE address complementary cases~\citep{lu2026isbench,ying2025agentsafe}. Keeping these outcomes separate prevents blanket refusal from appearing safe when the agent cannot mitigate a hazard and continue a benign task.

\textbf{Reporting metrics and controlling tests.}
Safety metrics need explicit denominators: rates computed only over successful episodes omit important failures, while mitigation success conditional on hazard detection should be distinguished from end-to-end performance over all hazard cases. Measures should cover both risk awareness (e.g., detection, localization, and calibration) and execution (e.g., violation severity, safety margins, and intervention cost), reported separately from task return as in Safety Gym~\citep{ray2019safetygym}. Tests should publish their predicates and trigger conditions, include hazardous and non-hazardous controls, and stratify results by hazard, scene, duration, and embodiment~\citep{sermanet2025constitutions,yang2026saferelbench}. State and trajectory logs are preferable to an LLM judge; semantic judging requires human validation, and repeated trials should report uncertainty.

\textbf{Attributing safety improvements.}
To determine which components improve safety, compare a baseline with safeguards, risk reasoning, memory, and active mitigation while holding the backbone, action policy, observation budget, tools, and compute fixed. This component-level design is consistent with diagnostic evaluations of embodied safeguards~\citep{son2025embodyguard,tabbara2025latent,seo2025uncertaintyfilter}. Include clean and adversarial scenes, out-of-distribution objects, corrupted memory, delayed sensing, and post-update regression tests so that gains are not confounded with additional model capacity or resources.

\textbf{Remaining assurance gaps.}
Current evaluations lack consistent specifications for hidden state, irreversible consequences, near misses, and prolonged exposure~\citep{li2026embodiedsurvey,son2025embodyguard,yang2026saferelbench}. Moreover, module-level safety does not guarantee system-level safety: planner, memory, policy, and controller may share failure modes or rely on incompatible assumptions~\citep{brunke2022safe,ames2019cbf}. Benchmarks must also reward hazard discovery and mitigation without encouraging unnecessary refusal, delay, repeated help seeking, or risky mitigation~\citep{ren2023robotsask,lu2026isbench}. Thus, accident-free trials support claims only within the tested conditions; they cannot establish zero risk. Stronger assurance requires a coherent evidence chain alongside more capable models.

%% file: sec/7_infra.tex
\begin{figure*}[t]
\centering
\includegraphics[width=\linewidth]{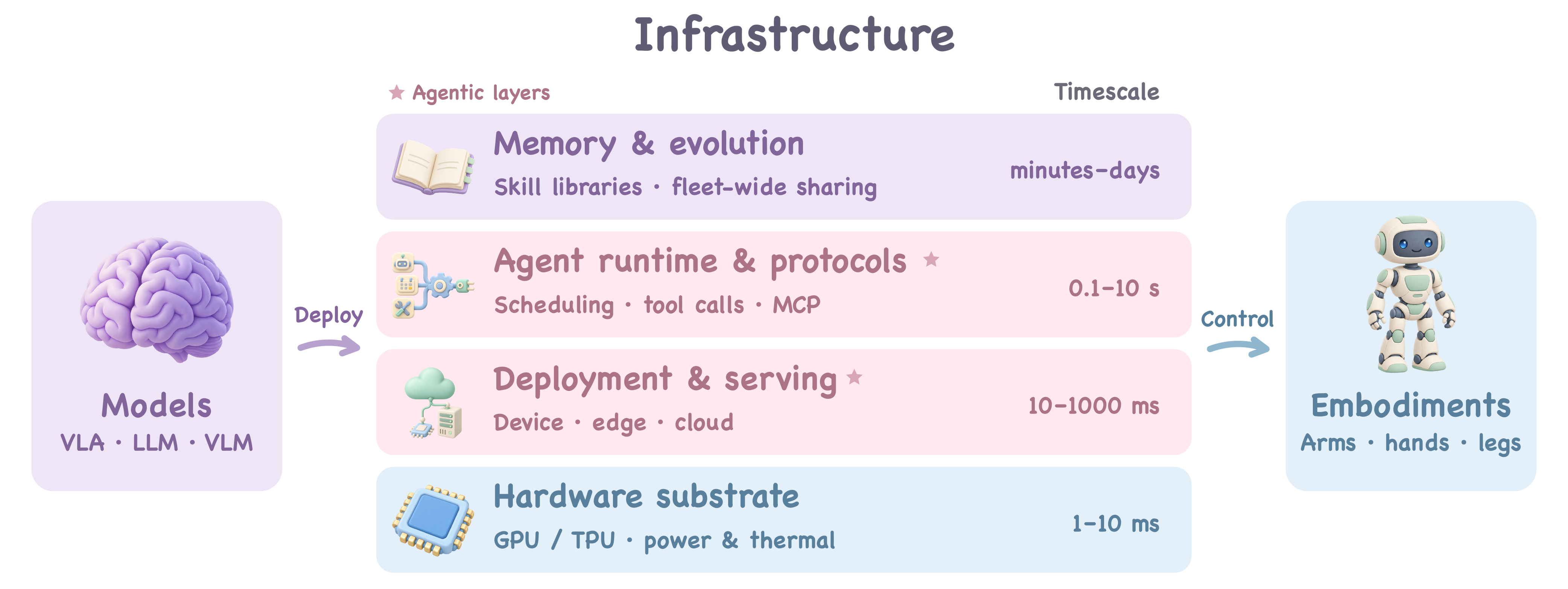}
\caption{\textbf{Positioning of infrastructure as the bridge between models and embodiments.} It receives algorithms from above and manages hardware from below.}
\label{fig:ch7-bridge}
\end{figure*}

The capability ceiling of agentic embodied intelligence is no longer determined by models alone.
VLA models attain strong single-task performance, yet their deployment encounters two structural obstacles: a \emph{capacity mismatch} between billion-parameter models and palm-sized onboard compute, and an \emph{architectural limitation} of single-shot mappings over long-horizon tasks.
The agentic paradigm resolves the second obstacle by decomposing the system into a deliberating model, fast-executing policies, memory, and standardized tool interfaces---and in doing so it multiplies the demands placed on the layer beneath.
This section therefore treats computing infrastructure as the bridge between models and embodiments. As shown in Figure~\ref{fig:ch7-bridge}, a model is not embodied until its outputs are realized on actuators, and a body remains unproductive until it receives structured intent.
We organize the discussion in four parts.
Section~\ref{sec:infra-req} analyzes how the agentic shift changes the infrastructure requirement, and introduces the four-layer stack used throughout.
Section~\ref{sec:infra-deploy} studies deployment: how computation is mapped onto device, edge, and cloud, and where the split is placed.
Section~\ref{sec:infra-runtime} examines the agent runtime and the Model Context Protocol (MCP), which integrate upward into the algorithm stack and downward into the hardware stack, respectively.

\subsection{The Infrastructure Requirement}
\label{sec:infra-req}

Two quantitative observations anchor the chapter.
First, open VLAs now span roughly 1B to 55B parameters~\cite{kim2024openvla,loop_brohan2023rt2}, whereas robot-onboard compute is an edge module bound by hard power, thermal, and cost budgets.
Second, as formalized in Table~\ref{tab:embodied_autonomy_levels}, end-to-end policies operate largely within L2. The transition to L3--L5 requires task decomposition, tool invocation, long-horizon memory, and closed-loop recovery.
These capabilities do not reside in the policy network itself, but are carried by the infrastructure layer.
The higher the autonomy level, the more the differentiator is the substrate rather than the model.

Under the end-to-end paradigm, this layer faced a single question: how to execute one model efficiently.
However, under the agentic paradigm, it faces a chain of systems questions concerning computation placement, communication, scheduling, memory, and recovery.
Whether inference should reside onboard or be consumed as a hosted service remains itself an open debate~\cite{apisurvey2026}.
To structure these questions, we adopt a four-layer stack with explicit per-layer timescales: hardware, deployment and serving, agent runtime and protocols, and memory.
Real-time constraints tighten toward the bottom: control loops operate at millisecond rates, policies at tens of milliseconds, and deliberation at second rates.

\subsection{Deployment across Device, Edge, and Cloud}
\label{sec:infra-deploy}

The \emph{device side} is the compute module carried by the robot. The \emph{edge side} is an on-site server, canonically a single workstation-class machine in a factory. The \emph{cloud side} is data-center-scale accelerators.

The deployment problem originates in a simple comparison: OpenVLA~\cite{kim2024openvla} carries approximately 7B parameters, whereas the lightweight action predictor that executes in real time onboard in VLA-ULAP~\cite{ulap2026} carries 7.4M, a gap of nearly three orders of magnitude.
Engineering narrows this gap but does not close it. Onboard compute hosts reflex-scale competence, not full model capability.
Therefore, the deployment is classified into three routes:
\begin{enumerate}
\item Route A (fully onboard) compresses the model onto the robot~\citep{deer2024,hirt2024,fasa2026}. Compute, energy, and thermals are paid by the platform, and performance is capped by onboard hardware.
\item Route B (fully cloud) relocates inference wholesale to the data center~\citep{loop_brohan2023rt2, kuffner2010cloud, fog2019, fogros2021}.
The route is fragile to network jitter, bandwidth, and privacy constraints.
\item Route C (hybrid) combines cloud deliberation with onboard reflexes~\citep{ulap2026,ecovla2026,comvla2026,vlaserving2026}.
\end{enumerate}
The three deployment routes differ not only in where inference runs, but also in whether they can meet the timing requirements of each control loop. 
Fast feedback and motor control require predictable, low-latency responses~\citep{ulap2026}, whereas high-level deliberation can tolerate slower computation. Latency, especially tail latency and disconnection risk, therefore determines which deployment route is suitable for each function~\citep{hirt2024}.

\begin{figure}[t]
\centering
\includegraphics[width=\linewidth]{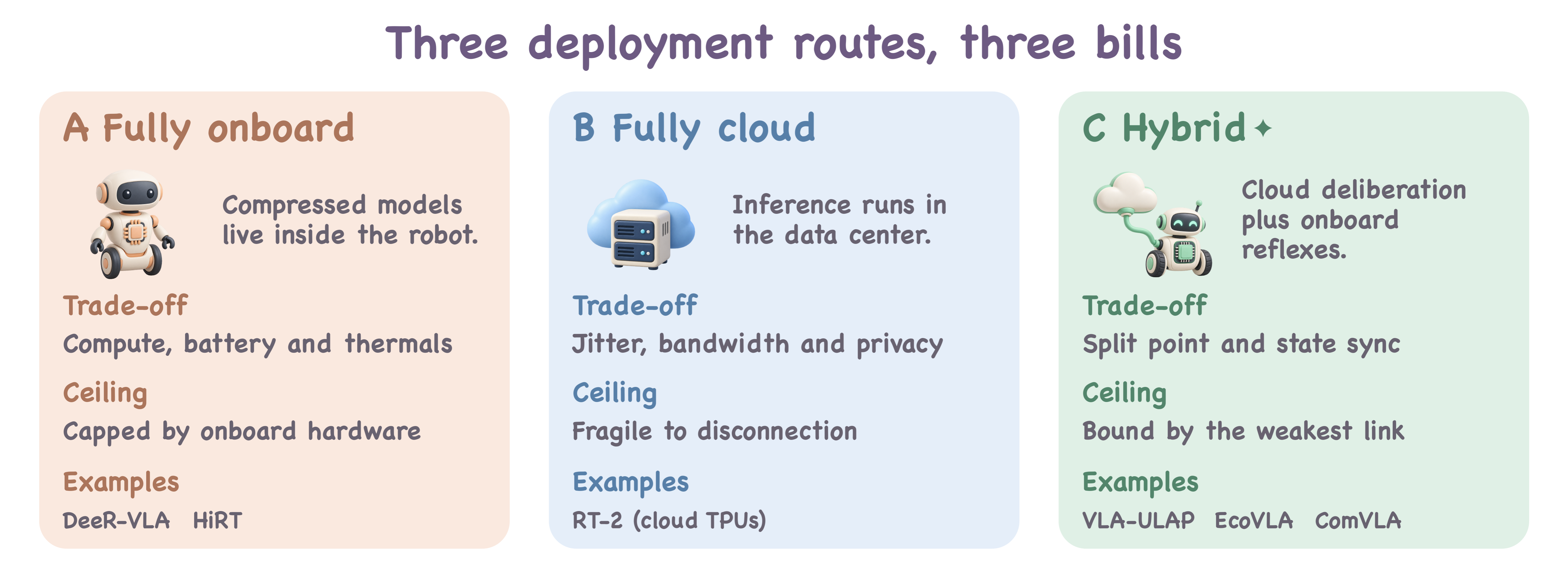}
\caption{\textbf{\textbf{Three deployment routes and their cost profiles.}}}
\label{fig:ch7-routes}
\end{figure}

The case for hybrid organization must be made per control loop.
The fast loop operates at tens of hertz with a budget of tens of
milliseconds~\cite{ulap2026}, whereas a cloud round trip already averages $\sim$100\,ms. Crucially, tail latency and disconnection risk make a purely cloud-fed fast loop unsuitable. The slow deliberation loop can better tolerate such delays~\cite{hirt2024}. This difference motivates a hybrid organization in which an onboard predictor emits action chunks while a cloud VLA is invoked on demand, with their outputs interleaved
during execution~\cite{ulap2026}.
Once inference is distributed across device and remote resources, the split must account for more than latency alone. Energy and deadline budgets can make the choice between local execution and offloading vary from step to step~\cite{ecovla2026,edgedlm2026}. Communication capacity further limits the amount of sensor data that can be transmitted for each decision, especially in wireless and 6G-connected settings~\cite{comvla2026}. At the edge, server load and per-robot service-level objectives constrain
how inference can be shared across a  fleet~\cite{kwon2023vllm,zheng2024sglang,vlaserving2026}.
Together, these constraints make the split point an online scheduling decision rather than a fixed topology choice. The system must continually balance deadlines, energy, bandwidth, and compute.


\subsection{Agent Runtimes and Interfaces}
\label{sec:infra-runtime}

Runtime and interface design addresses how models, policies, tools, and robots coordinate during execution. The runtime schedules these components and manages their state, while interfaces define how they exchange information and invoke capabilities.

At the tool boundary, the Model Context Protocol provides a common way to expose callable functions, readable resources, and reusable prompts. Early robotics integrations show how this interface can make robot data and domain capabilities accessible to model-driven agents~\cite{rosbagmcp2025,drivemcp2026}. Its role is to standardize higher-level tool access, not to replace the communication mechanisms required for low-latency control. A useful interface must also make each capability's inputs, outputs, authority, and execution status clear enough for the agent to select and verify it.

Above these interfaces, an agent runtime coordinates task scheduling, state transfer across models and policies, capability handoffs, failure handling, and shared resources. Emerging embodied operating systems illustrate this supervisory role across heterogeneous robots and persistent memory~\cite{roboos2025,roboosnext2025}. The runtime must also bridge interfaces that operate at different timescales. ROS~2 and its Data Distribution Service transport support high-frequency communication between robot components, whereas MCP is designed for higher-level tool invocation. Keeping these layers distinct allows control traffic to use its native transport while the runtime translates task-level requests into robot capabilities and returns execution outcomes to the agent.

This runtime layer also connects online execution with longer-term learning. Interaction histories and fleet-wide experience can support offline training, but their value depends on whether relevant evidence can be retrieved before a decision and whether its provenance and freshness are preserved~\cite{roboosnext2025,datafactory2026}. The central systems challenge is therefore to coordinate immediate execution with reliable access to prior experience, without confusing historical records with the robot's current state.

\subsection{Hardware and Embodiments}
\label{sec:infra-hw}

Hardware sets the operating limits within which models and runtimes must function. Onboard compute, energy, and thermal capacity constrain the size and rate of local inference, while cloud and edge resources offer greater capacity at the cost of communication delay and dependence. This gap makes deployment across local and remote resources a persistent systems requirement. Integrated model and hardware efforts, such as GR00T N1, illustrate the value of aligning deliberative computation with fast onboard control, but do not remove the underlying limits of robot-side resources~\cite{nvidia2025groot}. 

The robot's morphology and end effector further determine which actions are physically available and at what speed and precision they can be executed. These properties shape the policies a system can use, the skills it can compose, and the control workloads that must remain onboard. A mobile base, a humanoid platform, and a dexterous hand therefore imply different execution and resource requirements rather than interchangeable hardware choices. A detailed comparison of robot platforms and end effectors is beyond this survey's scope; we treat embodiment as a constraint on planning, control, and deployment.

%% file: sec/9_data.tex
\begin{figure*}[t]
\centering
\includegraphics[width=\textwidth]{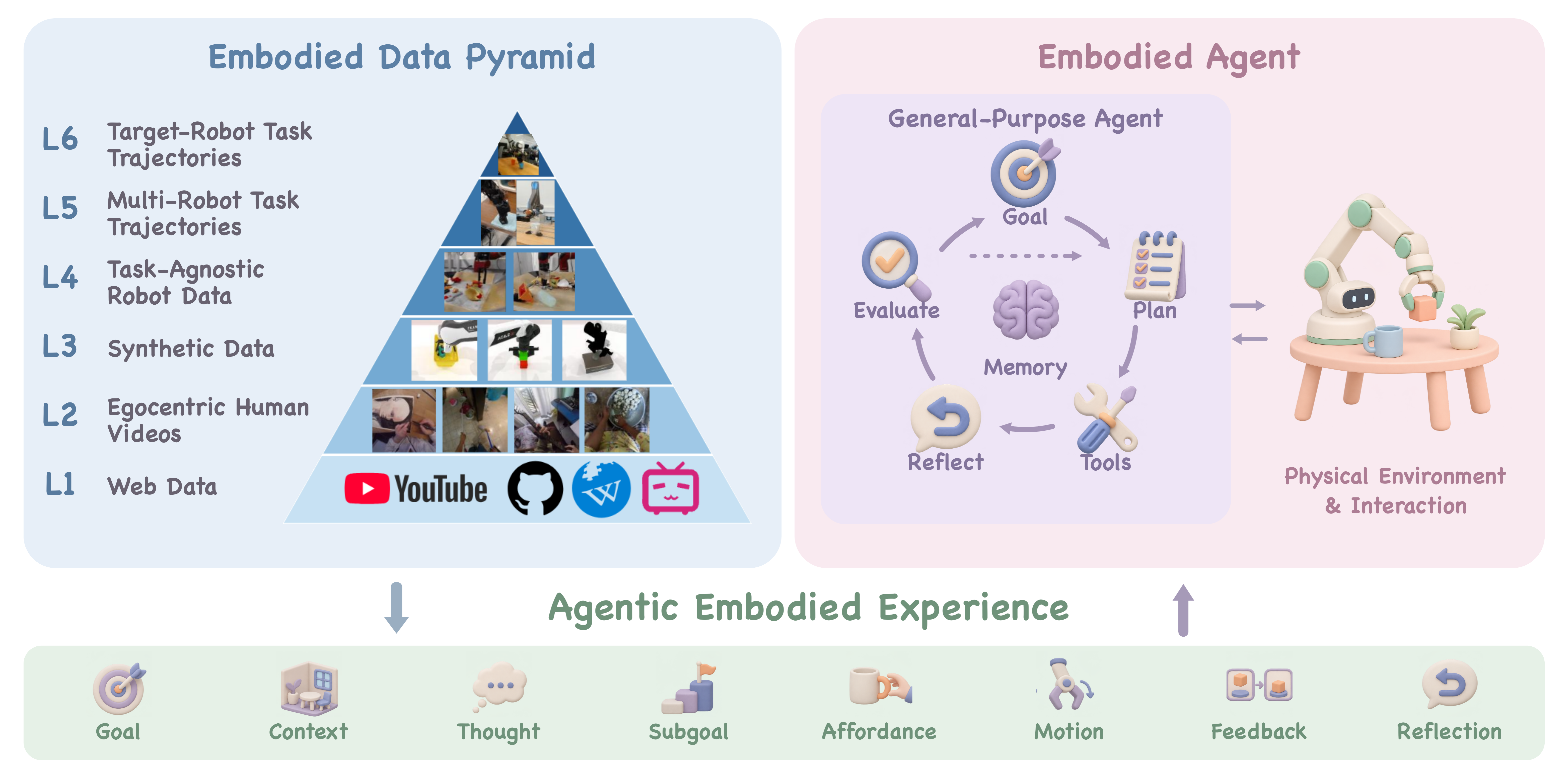}
\caption{\textbf{From the embodied data pyramid to Agentic Embodied Experience.}
The six-level data pyramid of Motus~\citep{bi2025motus} organizes
complementary sources. To enable embodied agents to integrate goal-directed planning, tool use, evaluation, and reflection with physical interaction, data resources connect decisions to execution, observed consequences, and subsequent revision. 
The Agentic Embodied Experience schema captures these connections through eight linked elements, organizing
heterogeneous data into experience that supports adaptive embodied behavior.}
\label{fig:agentic_embodied_experience}
\end{figure*}

\input{tab/9_data_sources}

Agentic embodied intelligence requires data that connect task goals and decision rationale to physical execution, observed consequences, and subsequent correction. Such experience supports long-horizon behavior, failure recovery, and adaptation as tasks and environments change. This chapter examines the data ecosystem through the lens of \emph{Agentic Embodied Experience}, focusing on how experience is organized, generated, and used for continued learning.

\subsection{From the Data Pyramid to Agentic Embodied Experience}
\label{sec:causal_trajectory}


\noindent\textbf{Data Pyramid.} The data pyramid is a widely used view of embodied data organization. Here we adopt the six-level formulation of Motus~\cite{bi2025motus}, which organizes resources by source and training role relative to a target robot. From bottom to top, its levels are:

\begin{enumerate}
\item[$\circ$] \textbf{D1: Web Data.} Text, images, and videos supply visual, linguistic, and world-knowledge priors.
\item[$\circ$] \textbf{D2: Egocentric Human Videos.} Human recordings capture interactions, motions, and task knowledge.
\item[$\circ$] \textbf{D3: Synthetic Data.} Simulation and digital twins provide scalable scenes and interaction data.
\item[$\circ$] \textbf{D4: Task-Agnostic Data.} Robot sampling covers motion and interaction without semantic goals.
\item[$\circ$] \textbf{D5: Multi-Robot Task Trajectory Data.} Demonstrations support cross-embodiment knowledge sharing.
\item[$\circ$] \textbf{D6: Target-Robot Task Trajectory Data.} Target-platform demonstrations support learning and adaptation.
\end{enumerate}

The pyramid captures a broad shift from abundant general priors to targeted robot experience. For agentic learning, a further question is whether the data connect a decision to its actual execution, observed consequences, and subsequent revision. Figure~\ref{fig:agentic_embodied_experience} illustrates how the data pyramid and embodied agent requirements motivate this experience view.

\noindent\textbf{Agentic Embodied Experience Schema.}
General-purpose agents pursue goals through a recurring cycle of deciding,
acting, observing outcomes, and revising subsequent behavior. ReAct~\cite{loop_yao2023react}
illustrates the interplay of reasoning and action, while
Reflexion~\cite{loop_shinn2023reflexion} shows how feedback can inform later
attempts. For embodied agents, this cycle must also connect decisions to
interaction targets, executed motion, and physical consequences. We therefore
introduce the following schema for organizing
\emph{Agentic Embodied Experience}:
\begin{equation}
\label{eq:causal_trajectory}
\begin{aligned}
\mathcal{E}_{\mathrm{agentic}}=\big\langle
&\mathcal{G},\mathcal{C}, \mathcal{T},\mathcal{Z}, \mathcal{A},\mathbf{a},\mathcal{O},\mathcal{R}
\big\rangle .
\end{aligned}
\end{equation}
\begin{enumerate}
\item[$\circ$] \textbf{Goal ($\mathcal{G}$).} An explicit task objective or specification establishes the intended outcome.
\item[$\circ$] \textbf{Context ($\mathcal{C}$).} Visual observations, states, and  history ground decisions in the current situation.
\item[$\circ$] \textbf{Thought ($\mathcal{T}$).} Annotated or generated explanatory rationales link available evidence to decision choices.
\item[$\circ$] \textbf{Subgoal ($\mathcal{Z}$).} Plan steps or temporally annotated action/skill segments organize execution into stages.
\item[$\circ$] \textbf{Affordance ($\mathcal{A}$).} Contact/interaction annotations, and feasibility labels ground where and how to interact, with different degrees of physical specificity.
\item[$\circ$] \textbf{Motion ($\mathbf{a}$).} State trajectories, control actions or targets, and human hand poses or position sequences record the embodied realization of behavior.
\item[$\circ$] \textbf{Feedback ($\mathcal{O}$).} Interaction effects provide evidence of progress and deviation, including passively observed physical events.
\item[$\circ$] \textbf{Reflection ($\mathcal{R}$).} Post-execution diagnosis or corrective commentary supports behavioral revision and later attempts.
\end{enumerate}

Table~\ref{tab:pyramid_causal_fields} relates two complementary views of embodied data. The pyramid describes where data come from and what training role they serve, while the experience fields describe which parts of an agent's interaction they record. Web data mainly provide visual and language context, but rarely connect a decision to a robot action or its physical outcome~\cite{schuhmann2022laion5b,bain2021frozen}. Human activity videos provide richer evidence about action sequences, subgoals, contact, and observed consequences, although human motion is not equivalent to robot control and the availability of explicit goals or reasoning depends on annotation~\cite{grauman2022ego4d,damen2020rescaling,grauman2024egoexo4d}. Simulation offers controllable interactions and repeatable state changes, which are useful for studying physical consequences and collecting failure cases, but its value for real robots depends on transfer across the simulation gap~\cite{ahmed2020causalworld,duan2024aha,loop_liu2023reflect}. Task-agnostic robot data contribute robot-specific motion and interaction outcomes without consistently specifying the task that caused them. Multi-robot task trajectories connect goals and context to robot actions across embodiments, while target-robot trajectories provide the closest match to the platform that will be deployed. These task-oriented sources therefore offer stronger action grounding, but their breadth and annotation of reasoning, affordances, and reflection vary across resources and subsets~\cite{ebert2017selfsupervised,dasari2019robonet,fang2023rh20t, loop_oxe2023,wu2024robomind,zhao2023learning,loop_khazatsky2024droid}.  

Context, motion, and observed outcomes are recorded more consistently than the decision processes connecting them. Explicit thought, subgoal, affordance, and reflection annotations appear mainly in selected resources or subsets. Coverage of all eight fields does not guarantee their co-occurrence within samples or temporal alignment. Failure records are particularly useful when they preserve both deviations and recovery: outcomes ground diagnoses, and recovery trajectories test whether corrections worked. Such records supervise behavioral revision beyond binary failure labels~\cite{wu2024robomind,loop_liu2023reflect,duan2024aha}. The central challenge is to align decisions, actions, outcomes, and revisions across heterogeneous sources while preserving temporal order and provenance.

\subsection{Generating Agentic Embodied Experience}
\label{sec:multi_source_agentic_experience}

The Agentic Embodied Experience schema identifies the information needed to connect decisions, physical execution, and revision. As shown in Table~\ref{tab:agentic_experience_sources}, this information is acquired through active exploration, human videos, simulation, and human corrections. These routes enrich existing resources through new interactions and annotations, contributing complementary evidence about goals, action choices, and outcomes.

\begin{table}[t]
\centering
\small
\setlength{\tabcolsep}{3pt}
\renewcommand{\arraystretch}{1.15}
\caption{\textbf{Complementary routes to generating Agentic Embodied Experience.}}
\label{tab:agentic_experience_sources}
\begin{tabular}{@{}p{0.24\linewidth}p{0.42\linewidth}p{0.28\linewidth}@{}}
\toprule
\textbf{Source} & \textbf{Primary signal} & \textbf{Strength} \\
\midrule
\rowcolor{AEIrow}
Active exploration & Physical responses to selected actions & Direct interaction feedback \\
Human videos & Intent, attention, and affordance annotations & Broad task diversity \\
\rowcolor{AEIrow}
Simulation & Plans, critiques, and tested revisions & Controllable and scalable \\
Human corrections & Language feedback and corrective actions & Targeted supervision \\
\bottomrule
\end{tabular}
\end{table}

\textbf{Active Exploration.} Active exploration treats action as an information-seeking query, targeting reducible epistemic uncertainty by selecting an admissible probe $a_t^{\star}=\arg\max_{a\in\mathcal{U}_t} I\!\left(\Theta;o_{t+1}\mid h_t,a\right)$, where $h_t$ is interaction history and $\Theta$ denotes unknown scene properties or dynamics. Information-gain and predictive-disagreement objectives guide probes that reveal hidden dynamics, geometry, and contact; visual, tactile, and physical exploration can turn these interactions into reusable transitions for modeling and downstream manipulation~\citep{bohg2017interactive,houthooft2016vime,pathak2019disagreement,sekar2020planning,schneider2022activeexploration,choi2025activevisiontouch,iten2026tactilecuriosity}. On a broader scale, self-supervised poking and cross-robot collections capture action-conditioned visual effects~\citep{agrawal2016learning,dasari2019robonet,ren2024exploreconfident}. Useful records connect the gap motivating a probe to its context, observations, measured outcome, and resulting update, retaining informative failures and successful contacts.

\textbf{Human Videos.}
Human activity videos broaden experience across everyday objects, tasks, and environments: narrated egocentric and kitchen recordings capture object interactions and state changes, while synchronized views, gaze, temporal descriptions, hand--object grounding, and rationales connect behavior to attention and task stages~\citep{grauman2022ego4d,damen2020rescaling,grauman2024egoexo4d,lai2023gazeanticipation,jawaid2025openego,pei2025egothinker}. Complementary affordance annotations specify contact points, trajectories, and 3D interaction regions, and human-derived interaction priors can guide robot exploration~\citep{yu2023fineaffordance,yoshida2024textaff,liu2024vsad,bahl2023affordances}. Together, video-based experience links activity context to relevant objects, action phases, and likely interaction sites, but annotations remain distributed across resources and subsets rather than forming a uniformly aligned, robot-executable schema.

\textbf{Simulation.} Simulation expands experience in controllable settings by reconstructing scenes from video or composing new physical arrangements, then generating tasks, demonstrations, and supervision at scale~\citep{xia2026simrecon,dong2026replicateanyscene,lyu2026physga,loop_wang2024gensim,loop_wang2024robogen}. Interactive rollouts extend these synthetic demonstrations by capturing how plans and actions change as simulator observations trigger critiques, replanning, constraints, or risk-sensitive revisions~\citep{singh2024malmm,zhang2024read,gokhale2025logicguard,wang2025madra}. Preserving each rollout's initial state and goal, tool calls, failed branches, revisions, and outcomes creates supervision for action selection and recovery; rationale-enriched expert transitions provide a complementary distillation signal~\citep{choi2024deder}. These records support scalable learning, but simulated outcomes remain conditional on simulator assumptions and do not alone establish physical transfer.

\textbf{Human Corrections.} Human feedback provides targeted supervision when agent behavior diverges from intent: language corrections can revise goals, action order, or local motion and support planning, shared autonomy, iterative policy improvement, or reusable correction memories~\citep{loop_huang2022innermonologue,cui2023lilac,shi2024yell,zha2023droc}. Action-level interventions, from takeover and corrective teleoperation to rewind-and-retry, generated corrective trajectories, and compliant incremental guidance, concentrate demonstrations around learner errors and recovery states~\citep{ross2011reduction,kelly2018hg,mandlekar2020iwr,hu2025rac,hoque2024intervengen,xu2025crdagger}. Physical corrections can also be translated into language refinements for later planning~\citep{zhang2024physicalcorrection,jin2025talktweak}. Recording the original context and behavior, correction signal, agent interpretation, and subsequent outcome turns one-off interventions into reusable evidence of how feedback changes decisions and execution.

\subsection{Closing the Loop: The Agentic Data Flywheel}
\label{sec:self_improving_agentic_flywheel}

The preceding section describes how Agentic Embodied Experience is
generated. Turning this experience into sustained capability growth requires connecting task selection, execution evaluation, and policy improvement.
The agentic data flywheel links these stages in a recurring cycle:
curricula select informative tasks, execution produces new experience, evaluators add feedback, and learning converts that experience into improved decisions and skills. Updated agents then shape the next round of
collection.

\textbf{Choosing What to Collect Next.} 
Adaptive curricula steer collection toward interactions suited to the learner's current competence, using learning progress and generated goals to expand task difficulty and diversity across simulated and real-world settings~\citep{florensa2018automatic,portelas2019teacher,loop_wang2024gensim,loop_ahn2024autort}. In the data flywheel, verified successes, failures, and recovery attempts expose gaps in existing experience and guide the next collection round, progressively extending coverage from familiar behaviors to underrepresented conditions.

\textbf{Assessing Interaction Quality and Generating Feedback.} Interaction assessment converts execution records into complementary supervision using human labels, vision--language comparisons, annotated perturbations, or generated reward functions~\citep{du2023successdetectors,wang2024rlvlmf,duan2024aha,loop_ma2024eureka}. When aligned with the corresponding observations and actions, these signals support policy and recovery learning while exposing experience gaps for the next curriculum.

\textbf{Learning from Experience to Improve Policies.} Experience can improve both deliberation and control: feedback-derived reflections and rationale supervision inform later decisions, while optimized or self-generated action-labeled trajectories train motor policies~\citep{loop_shinn2023reflexion,choi2024deder,levine2016end,loop_bousmalis2023robocat}. Linking these updates to observations and execution outcomes makes experience reusable for future decisions and actions. Improved agents can then tackle more demanding tasks, with their successes and remaining failures guiding the next curriculum and closing the loop between data collection and capability growth.

\begin{figure}[t]
\centering
\includegraphics[width=0.7\linewidth]{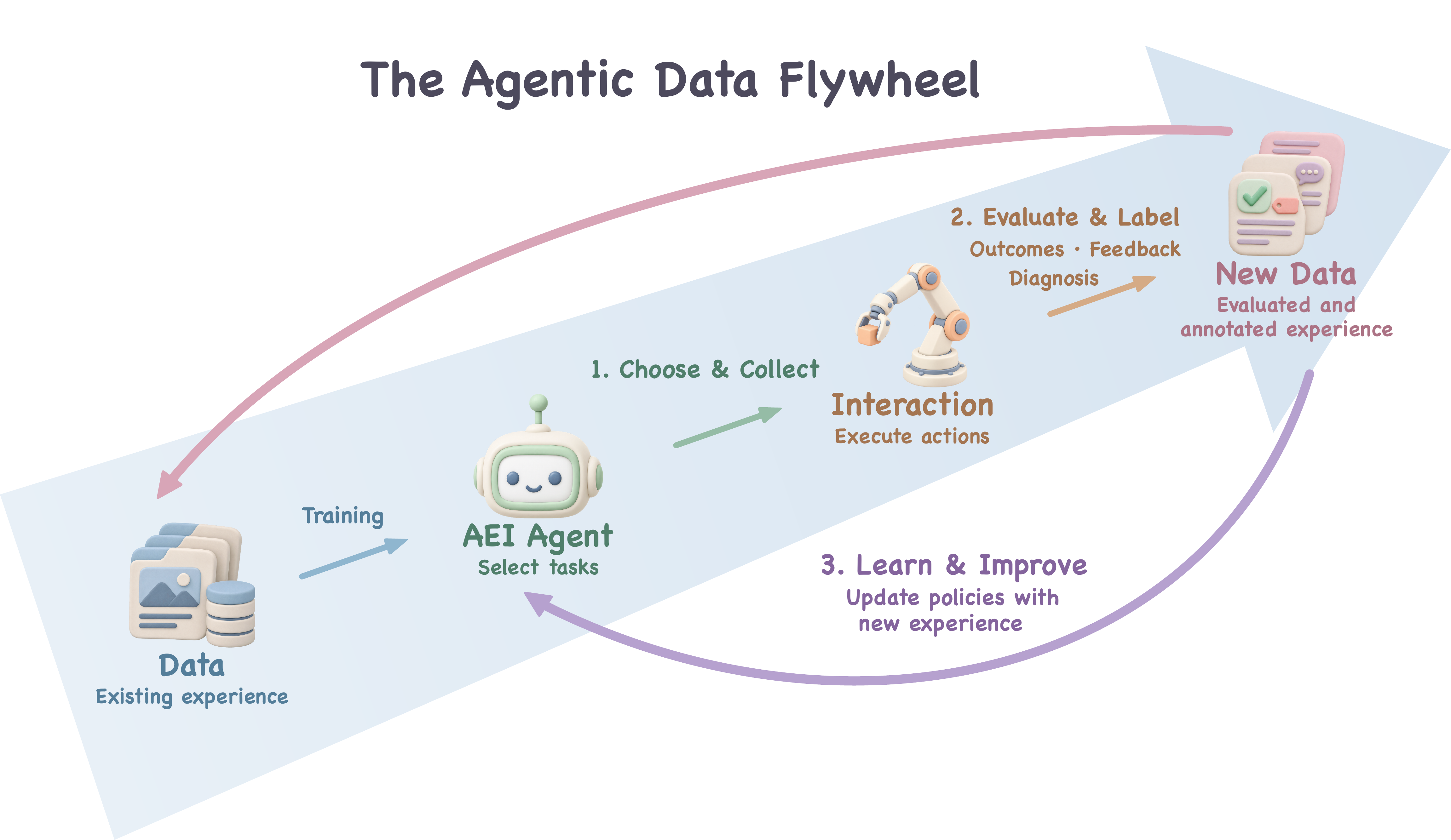}
\caption{\textbf{The agentic data flywheel.} Agents trained on existing experience select tasks and collect interactions. Assessing interaction outcomes produces feedback, diagnoses, and annotated experience for policy updates.
Improved agents then drive the next round of collection, linking data growth to capability improvement.}
\label{fig:agentic_data_flywheel}
\end{figure}

%% file: tab/9_data_sources.tex
\begin{table*}[!t]
\setlength{\tabcolsep}{6pt}
\renewcommand{\arraystretch}{1.05}
\centering
\caption{\textbf{Data resources across the embodied data pyramid, ordered from
D1 to D6, and their explicit coverage of Agentic Embodied
Experience fields.} A checkmark indicates that the named resource or specified subset
provides the field. a cross indicates that it is not explicitly supplied under the definitions above.}
\label{tab:pyramid_causal_fields}
\begin{tabular}{p{0.55\textwidth}c*{8}{c}}
\toprule
\textbf{Dataset / data resource} & \textbf{Level}
& $\mathcal{G}$
& $\mathcal{C}$
& $\mathcal{T}$
& $\mathcal{Z}$
& $\mathcal{A}$
& $\mathbf{a}$
& $\mathcal{O}$
& $\mathcal{R}$ \\
\midrule
LAION-5B~\cite{schuhmann2022laion5b} & \multirow[c]{2}{*}{D1} & $\times$ & $\checkmark$ & $\times$ & $\times$ & $\times$ & $\times$ & $\times$ & $\times$ \\*
WebVid~\cite{bain2021frozen} &  & $\times$ & $\checkmark$ & $\times$ & $\times$ & $\times$ & $\times$ & $\times$ & $\times$ \\
\midrule
Ego4D~\cite{grauman2022ego4d} & \multirow[c]{11}{*}{D2} & $\times$ & $\checkmark$ & $\times$ & $\checkmark$ & $\checkmark$ & $\checkmark$ & $\checkmark$ & $\times$ \\*
EPIC-KITCHENS-100~\cite{damen2020rescaling} &  & $\times$ & $\checkmark$ & $\times$ & $\checkmark$ & $\checkmark$ & $\times$ & $\checkmark$ & $\times$ \\*
Ego-Exo4D (annotated subsets)~\cite{grauman2024egoexo4d} &  & $\checkmark$ & $\checkmark$ & $\checkmark$ & $\checkmark$ & $\checkmark$ & $\checkmark$ & $\checkmark$ & $\checkmark$ \\*
EgoDex~\cite{hoque2025egodex} &  & $\checkmark$ & $\checkmark$ & $\times$ & $\times$ & $\times$ & $\checkmark$ & $\checkmark$ & $\times$ \\*
OpenEgo~\cite{jawaid2025openego} &  & $\checkmark$ & $\checkmark$ & $\times$ & $\checkmark$ & $\times$ & $\checkmark$ & $\checkmark$ & $\times$ \\*
EgoRe-5M (CoT / temporal QA / grounding subsets)~\cite{pei2025egothinker} &  & $\times$ & $\checkmark$ & $\checkmark$ & $\checkmark$ & $\checkmark$ & $\times$ & $\checkmark$ & $\times$ \\*
EPIC fine-grained affordance annotations~\cite{yu2023fineaffordance} &  & $\times$ & $\checkmark$ & $\times$ & $\checkmark$ & $\checkmark$ & $\times$ & $\checkmark$ & $\times$ \\*
TextAFF80K~\cite{yoshida2024textaff} &  & $\checkmark$ & $\checkmark$ & $\times$ & $\times$ & $\checkmark$ & $\times$ & $\times$ & $\times$ \\*
VSAD~\cite{liu2024vsad} &  & $\times$ & $\checkmark$ & $\times$ & $\times$ & $\checkmark$ & $\times$ & $\checkmark$ & $\times$ \\*
Causal Triplet (EPIC-KITCHENS subset)~\cite{liu2023causaltriplet} &  & $\times$ & $\checkmark$ & $\times$ & $\times$ & $\times$ & $\times$ & $\checkmark$ & $\times$ \\*
RoboVQA (human video--QA subset)~\cite{sermanet2023robovqa} &  & $\checkmark$ & $\checkmark$ & $\times$ & $\checkmark$ & $\checkmark$ & $\times$ & $\checkmark$ & $\times$ \\
\midrule
Causal Triplet (ProcTHOR subset)~\cite{liu2023causaltriplet} & \multirow[c]{7}{*}{D3} & $\times$ & $\checkmark$ & $\times$ & $\times$ & $\times$ & $\times$ & $\checkmark$ & $\times$ \\*
CLEVRER~\cite{yi2019clevrer} &  & $\times$ & $\checkmark$ & $\times$ & $\times$ & $\times$ & $\times$ & $\checkmark$ & $\times$ \\*
CausalWorld~\cite{ahmed2020causalworld} &  & $\checkmark$ & $\checkmark$ & $\times$ & $\times$ & $\times$ & $\checkmark$ & $\checkmark$ & $\times$ \\*
RoboTwin~\cite{mu2025robotwin} &  & $\checkmark$ & $\checkmark$ & $\times$ & $\times$ & $\times$ & $\checkmark$ & $\checkmark$ & $\times$ \\*
robomimic (simulation subset)~\cite{loop_mandlekar2021demonstrations} &  & $\checkmark$ & $\checkmark$ & $\times$ & $\times$ & $\times$ & $\checkmark$ & $\checkmark$ & $\times$ \\*
AHA~\cite{duan2024aha} &  & $\checkmark$ & $\checkmark$ & $\times$ & $\checkmark$ & $\times$ & $\times$ & $\checkmark$ & $\checkmark$ \\*
RoboFail (simulation subset)~\cite{loop_liu2023reflect} &  & $\checkmark$ & $\checkmark$ & $\times$ & $\checkmark$ & $\times$ & $\checkmark$ & $\checkmark$ & $\checkmark$ \\
\midrule
BAIR robot pushing~\cite{ebert2017selfsupervised} & \multirow[c]{3}{*}{D4} & $\times$ & $\checkmark$ & $\times$ & $\times$ & $\times$ & $\checkmark$ & $\checkmark$ & $\times$ \\*
RoboNet~\cite{dasari2019robonet} &  & $\times$ & $\checkmark$ & $\times$ & $\times$ & $\times$ & $\checkmark$ & $\checkmark$ & $\times$ \\*
AnyPos~\cite{tan2025anypos} &  & $\times$ & $\checkmark$ & $\times$ & $\times$ & $\times$ & $\checkmark$ & $\times$ & $\times$ \\
\midrule
RH20T~\cite{fang2023rh20t} & \multirow[c]{7}{*}{D5} & $\checkmark$ & $\checkmark$ & $\times$ & $\times$ & $\times$ & $\checkmark$ & $\checkmark$ & $\times$ \\*
Open X-Embodiment~\cite{loop_oxe2023} &  & $\checkmark$ & $\checkmark$ & $\times$ & $\times$ & $\times$ & $\checkmark$ & $\checkmark$ & $\times$ \\*
RoboMIND~\cite{wu2024robomind} &  & $\checkmark$ & $\checkmark$ & $\times$ & $\times$ & $\times$ & $\checkmark$ & $\checkmark$ & $\times$ \\*
RoboMIND (fine-grained language subset)~\cite{wu2024robomind} &  & $\checkmark$ & $\checkmark$ & $\times$ & $\checkmark$ & $\times$ & $\checkmark$ & $\checkmark$ & $\times$ \\*
RoboMIND (failure subset)~\cite{wu2024robomind} &  & $\checkmark$ & $\checkmark$ & $\times$ & $\times$ & $\times$ & $\checkmark$ & $\checkmark$ & $\checkmark$ \\*
RH20T-P~\cite{chen2024rh20tp} &  & $\checkmark$ & $\checkmark$ & $\times$ & $\checkmark$ & $\times$ & $\checkmark$ & $\checkmark$ & $\times$ \\*
ShareRobot~\cite{ji2025robobrain} &  & $\checkmark$ & $\checkmark$ & $\times$ & $\checkmark$ & $\checkmark$ & $\times$ & $\checkmark$ & $\times$ \\
\midrule
RoboTurk~\cite{mandlekar2019roboturk} & \multirow[c]{9}{*}{D6} & $\checkmark$ & $\checkmark$ & $\times$ & $\times$ & $\times$ & $\checkmark$ & $\checkmark$ & $\times$ \\*
robomimic (real subset)~\cite{loop_mandlekar2021demonstrations} &  & $\checkmark$ & $\checkmark$ & $\times$ & $\times$ & $\times$ & $\checkmark$ & $\checkmark$ & $\times$ \\*
ALOHA~\cite{zhao2023learning} &  & $\checkmark$ & $\checkmark$ & $\times$ & $\times$ & $\times$ & $\checkmark$ & $\checkmark$ & $\times$ \\*
Mobile ALOHA~\cite{loop_fu2024mobilealoha} &  & $\checkmark$ & $\checkmark$ & $\times$ & $\times$ & $\times$ & $\checkmark$ & $\checkmark$ & $\times$ \\*
BridgeData V2~\cite{walke2023bridgedatav2} &  & $\checkmark$ & $\checkmark$ & $\times$ & $\times$ & $\times$ & $\checkmark$ & $\checkmark$ & $\times$ \\*
DROID~\cite{loop_khazatsky2024droid} &  & $\checkmark$ & $\checkmark$ & $\times$ & $\times$ & $\times$ & $\checkmark$ & $\checkmark$ & $\times$ \\*
ECoT~\cite{zawalski2024ecot} &  & $\checkmark$ & $\checkmark$ & $\checkmark$ & $\checkmark$ & $\checkmark$ & $\checkmark$ & $\checkmark$ & $\times$ \\*
RoboVQA (robot video--QA subset)~\cite{sermanet2023robovqa} &  & $\checkmark$ & $\checkmark$ & $\times$ & $\checkmark$ & $\checkmark$ & $\times$ & $\checkmark$ & $\times$ \\*
RoboFail (real-robot subset)~\cite{loop_liu2023reflect} &  & $\checkmark$ & $\checkmark$ & $\times$ & $\checkmark$ & $\times$ & $\checkmark$ & $\checkmark$ & $\checkmark$ \\
\bottomrule
\end{tabular}
\end{table*}

%% file: sec/10_bench.tex
Evaluating AEI requires more than testing whether a robot can complete a fixed task. Classical embodied benchmarks~\citep{kolve2017ai2, savva2019habitat, james2020rlbench, szot2021habitat} enabled reproducible and comparable evaluation by standardizing environments, action spaces, and task success criteria.  
These benchmarks remain important, but they usually evaluate whether an agent completes a predefined task. This is not sufficient for AEI, where successful task execution often depends on how the agent handles uncertainty and unexpected changes during interaction. Beyond completing the assigned goal, the agent may need to clarify ambiguous instructions, gather missing information, use or create tools, maintain task state over long horizons, detect failures, and revise its plan accordingly. Therefore, AEI evaluation should assess not only the final task outcome, but also the agent's behavior throughout the interaction process.

This section organizes AEI evaluation into three connected layers. First, a simulator defines the world in which agentic behavior can be exercised. For AEI, this world must expose not only geometry and rigid-body motion but also contact, multimodal sensing, persistent object states, and long-horizon state transitions. Second, a benchmark defines the capability being challenged. We therefore organize tasks by their cognitive and physical demands rather than only by surface categories such as navigation or grasping. Third, an evaluation protocol determines how the agent's performance is measured. It should evaluate not only whether the task is completed, but also the quality of the agent's reasoning, its ability to recover from failures, the physical validity of its actions, and its safety. It should also examine whether the results obtained in simulation are consistent with real-world robot performance.

The three layers should be designed together. First, the simulator must contain the physical properties needed by the benchmark. For example, a benchmark cannot test whether an agent knows that a sponge can absorb liquid if the simulator does not model that property. Second, the benchmark must expose the capability we want to measure. Even in a rich simulator, a task that only checks final success cannot tell whether the agent planned well, recovered from mistakes, or simply found a shortcut. Third, the metric should reward behavior that remains meaningful on real robots. A policy should not receive a high score only because it exploits a simulator-specific artifact that would not work in the real world. We therefore view AEI evaluation as an alignment problem among \emph{world expressiveness}, \emph{capability coverage}, and \emph{measurement validity}.

\subsection{Simulation Foundations}
\label{subsec:physics_semantic_sim}

A conventional simulator models how physical states change under actions. For agentic embodied systems, this is not enough: the environment must also preserve task-relevant states and relations when they are temporarily unobserved. We represent this joint state as
\begin{equation}
    (x_{t+1}, z_{t+1}) =
    F(x_t, z_t, a_t; \theta),
\end{equation}
where $x_t$ denotes physical state, $z_t$ denotes semantic states and relations, $a_t$ is the agent action, and $\theta$ contains physical properties such as mass and friction. This formulation highlights four requirements for simulation: modeling diverse physical processes, providing informative sensory feedback, maintaining persistent task state, and supporting scalable evaluation across platforms.

\textbf{Contact dynamics and heterogeneous physical processes.}
Rigid and articulated bodies remain the foundation for manipulation, where collision, friction, joint limits, and stable contact determine whether actions are executable. GPU-parallel platforms have made these interactions more practical to simulate at scale, supporting broader policy training and controlled variation~\citep{tao2025maniskill3,zakka2025mujocoplayground,mittal2025isaaclab}. Everyday tasks, however, also involve deformable objects, fluids, and material changes that rigid-body models cannot represent adequately. Recent simulators extend coverage to these processes, pointing to a broader requirement: simulation fidelity should match the physical effects that determine task feasibility, rather than focus only on processes that are computationally convenient~\citep{cao2026dlolab,huang2026soma,zhu2026flatlab}.

\textbf{Multimodal sensing for active physical inquiry.}
An agent often needs more than visual feedback to determine what happened during contact. Depth, proprioception, force, and tactile signals can reveal properties and interaction outcomes that are hidden from the camera. Simulating these channels allows benchmarks to test whether agents use actions to reduce uncertainty, for example by probing an object's weight or testing whether a surface is slippery. Such actions make sensing part of reasoning rather than a passive input, and prevent benchmarks from avoiding physical inference by omitting relevant feedback~\citep{li2025taccel}.

\textbf{Persistent semantic states and transition rules.}
Many long-horizon tasks depend on states such as whether an object is open, filled, clean, cooked, or broken, as well as relations such as containment and attachment. These states must persist across time and change consistently with physical interaction. Task-rich environments and persistent world models provide complementary foundations for representing such state changes and relations~\citep{li2023behavior1k,nasiriany2026robocasa365,zhou2025persistentwm}. This semantic layer determines which causal effects and plans a simulator can test. It also enables a useful separation in evaluation: the simulator can retain privileged ground truth, while the agent must infer and update its own state from partial observations.

\textbf{Scalable, interoperable simulation ecosystems.}
No single platform can optimize physical fidelity, semantic richness, parallel throughput, and real-world validation for every task. Simulation ecosystems therefore aim to make assets, physics backends, world-state interfaces, and evaluation tools reusable across platforms. Current efforts explore common interfaces across backends, large collections of scenes and assets, and learned interactive models as simulation surrogates~\citep{geng2025roboverse,kim2026molmospaces,wang2026interactiveworld}. This direction supports choosing or combining platforms according to the capability being tested and the evidence required, instead of treating one simulator as a universal solution.

Table~\ref{tab:aei_simulators} compares representative platforms across physical modeling, world and task modeling, scalability, and sim-to-real validation. The comparison shows that their strengths are complementary. High-throughput platforms support large-scale experiments, while task-rich environments represent persistent states and longer activities. Accordingly, platform choice should follow the evaluation claim: the simulator must provide the physical and semantic conditions needed to test that claim, and its results should be validated before they are treated as evidence of real-world performance.

\input{tab/10_simulator}

\begin{figure}[!t]
    \centering
    \includegraphics[width=\linewidth]{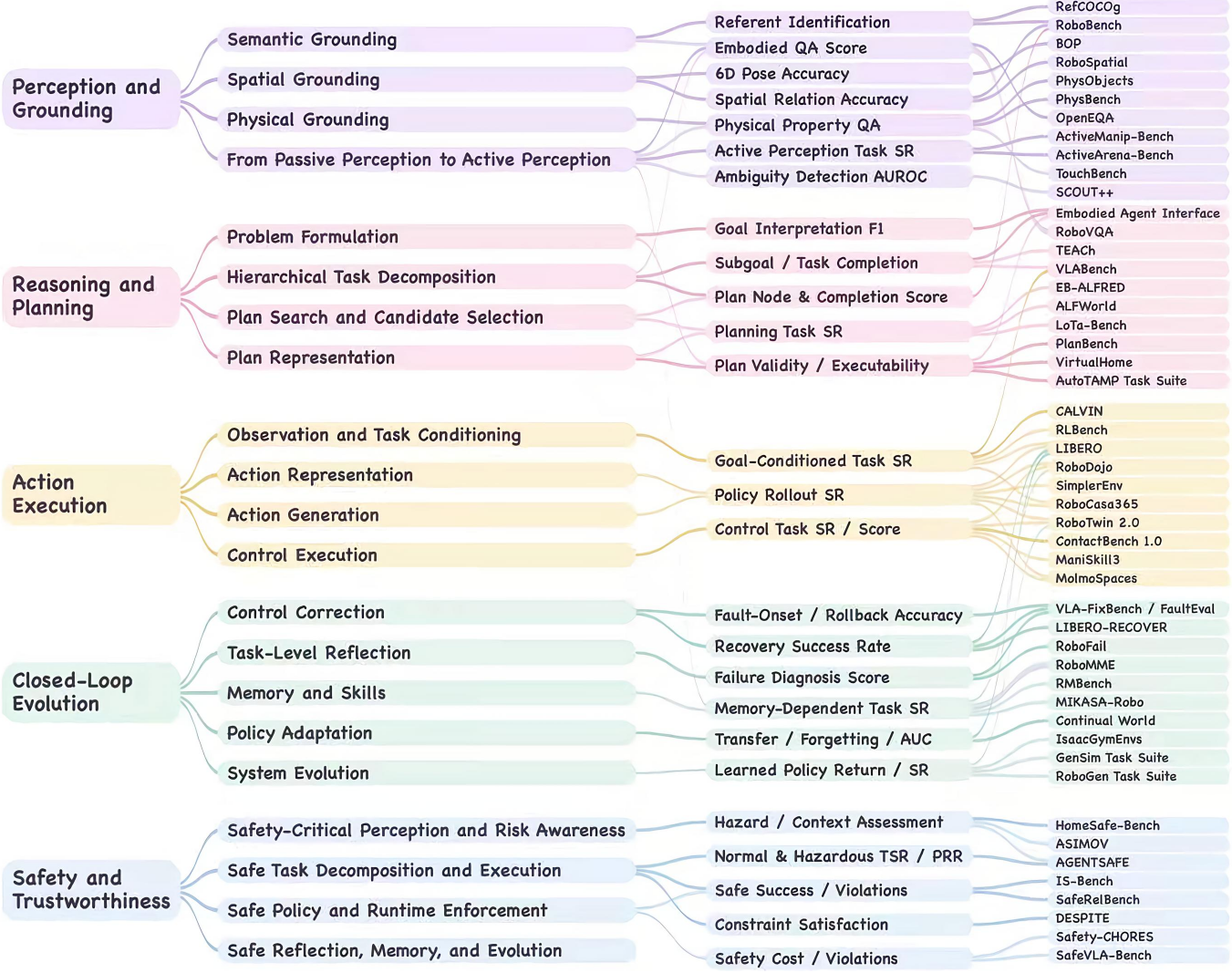}
    \caption{\textbf{The summary of representative benchmarks for the agentic embodied intelligence system.}}
    \label{fig:bench}
\end{figure}

\subsection{Benchmark Spectrum}
\label{subsec:aei_benchmark_ladder}

\begin{figure}[t]
\centering
\includegraphics[width=\linewidth]{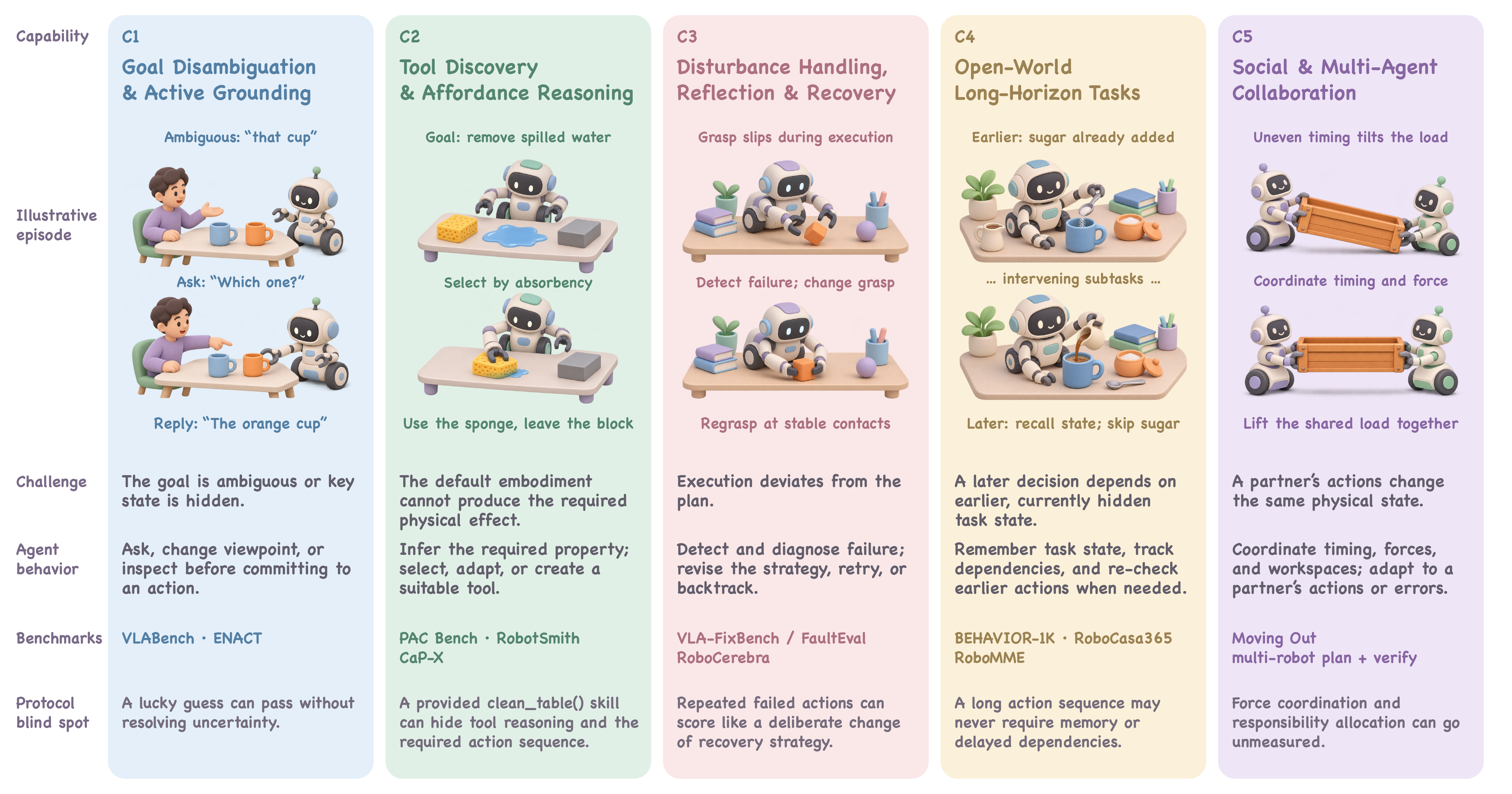}
\caption{The AEI cognitive capability ladder. Each tier is characterized by the \emph{reason} a task is hard rather than by the motor primitive it uses, so that a benchmark result can be attributed to a specific capability. Moving from C1 to C5 increases the information the agent must acquire, the number of intermediate states it must manage, and the extent to which success depends on adaptation rather than direct execution. The bottom row records what a protocol fails to detect when it scores only endpoint success. These are benchmark regimes, not a second autonomy taxonomy; they are distinct from the L0--L5 levels of Table~\ref{tab:embodied_autonomy_levels}.}
\label{fig:aei_capability_ladder}
\end{figure}

We summarize embodied benchmarks in Figure~\ref{fig:bench}, which are often organized by mechanical task type: navigation, grasping, pushing, rearrangement, or locomotion. This taxonomy is useful for controller design, but it is not sufficient for AEI. Two tasks with the same motor primitive can require very different levels of intelligence. Picking up a visible mug from an uncluttered table mainly tests perception and control. Picking a suitable mug when the instruction is ambiguous, the target is occluded, and the agent must first move an obstacle or ask a question tests active perception, uncertainty management, and language grounding. AEI benchmarks should therefore expose the \emph{reason for difficulty}, not only the surface form of the action.
As shown in Figure~\ref{fig:aei_capability_ladder}, we organize this space as a progressive cognitive capability ladder. The tiers increase the amount of information the agent must acquire, the number of intermediate states it must manage, and the extent to which success depends on adaptation rather than direct execution. These benchmarks mark a shift from action-centric evaluation toward capability-centric evaluation. 

\textbf{C1: Goal disambiguation and active physical grounding.}
The first capability is to recognize when current evidence is insufficient and acquire information before committing to an action. Ambiguous references, occlusions, or hidden object states may require a viewpoint change, scene interaction, or clarification rather than a guess. Benchmarks should test whether these actions reduce uncertainty and support a grounded decision. Recent work probes these behaviors through uncertainty-aware help seeking, implicit-intent understanding, and action-conditioned prediction of state changes~\citep{ren2023robotsask,zhang2025vlabench,wang2026enact}. Active grounding thus links partial observability to information-seeking behavior and evidence-based action.

\textbf{C2: Tool discovery, affordance reasoning, and physical reachability creation.}
The second capability is to extend the robot's effective affordances when its default embodiment cannot produce the required physical effect. The agent must reason backward from the goal to relevant object properties and physical constraints, then select, adapt, or create a tool. Recent benchmarks probe these competencies through affordance reasoning, tool synthesis, and executable skill composition~\citep{gundawar2025pacbench,lin2025robotsmith,fu2026capx}. Evaluation should distinguish such competence from capabilities supplied by hand-designed primitives: a provided \texttt{clean-table()} skill, for example, bypasses the tool selection, wiping, and regrasping decisions that the task is intended to test.

\textbf{C3: Online disturbance handling, reflection, and recovery.}
A defining property of closed-loop agency is the ability to detect and recover from deviations that arise during execution, not merely tolerate randomized initial conditions. Benchmarks should perturb ongoing tasks and assess whether an agent recognizes the failure, diagnoses its source, and selects an effective retry or rollback. Recent work begins to evaluate this process across perception, planning, control, and long-horizon reflection~\citep{yan2026vlafix,han2025robocerebra}. Scoring should distinguish genuine diagnosis and recovery from repeated attempts that happen to reach the goal.


\textbf{C4: Open-world, long-horizon ecological tasks.}
A meaningful long-horizon benchmark tests more than the number of actions: it requires agents to retain temporarily hidden state, track subtask dependencies, and use earlier events to handle delayed consequences and exceptions. Ecological task suites and history-dependent manipulation benchmarks provide complementary settings for evaluating these demands, including memory for temporal, spatial, object, and procedural information~\citep{li2023behavior1k,nasiriany2026robocasa365,dai2026robomme}. The central criterion is whether the agent maintains and uses a coherent task state, rather than merely executing a longer sequence.


\textbf{C5: Physical social and multi-agent collaboration.}
At this tier, success depends on partners whose actions alter a shared physical state, requiring coordination under behavioral uncertainty, coupled workspace and timing constraints. Initial benchmarks and systems explore physically grounded human--robot collaboration and multi-robot execution, but systematic coverage remains limited~\citep{kang2026movingout,he2026multiagent}. Important open evaluation targets include implicit coordination, shared uncertainty, responsibility allocation, and recovery from partner errors, making collaboration a frontier of AEI benchmarking rather than a mature capability category.

\subsection{Evaluation Metrology}
\label{subsec:decoupled_metrology}

A task-success rate records whether the goal was reached, but not whether the agent made feasible decisions, recovered from errors, or acted safely. AEI evaluation should therefore characterize the full interaction trajectory through goal completion, cognitive rationality, recovery, physical execution, and safety violations:
\begin{equation}
    \mathbf{M}(\tau)=
    \left[
    G(\tau),
    M_{\mathrm{cog}}(\tau),
    M_{\mathrm{rec}}(\tau),
    M_{\mathrm{phy}}(\tau),
    M_{\mathrm{safe}}(\tau)
    \right],
\end{equation}
where $\tau$ is the interaction trajectory and $G(\tau)$ indicates task completion. These dimensions distinguish planning errors from control failures, and graded execution quality from violations of hard safety constraints. Evidence from safety-aware and recovery-focused evaluations reinforces the need to assess the process as well as its outcome~\citep{bai2026safelab,balaji2026oopsieverse,yan2026vlafix,han2025robocerebra}. Table~\ref{tab:metric_matrix} summarizes representative metrics and the questions they address. Metrics should be selected to match the benchmark claim, with the action interface and tool budget reported so that supplied capabilities are not mistaken for agent reasoning~\citep{fu2026capx}.


\textbf{Evidence and scoring validity.}
Metrics are only meaningful when supported by evidence suited to the property being assessed. Simulator state and calibrated sensors can verify objective conditions such as contact, containment, force, and temperature. Semantic qualities that lack explicit state predicates may require learned or human judgment. Learned judges can scale evaluation, but should be calibrated and audited against human assessments rather than treated as ground truth~\citep{jangir2026robotarenainf,atreya2025roboarena}. Benchmarks must also distinguish goal completion from physically valid completion. A successful outcome should not count as valid if it depends on unsafe behavior or simulator-specific shortcuts:
\begin{equation}
    S_{\mathrm{valid}}(\tau)
    =
    G(\tau)\,
    \mathbf{1}\!\left[
        C_{\mathrm{physics}}(\tau)
        \wedge
        C_{\mathrm{safety}}(\tau)
    \right].
\end{equation}
The predicates, privileged evaluator signals, and known simulator limitations should be disclosed. Valid success should be reported alongside goal completion and safety violations, so that a combined score does not hide why an episode failed~\citep{bai2026safelab,balaji2026oopsieverse}.

\begin{table}[!t]
\centering
\small
\caption{\textbf{Decoupled metric matrix for AEI evaluation. }Metrics should be selected according to the benchmark claim.}
\label{tab:metric_matrix}
\renewcommand{\arraystretch}{1.2} 
\begin{tabularx}{\textwidth}{>{\raggedright\arraybackslash}p{0.18\textwidth} >{\raggedright\arraybackslash}p{0.40\textwidth} >{\raggedright\arraybackslash}X}
\toprule
\textbf{Dimension} & \textbf{Example metrics} & \textbf{What the metrics diagnose} \\
\midrule
\rowcolor{AEIrow}
Goal attainment
& Success rate; partial completion; terminal-state validity
& Whether the requested outcome was achieved. \\
Cognitive rationality
& Plan feasibility; impossible-action rate; subgoal-order errors; dead-end frequency
& Whether decisions respect task preconditions, physical constraints, and the action interface. \\
\rowcolor{AEIrow}
Reflexivity and recovery
& Fault-detection recall; diagnosis accuracy; recovery success and latency; repeated-error rate
& Whether the agent recognizes and repairs deviations at reasonable cost. \\
Physical compliance
& Jerk; contact-force margin; trajectory and energy cost; collision severity; object damage
& Whether execution is physically controlled and efficient. \\
\rowcolor{AEIrow}
Hard safety boundaries
& Forbidden-region, force, speed, or proximity violations; irreversible hazards
& Whether the trajectory remains within declared safety and authority constraints. \\
\bottomrule
\end{tabularx}
\end{table}

\textbf{Reliable comparison and transfer.}
Physical evaluation is costly, so policy comparisons should use repeated and task-stratified trials, report uncertainty, and apply paired or sequential analysis when appropriate. Automated evaluation can reduce the cost of collecting results, but does not remove the need to quantify uncertainty~\citep{snyder2026beyondbinary,zhou2025autoeval}. Simulation can extend evaluation only when it provides evidence about real-world performance. Existing sim-to-real studies examine whether policy rankings persist across settings and how performance changes with visual, control, and initialization differences~\citep{li2025simpler,jain2026polaris,yang2026robolab,kim2026molmospaces}. Rank agreement is useful, but a stronger validation also checks score calibration and whether simulation reproduces real failure modes. The appropriate standard therefore depends on the claim: a simulator used to rank policies need not support the same conclusions as one used to diagnose failures.

Together, these principles make evaluation claim-driven. Define the capability being tested, select trajectory-level measures for it, ground those measures in suitable evidence, and establish that comparisons are reliable and relevant to physical deployment.

%% file: tab/10_simulator.tex
\begin{table*}[!t]
\centering
\small

\setlength{\tabcolsep}{2pt}
\renewcommand{\arraystretch}{1.0}

\caption{
\textbf{Capability comparison of representative simulation platforms.}
}
\label{tab:aei_simulators}

\begin{tabularx}{\textwidth}{@{}
>{\raggedright\arraybackslash}m{0.19\textwidth}
*{7}{>{\centering\arraybackslash}X}
@{}}

\specialrule{\heavyrulewidth}{\abovetopsep}{0pt}


\multirow[c]{2}{0.19\textwidth}[-0.2ex]{%
    \centering
    \small\bfseries
    System
}
&
\multicolumn{2}{c}{%
    \cellcolor{AEIphysical}
    \AEIgroup{Physical Modeling}
}
&
\multicolumn{2}{c}{%
    \cellcolor{AEIworld}
    \AEIgroup{World \& Task Modeling}
}
&
\multicolumn{2}{c}{%
    \cellcolor{AEIscale}
    \AEIgroup{Scalability}
}
&
\cellcolor{AEIvalid}
\AEIgroup{Validation}
\\[-0.2ex]


&
\cellcolor{AEIphysical}
\AEIsub{Rigid /}{Articulated}
&
\cellcolor{AEIphysical}
\AEIsub{Deformable}{/ Fluid}
&
\cellcolor{AEIworld}
\AEIsub{Semantic}{States}
&
\cellcolor{AEIworld}
\AEIsub{Long-Horizon}{Tasks}
&
\cellcolor{AEIscale}
\AEIsub{Parallel}{Simulation}
&
\cellcolor{AEIscale}
\AEIsub{Cross-}{Simulator}
&
\cellcolor{AEIvalid}
\AEIsub{Sim-to-Real}{Validation}
\\[-0.2ex]

\midrule


\rowcolor{AEIrow}
OmniGibson~\cite{li2023behavior1k}
&
\cmark
&
\cmark
&
\cmark
&
\cmark
&
\pmark
&
\xmark
&
\pmark
\\


ManiSkill3~\cite{tao2025maniskill3}
&
\cmark
&
\xmark
&
\xmark
&
\pmark
&
\cmark
&
\xmark
&
\pmark
\\


\rowcolor{AEIrow}
MuJoCo~\cite{zakka2025mujocoplayground}
&
\cmark
&
\xmark
&
\xmark
&
\xmark
&
\cmark
&
\xmark
&
\cmark
\\


Isaac Lab~\cite{mittal2025isaaclab}
&
\cmark
&
\pmark
&
\xmark
&
\pmark
&
\cmark
&
\xmark
&
\pmark
\\


\rowcolor{AEIrow}
RoboVerse~\cite{geng2025roboverse}
&
\cmark
&
\pmark
&
\pmark
&
\pmark
&
\pmark
&
\cmark
&
\pmark
\\


RoboCasa365~\citep{nasiriany2026robocasa365}
&
\cmark
&
\xmark
&
\pmark
&
\cmark
&
\pmark
&
\xmark
&
\xmark
\\


\rowcolor{AEIrow}
MolmoSpaces~\citep{kim2026molmospaces}
&
\cmark
&
\pmark
&
\pmark
&
\cmark
&
\pmark
&
\cmark
&
\cmark
\\

\bottomrule

\end{tabularx}

\vspace{3pt}

\footnotesize
\cmark~Explicit support
\qquad
\pmark~Partial / backend-dependent
\qquad
\xmark~Not a primary feature

\end{table*}

%% file: sec/11_future.tex
The open challenges of agentic embodied intelligence are connected rather than independent. An agent must first form a trustworthy belief about the physical world, use it to produce a plan that available policies can execute, and act within physical and computational limits. Execution outcomes must then support targeted and safe improvement. Computing infrastructure and data enable this loop, safety constrains it throughout, and evaluation determines whether the resulting capability is reliable beyond isolated demonstrations.

\subsection{Perception and Grounding}

The central challenge in agentic perception is to maintain a task-relevant belief that is calibrated, spatially consistent, physically informative, and valid as the environment changes. Current models remain unreliable about uncertainty, robot-centric geometry, contact properties, and whether previously observed facts are still current, limiting their ability to decide when to act, gather more evidence, or abstain~\citep{vsibench,mmsi,comfort,mvrobobench,physbench,physion}. Active sensing further requires the agent to weigh the expected decision value of looking, touching, or asking against their time, effort, and physical risk. Progress therefore depends on beliefs that expose their evidence and uncertainty, support inspection by downstream policies, and are evaluated for task-dependent accuracy, calibration, temporal validity, and information-gathering cost. Such evaluation must also rule out language-only shortcuts and test whether beliefs remain useful after unobserved changes~\citep{blindtest,ren2023robotsask,abstaineqa,openeqa,dai2026robomme,geminirobotics,yang2025embodiedbench,wang2026enact}.

\subsection{Reasoning and Planning}

Long-horizon planning remains unreliable when search, task decomposition, plan representation, and motor-policy capability are treated separately. Search can amplify inaccurate outcome predictions or prune information-gathering actions; subgoals can exceed the capabilities of available policies; and translation between language, programs, and skills can lose preconditions or termination conditions~\citep{hu2024treeplanner,du2023vlp,chen2026parallelworld,ahn2022saycan,lin2023text2motion,huang2026roboharness,loop_liang2022codepolicies,chen2024autotamp,huang2024rekep}. The key challenge is to couple these decisions: allocate search according to uncertainty, generate subgoals from policy-conditioned feasibility and outcome estimates, and preserve checkable task contracts across representations. Evaluation should isolate planning quality from differences in motor policies, observation access, tool availability, and computation budget~\citep{li2024eai}.

\subsection{Action Execution}

Action execution must generalize across contact-rich tasks, robot embodiments, and deployment constraints while remaining stable and timely. Generalist policies still lean toward simpler gripper-based manipulation, while dexterous hands introduce higher-dimensional actions and more complex contact dynamics; transferring behavior across robots additionally requires representations that preserve both shared interaction structure and embodiment-specific limits~\citep{loop_brohan2022rt1,kim2024openvla,loop_octo2024,wen2025dexvla,xu2025dexumi,yuan2025crossdex,wei2024dro,fei2025t,wu2026cedex}. At the same time, large policies must meet real-time latency and resource budgets on physical platforms. The open challenge is to develop contact-aware policies and action representations that transfer with limited adaptation, and execution pipelines that reduce latency without sacrificing control quality~\citep{pertsch2025fast,kim2025fine,shukor2025smolvla,black2025rtc}.

\subsection{Closed-Loop Evolution}

The central challenge in closed-loop evolution is to show that physical experience produces targeted, transferable improvements rather than merely a successful retry. This requires diagnosing the cause of a failure, choosing the appropriate update target, and independently validating the change: a memory or skill must remain applicable on new tasks, while model and system updates must improve held-out performance without erasing earlier capabilities. Existing work addresses parts of this process, including software-task verification, robot failure correction, skill reuse, and policy adaptation, but no single protocol establishes open-ended physical self-improvement~\citep{loop_jimenez2024swebench,loop_liu2023reflect,loop_wang2023voyager,loop_bousmalis2023robocat}. Future evaluations should match interaction and computation budgets, distinguish generation from validation and held-out testing, track regressions and safety violations across versions, and account for human labor, resets, and hardware time. Safe evolution ultimately requires provenance, applicability checks, independent regression tests, and a rollback path.

\subsection{Computing Infrastructure}
\label{sec:infra-summary}

The infrastructure challenge shifts from efficiently running one model to coordinating models, tools, memories, and robots across device, edge, and cloud. These components operate at different timescales: semantic deliberation may tolerate network delay, whereas control and safety responses require predictable low latency. Future systems must therefore schedule computation dynamically under changing energy, bandwidth, and workload conditions, preserve real-time semantics across agent--robot interfaces, and degrade safely during network loss. This requires deployment protocols that connect high-level tool orchestration to existing control infrastructure, as well as representative workload measurements to guide hardware and runtime design. Infrastructure should be evaluated as part of the deployed agent, not treated as an invisible source of compute.

\subsection{Safety and Trustworthiness}

The central safety challenge is to turn useful local safeguards into evidence for the behavior of the complete embodied loop. Risk must be grounded in physical state and uncertainty, planning must preserve safety preconditions and recovery obligations, runtime mechanisms must constrain unsafe actions, and persistent memories or updates must not propagate stale or unsafe assumptions~\citep{yang2024safetychip,ren2023robotsask,lu2026homeguard,sima2026jitmemory,zhang2025safevla}. These protections do not automatically compose into an end-to-end guarantee. Progress depends on explicit contracts between perception, planning, execution, and learning, backed by independent testing across embodiments, long horizons, and real deployment conditions. Until such evidence is routine, safety claims should remain bounded by their stated assumptions and operating domains.

\subsection{Agentic Data Ecosystem}
\label{sec:agentic_data_challenges_horizons}

The data challenge is to turn heterogeneous observations and actions into traceable experience that links goals, decisions, execution, feedback, and outcomes. Human video, robot trajectories, generated rationales, and failure annotations differ in embodiment, temporal resolution, and quality; without aligned events, entity references, and provenance, errors can propagate through policy updates and future data selection. Meanwhile, costly and safety-constrained physical collection leaves rare failures and complete recovery episodes underrepresented, while simulation introduces transfer gaps~\citep{dalal2018safe,brunke2022safe,yan2026vlafix,balaji2026oopsieverse}. A useful direction is an event-linked experience record with versioned labels and independently checked outcomes, complemented by data collection that targets failure and recovery rather than familiar successes alone. For L4, records must capture long-horizon operation, domain-boundary events, and safe halting; for L5, they must extend to unfamiliar situations, skill transfer, and longitudinal capability change.

\subsection{Simulators, Benchmarks, and Evaluation Protocols}
\label{sec:bench-open-challenges}

Evaluation must progress from measuring success on resettable episodes to establishing the conditions under which autonomy can be granted. For L4, benchmarks should specify the operational domain and measure boundary recognition, safe stopping, and the cost of unnecessary escalation; for longer-term autonomy claims, protocols must include uninterrupted operation, accumulated damage, recovery, and regressions after system updates~\citep{li2026saewc,abstaineqa,seo2025uncertaintyfilter,sentinel,rajaprakash2026safecast,kim2026safelonghorizon,li2023behavior1k,nasiriany2026robocasa365,dai2026robomme,jiang2026rollout,qin2026embodiedgovbench}. For L5, fixed task suites are insufficient, motivating generated and adversarial tasks whose physical solvability and diagnostic value are independently checked~\citep{he2026tea,jiang2026benchforge,lai2026benchautomation,zheng2026benchbench,nguyen2025regen,kang2026movingout,he2026multiagent}. Because rare failures cannot be established by nominal sampling alone, evidence should combine boundary-focused simulation, failure-mode transfer checks, sequential real-robot testing, and explicit residual-risk reporting; simulator error and evaluator limitations must themselves be accounted for~\citep{li2025simpler,jain2026polaris,yang2026robolab,wang2026interactiveworld,zhou2025persistentwm,badithela2026imperfect,wan2026nofreechecker,snyder2026beyondbinary}. Thus, evaluation is not merely a final score: it is the evidence layer that determines whether a system's autonomy claim is justified.

%% file: sec/12_conclusion.tex
This survey has examined agentic embodied intelligence as a system-level transition from conditional action generation toward sustained, feedback-driven physical behavior. The central requirement is not simply to produce a plausible action sequence, but to maintain a task-relevant account of the world, reason about goals and constraints over time, execute through policies and control interfaces, and use observed outcomes to continue, recover, or revise the plan. Perception and grounding, reasoning and planning, action execution, and closed-loop evolution are therefore best understood as coupled functions linked by explicit information and control interfaces. 
Their performance depends on the surrounding infrastructure that schedules deliberation and control, mediates skill and tool calls, preserves state, and handles communication and exceptions. It also depends on data that connect goals and decisions to physical execution, feedback, and correction, and on evaluation that tests recovery, uncertainty, safety, and transfer rather than final success alone.

The autonomy ladder clarifies the remaining gap. Teleoperation, assisted primitives, and bounded single-task policies are established operating modes; language-conditioned foundation policies extend capability but remain conditional on their training and operating conditions. L4 requires sustained mission execution within a declared operational domain, including verification, routine recovery, boundary recognition, and safe escalation. The literature reviewed here does not establish sustained unattended L4 operation across a broad, independently validated domain, much less L5 generality. An agentic closed loop is a necessary systems condition for progressing beyond conditional autonomy.

The path from foundation models to physical agency is therefore a joint challenge in learning, control, infrastructure, data, and assurance. Model scale can broaden what an agent recognizes or attempts; dependable embodied autonomy will depend on whether the complete system can ground its decisions, act within physical limits, detect when reality diverges from expectation, and respond appropriately. Agentic embodied intelligence offers a framework for organizing that work and for measuring progress toward useful, bounded autonomy and, ultimately, more general physical agency.

%% file: main.bib
@article{ashby1957introduction,
  title={An introduction to cybernetics},
  author={Ashby, W Ross and Pierce, JR},
  journal={Physics Today},
  year={1957}}

@book{moravec1988mind,
  title={Mind children: The future of robot and human intelligence},
  author={Moravec, Hans},
  year={1988},
  publisher={Harvard University Press}
}

@article{brooks1991intelligence,
  title={Intelligence without representation},
  author={Brooks, Rodney A},
  journal={Artificial intelligence},
  year={1991},
}

@book{varela2017embodied,
  title={The embodied mind: Cognitive science and human experience},
  author={Varela, Francisco J and Thompson, Evan and Rosch, Eleanor},
  year={2017},
  publisher={The MIT press}
}

@book{clark1997being,
  title={Being there: Putting brain, body, and world together again},
  author={Clark, Andy and Boden, Margaret A},
  year={1997},
  publisher={MIT press Cambridge, MA}
}

@article{barsalou1999perceptual,
  title={Perceptual symbol systems.},
  author={Barsalou, Lawrence W},
  journal={Behavioral \& Brain Sciences},
  year={1999}
}

@article{lakoff1999review,
  title={Review of Philosophy in the Flesh: The embodied mind and its challenge to Western thought},
  author={Lakoff, George and Johnson, Mark and Sowa, John F},
  journal={Computational Linguistics},
  year={1999}
}

@book{gibson2014ecological,
  title={The ecological approach to visual perception: classic edition},
  author={Gibson, James J},
  year={2014},
  publisher={Psychology press}
}

@article{lungarella2003developmental,
  title={Developmental robotics: a survey},
  author={Lungarella, Max and Metta, Giorgio and Pfeifer, Rolf and Sandini, Giulio},
  journal={Connection science},
  year={2003},
}

@article{zhao2023learning,
  title={Learning fine-grained bimanual manipulation with low-cost hardware},
  author={Zhao, Tony Z and Kumar, Vikash and Levine, Sergey and Finn, Chelsea},
  journal={arXiv preprint arXiv:2304.13705},
  year={2023}
}

@article{argall2009survey,
  title={A survey of robot learning from demonstration},
  author={Argall, Brenna D and Chernova, Sonia and Veloso, Manuela and Browning, Brett},
  journal={Robotics and autonomous systems},
  year={2009},
}

@article{ijspeert2013dynamical,
  title={Dynamical movement primitives: learning attractor models for motor behaviors},
  author={Ijspeert, Auke Jan and Nakanishi, Jun and Hoffmann, Heiko and Pastor, Peter and Schaal, Stefan},
  journal={Neural computation},
  year={2013},
}

@article{mahler2017dex,
  title={Dex-net 2.0: Deep learning to plan robust grasps with synthetic point clouds and analytic grasp metrics},
  author={Mahler, Jeffrey and Liang, Jacky and Niyaz, Sherdil and Laskey, Michael and Doan, Richard and Liu, Xinyu and others},
  journal={arXiv preprint arXiv:1703.09312},
  year={2017}
}

@inproceedings{kuffner2000rrt,
  title={RRT-connect: An efficient approach to single-query path planning},
  author={Kuffner, James J and LaValle, Steven M},
  booktitle={ICRA},
  year={2000},
}

@article{levine2016end,
  title={End-to-end training of deep visuomotor policies},
  author={Levine, Sergey and Finn, Chelsea and Darrell, Trevor and Abbeel, Pieter},
  journal={Journal of Machine Learning Research},
  year={2016}
}

@inproceedings{kalashnikov2018scalable,
  title={Scalable deep reinforcement learning for vision-based robotic manipulation},
  author={Dmitry Kalashnikov and Alex Irpan and Peter Pastor and Julian Ibarz and Alexander Herzog and Eric Jang and others},
  booktitle={CoRL},
  year={2018},
}

@inproceedings{fang2020graspnet,
  title={Graspnet-1billion: A large-scale benchmark for general object grasping},
  author={Fang, Hao-Shu and Wang, Chenxi and Gou, Minghao and Lu, Cewu},
  booktitle={CVPR},
  year={2020},
}

@inproceedings{zeng2021transporter,
  title={Transporter networks: Rearranging the visual world for robotic manipulation},
  author={Andy Zeng and Pete Florence and Jonathan Tompson and Stefan Welker and Jonathan Chien and Maria Attarian and others},
  booktitle={CoRL},
  year={2021},
}

@article{chi2025diffusion,
  title={Diffusion policy: Visuomotor policy learning via action diffusion},
  author={Chi, Cheng and Xu, Zhenjia and Feng, Siyuan and Cousineau, Eric and Du, Yilun and Burchfiel, Benjamin and others},
  journal={The International Journal of Robotics Research},
  year={2025},
}

@article{akkaya2019solving,
  title={Solving rubik's cube with a robot hand},
  author={{OpenAI} and Ilge Akkaya and Marcin Andrychowicz and Maciek Chociej and Mateusz Litwin and Bob McGrew and others},
  journal={arXiv preprint arXiv:1910.07113},
  year={2019}
}

@article{kim2024openvla,
  title={Openvla: An open-source vision-language-action model, 2024},
  author={Moo Jin Kim and Karl Pertsch and Siddharth Karamcheti and Ted Xiao and Ashwin Balakrishna and Suraj Nair and others},
  journal={arXiv preprint arXiv:2406.09246},
  year={2024}
}

@article{black2024pi_0,
  title={$pi\_0$: A Vision-Language-Action Flow Model for General Robot Control},
  author={Kevin Black and Noah Brown and Danny Driess and Adnan Esmail and Michael Equi and Chelsea Finn and others},
  journal={arXiv preprint arXiv:2410.24164},
  year={2024}
}

@article{hu2024video,
  title={Video prediction policy: A generalist robot policy with predictive visual representations},
  author={Hu, Yucheng and Guo, Yanjiang and Wang, Pengchao and Chen, Xiaoyu and Wang, Yen-Jen and Zhang, Jianke and others},
  journal={arXiv preprint arXiv:2412.14803},
  year={2024}
}

@article{kim2026cosmos,
  title={Cosmos policy: Fine-tuning video models for visuomotor control and planning},
  author={Moo Jin Kim and Yihuai Gao and Tsung-Yi Lin and Yen-Chen Lin and Yunhao Ge and Grace Lam and others},
  journal={arXiv preprint arXiv:2601.16163},
  year={2026}
}

@article{pai2025mimic,
  title={mimic-video: Video-action models for generalizable robot control beyond vlas},
  author={Pai, Jonas and Achenbach, Liam and Montesinos, Victoriano and Forrai, Benedek and Mees, Oier and Nava, Elvis},
  journal={arXiv preprint arXiv:2512.15692},
  year={2025}
}

@article{li2026causal,
  title={Causal world modeling for robot control},
  author={Li, Lin and Zhang, Qihang and Luo, Yiming and Yang, Shuai and Wang, Ruilin and Han, Fei and others},
  journal={arXiv preprint arXiv:2601.21998},
  year={2026}
}

@article{wang2026openwam,
  title={OpenWAM: An open, modular exploration towards systematic world-action model pretraining},
  author={Yuran Wang and Siqiao Huang and Mingleyang Li and Chenhao Zhang and Jiaqi Liang and Weiyang Jin and others},
  journal={arXiv preprint arXiv:2609.07398},
  year={2026}
}

@misc{openai2026gpt,
  author       = {{OpenAI}},
  title        = {GPT-6 Astra},
  howpublished = {\url{https://openai.com/zh-Hans-CN/index/gpt-6-astra/}},
  year         = {2026},
}

@misc{anthropic2026opus,
  author       = {{Anthropic}},
  title        = {Claude Opus 5.5},
  howpublished = {\url{https://www.anthropic.com/claude-opus-5-5}},
  year         = {2026},
}

@article{zhang2026harness,
  title={Harness VLA: Steering Frozen VLAs into Reliable Manipulation Primitives via Memory-Guided Agents},
  author={Yixian Zhang and Huanming Zhang and Feng Gao and Xiao Li and Zhihao Liu and Yi Nie and others},
  journal={arXiv preprint arXiv:2607.08448},
  year={2026}
}

@misc{ding2026zetta,
  title={Zetta $\zeta$: An Efficient Closed-Loop Embodied Harness for Self-Evolving Physical Intelligence}, 
  author={Xin Ding and Liang Mi and Mingzhe Huang and Zixuan Wang and Chao Zhang and Zixu Hao and others},
  journal={arXiv preprint arXiv:2608.16590},
  year={2026},
}

@article{abbeel2010autonomous,
  title={Autonomous helicopter aerobatics through apprenticeship learning},
  author={Abbeel, Pieter and Coates, Adam and Ng, Andrew Y},
  journal={The International Journal of Robotics Research},
  year={2010},
}

@article{kober2008policy,
  title={Policy search for motor primitives in robotics},
  author={Kober, Jens and Peters, Jan},
  journal={NeurIPS},
  year={2008}
}

@inproceedings{ross2011reduction,
  title={A reduction of imitation learning and structured prediction to no-regret online learning},
  author={Ross, St{\'e}phane and Gordon, Geoffrey and Bagnell, Drew},
  booktitle={AISTATS},
  year={2011},
}

@article{levine2018learning,
  title={Learning hand-eye coordination for robotic grasping with deep learning and large-scale data collection},
  author={Levine, Sergey and Pastor, Peter and Krizhevsky, Alex and Ibarz, Julian and Quillen, Deirdre},
  journal={The International journal of robotics research},
  year={2018}}

@inproceedings{pinto2016supersizing,
  title={Supersizing self-supervision: Learning to grasp from 50k tries and 700 robot hours},
  author={Pinto, Lerrel and Gupta, Abhinav},
  booktitle={ICRA},
  year={2016},
}

@inproceedings{tobin2017domain,
  title={Domain randomization for transferring deep neural networks from simulation to the real world},
  author={Tobin, Josh and Fong, Rachel and Ray, Alex and Schneider, Jonas and Zaremba, Wojciech and Abbeel, Pieter},
  booktitle={IROS},
  year={2017},
}

@article{garcia2015safe,
  author  = {Garc{\'i}a, Javier and Fern{\'a}ndez, Fernando},
  title   = {A Comprehensive Survey on Safe Reinforcement Learning},
  journal = {Journal of Machine Learning Research},
  year    = {2015}
}

@article{brunke2022safe,
  author  = {Brunke, Lukas and Greeff, Melissa and Hall, Adam W. and Yuan, Zhaocong and Zhou, Siqi and Panerati, Jacopo and others},
  title   = {Safe Learning in Robotics: From Learning-Based Control to Safe Reinforcement Learning},
  journal = {Annual Review of Control, Robotics, and Autonomous Systems},
  year    = {2022},
}

@article{li2026embodiedsurvey,
  author  = {Xiao Li and Xiang Zheng and Yifeng Gao and Xinyu Xia and Yixu Wang and Xin Wang and others},
  title   = {Safety in Embodied AI: A Survey of Risks, Attacks, and Defenses},
  journal = {arXiv preprint arXiv:2605.02900},
  year    = {2026}
}

@inproceedings{achiam2017cpo,
  author    = {Achiam, Joshua and Held, David and Tamar, Aviv and Abbeel, Pieter},
  title     = {Constrained Policy Optimization},
  booktitle = {ICML},
  year      = {2017}
}

@inproceedings{dalal2018safe,
  author    = {Dalal, Gal and Dvijotham, Krishnamurthy and Vecerik, Matej and Hester, Todd and Paduraru, Cosmin and Tassa, Yuval},
  title     = {Safe Exploration in Continuous Action Spaces},
  journal={arXiv preprint arXiv:1801.08757},
  year      = {2018}
}

@inproceedings{cheng2019barrier,
  author    = {Cheng, Richard and Orosz, G{\'a}bor and Murray, Richard M. and Burdick, Joel W.},
  title     = {End-to-End Safe Reinforcement Learning through Barrier Functions for Safety-Critical Continuous Control},
  booktitle = {AAAI},
  year      = {2019},
}

@inproceedings{ames2019cbf,
  author    = {Ames, Aaron D. and Coogan, Samuel and Egerstedt, Magnus and Notomista, Gennaro and Sreenath, Koushil and Tabuada, Paulo},
  title     = {Control Barrier Functions: Theory and Applications},
  booktitle = {ECC},
  year      = {2019},
}

@article{wabersich2021psf,
  author  = {Wabersich, Kim P. and Zeilinger, Melanie N.},
  title   = {A Predictive Safety Filter for Learning-Based Control of Constrained Nonlinear Dynamical Systems},
  journal = {Automatica},
  year    = {2021},
}

@article{althoff2014reachability,
  author  = {Althoff, Matthias and Dolan, John M.},
  title   = {Online Verification of Automated Road Vehicles Using Reachability Analysis},
  journal = {IEEE Transactions on Robotics},
  year    = {2014},
}

@inproceedings{ren2023robotsask,
  author    = {Ren, Allen Z. and Dixit, Anushri and Bodrova, A. A. and Singh, Sumeet and Tu, Stephen and Brown, Noah and others},
  title     = {Robots That Ask for Help: Uncertainty Alignment for Large Language Model Planners},
  booktitle = {CoRL},
  year      = {2023}
}

@inproceedings{yang2024safetychip,
  author    = {Yang, Ziyi and Raman, Shreyas Sundara and Shah, Ankit and Tellex, Stefanie},
  title     = {Plug in the Safety Chip: Enforcing Constraints for LLM-Driven Robot Agents},
  booktitle = {ICRA},
  year      = {2024},
}

@misc{sermanet2025constitutions,
  author       = {Sermanet, Pierre and Majumdar, Anirudha and Irpan, Alex and Kalashnikov, Dmitry and Sindhwani, Vikas},
  title        = {Generating Robot Constitutions and Benchmarks for Semantic Safety},
  journal={arXiv preprint arXiv:2503.08663},
  year         = {2025}
}

@misc{lu2026homeguard,
  author       = {Lu, Xiaoya and Zhou, Yijin and Chen, Zeren and Wang, Ruocheng and Sima, Bingrui and Zhou, Enshen and others},
  title        = {HomeGuard: VLM-Based Embodied Safeguard for Identifying Contextual Risk in Household Task},
  howpublished = {Lecture Notes in Computer Science},
  year         = {2026},
}

@misc{son2025embodyguard,
  author       = {Son, Yejin and Kim, Minseo and Kim, Sungwoong and Han, Seungju and Kim, Jian and Jang, Dongju and others},
  title        = {Subtle Risks, Critical Failures: A Framework for Diagnosing Physical Safety of LLMs for Embodied Decision Making},
  howpublished = {EMNLP},
  year         = {2025}
}

@misc{yang2026saferelbench,
  author       = {Yang, Huaigang and Li, Ya and Ren, Min and Dai, Bo and Zhang, Zhenliang and He, Zhaofeng},
  title        = {SafeRelBench: A Spatial-Relation-Aware Benchmark for Process-Level Safety in VLM-Driven Embodied Agents},
  journal={arXiv preprint arXiv:2607.14543},
  year         = {2026}
}

@misc{li2026safetask,
  author       = {Li, Siyuan and Lang, Taiyan and Yan, Aoqi and Yu, Jia and Liu, Feifan and Du, Yihan and others},
  title        = {Safe Task Planning with Long-Term Graph Memory for Embodied Agents},
  journal={arXiv preprint arXiv:2609.08444},
  year         = {2026}
}

@misc{sima2026jitmemory,
  author       = {Sima, Bingrui and Wang, Lizhong and Lu, Xiaoya and He, Kun and Yang, Xiao},
  title        = {Self-Evolving Just-in-Time Memory for Proactive Embodied Safety},
  howpublished = {Lecture Notes in Computer Science},
  year         = {2026},
}

@inproceedings{lu2026isbench,
  author    = {Lu, Xiaoya and Chen, Zeren and Hu, Xuhao and Zhou, Yijin and Zhang, Weichen and Liu, Dongrui and others},
  title     = {IS-Bench: Evaluating Interactive Safety of VLM-Driven Embodied Agents in Daily Household Tasks},
  booktitle = {AAAI},
  year      = {2026},
}

@misc{ying2025agentsafe,
  author       = {Ying, Zonghao and Wang, Le and Xiao, Yisong and Wang, Jiakai and Ma, Yingjun and Guo, Jinyang and others},
  title        = {AGENTSAFE: Benchmarking the Safety of Embodied Agents on Hazardous Instructions},
  journal={arXiv preprint arXiv:2506.14697},
  year         = {2025}
}

@misc{zhang2025safevla,
  author       = {Zhang, Borong and Zhang, Yuhao and Ji, Jiaming and Lei, Yingshan and Cai, Yishuai and Dai, Juntao and others},
  title        = {SafeVLA: Towards Safety Alignment of Vision-Language-Action Model via Constrained Learning},
  journal={arXiv preprint arXiv:2503.03480},
  year         = {2025}
}

@misc{hu2025vlsa,
  author       = {Hu, Songqiao and Liu, Zeyi and Liu, Shuang and Cen, Jun and Meng, Zihan and Wang, Shihefeng and others},
  title        = {VLSA: Vision-Language-Action Models with Plug-and-Play Safety Constraint Layer},
  journal={arXiv preprint arXiv:2512.11891},
  year         = {2025}
}

@misc{tabbara2025latent,
  author       = {Tabbara, Ihab and Yang, Yuxuan and Hamzeh, Ahmad and Astafyev, Maxwell and Sibai, Hussein},
  title        = {Designing Latent Safety Filters using Pre-Trained Vision Models},
  journal={arXiv preprint arXiv:2509.14758},
  year         = {2025}
}

@misc{seo2025uncertaintyfilter,
  author       = {Seo, Junwon and Nakamura, Kensuke and Bajcsy, Andrea V.},
  title        = {Uncertainty-Aware Latent Safety Filters for Avoiding Out-of-Distribution Failures},
  journal={arXiv preprint arXiv:2505.00779},
  year         = {2025}
}

@article{haddadin2009safe,
  author  = {Haddadin, Sami and Albu-Schäffer, Alin and Hirzinger, Gerd},
  title   = {Requirements for Safe Robots: Measurements, Analysis and New Insights},
  journal = {The International Journal of Robotics Research},
  year    = {2009},
}

@article{villani2018hri,
  author  = {Villani, Valeria and Pini, Fabio and Leali, Francesco and Secchi, Cristian},
  title   = {Survey on Human--Robot Collaboration in Industrial Settings: Safety, Intuitive Interfaces and Applications},
  journal = {Mechatronics},
  year    = {2018},
}

@misc{ray2019safetygym,
  author       = {Ray, Alex and Achiam, Joshua and Amodei, Dario},
  title        = {Benchmarking Safe Exploration in Deep Reinforcement Learning},
  journal={arXiv preprint arXiv:1910.01708},
  year         = {2019}
}

@inproceedings{o2024open,
  title={Open x-embodiment: Robotic learning datasets and rt-x models: Open x-embodiment collaboration 0},
  author={O’Neill, Abby and Rehman, Abdul and Maddukuri, Abhiram and Gupta, Abhishek and Padalkar, Abhishek and Lee, Abraham and others},
  booktitle={ICRA},
  year={2024},
}

@article{kolve2017ai2,
  title={Ai2-thor: An interactive 3d environment for visual ai},
  author={Eric Kolve and Roozbeh Mottaghi and Winson Han and Eli VanderBilt and Luca Weihs and Alvaro Herrasti and others},
  journal={arXiv preprint arXiv:1712.05474},
  year={2017}
}

@inproceedings{savva2019habitat,
  title={Habitat: A platform for embodied ai research},
  author={Manolis Savva and Abhishek Kadian and Oleksandr Maksymets and Yili Zhao and Erik Wijmans and Bhavana Jain and others},
  booktitle={ICCV},
  year={2019},
}

@inproceedings{shridhar2020alfred,
  title={Alfred: A benchmark for interpreting grounded instructions for everyday tasks},
  author={Shridhar, Mohit and Thomason, Jesse and Gordon, Daniel and Bisk, Yonatan and Han, Winson and Mottaghi, Roozbeh and others},
  booktitle={CVPR},
  year={2020},
}

@article{james2020rlbench,
  title={Rlbench: The robot learning benchmark \& learning environment},
  author={James, Stephen and Ma, Zicong and Arrojo, David Rovick and Davison, Andrew J},
  journal={IEEE Robotics and Automation Letters},
  year={2020}}

@article{szot2021habitat,
  title={Habitat 2.0: Training home assistants to rearrange their habitat},
  author={Szot, Andrew and Clegg, Alexander and Undersander, Eric and Wijmans, Erik and Zhao, Yili and Turner, John and others},
  journal={NeurIPS},
  year={2021},
}

@article{loop_shinn2023reflexion,
  author = {Noah Shinn and Federico Cassano and Edward Berman and Ashwin Gopinath and Karthik Narasimhan and Shunyu Yao},
  title = {{Reflexion: Language Agents with Verbal Reinforcement Learning}},
  journal = {arXiv preprint arXiv:2303.11366},
  year = {2023},
}

@article{loop_madaan2023selfrefine,
  author = {Aman Madaan and Niket Tandon and Prakhar Gupta and Skyler Hallinan and Luyu Gao and Sarah Wiegreffe and others},
  title = {{Self-Refine: Iterative Refinement with Self-Feedback}},
  journal = {arXiv preprint arXiv:2303.17651},
  year = {2023},
}

@inproceedings{loop_yao2023react,
  author = {Shunyu Yao and Jeffrey Zhao and Dian Yu and Nan Du and Izhak Shafran and Karthik Narasimhan and others},
  title = {{ReAct: Synergizing Reasoning and Acting in Language Models}},
  booktitle = {ICLR},
  year = {2023},
}

@article{loop_wang2023voyager,
  author = {Guanzhi Wang and Yuqi Xie and Yunfan Jiang and Ajay Mandlekar and Chaowei Xiao and Yuke Zhu and others},
  title = {{Voyager: An Open-Ended Embodied Agent with Large Language Models}},
  journal = {arXiv preprint arXiv:2305.16291},
  year = {2023},
}

@inproceedings{loop_jimenez2024swebench,
  author = {Carlos E. Jimenez and John Yang and Alexander Wettig and Shunyu Yao and Kexin Pei and Ofir Press and others},
  title = {{SWE-bench: Can Language Models Resolve Real-World GitHub Issues?}},
  booktitle = {ICLR},
  year = {2024},
}

@article{loop_yang2024sweagent,
  author = {John Yang and Carlos E. Jimenez and Alexander Wettig and Kilian Lieret and Shunyu Yao and Karthik Narasimhan and others},
  title = {{SWE-agent: Agent-Computer Interfaces Enable Automated Software Engineering}},
  journal = {arXiv preprint arXiv:2405.15793},
  year = {2024},
}

@inproceedings{loop_wang2025openhands,
  author = {Xingyao Wang and Boxuan Li and Yufan Song and Frank F. Xu and Xiangru Tang and Mingchen Zhuge and others},
  title = {{OpenHands: An Open Platform for AI Software Developers as Generalist Agents}},
  booktitle = {ICLR},
  year = {2025},
}

@inproceedings{loop_gou2024critic,
  author = {Zhibin Gou and Zhihong Shao and Yeyun Gong and Yelong Shen and Yujiu Yang and Nan Duan and others},
  title = {{CRITIC: Large Language Models Can Self-Correct with Tool-Interactive Critiquing}},
  booktitle = {ICLR},
  year = {2024},
}

@inproceedings{loop_wang2023selfconsistency,
  author = {Xuezhi Wang and Jason Wei and Dale Schuurmans and Quoc Le and Ed Chi and Sharan Narang and others},
  title = {{Self-Consistency Improves Chain of Thought Reasoning in Language Models}},
  booktitle = {ICLR},
  year = {2023},
}

@article{loop_huang2022innermonologue,
  author = {Wenlong Huang and Fei Xia and Ted Xiao and Harris Chan and Jacky Liang and Pete Florence and others},
  title = {{Inner Monologue: Embodied Reasoning through Planning with Language Models}},
  journal = {arXiv preprint arXiv:2207.05608},
  year = {2022},
}

@article{loop_liang2022codepolicies,
  author = {Jacky Liang and Wenlong Huang and Fei Xia and Peng Xu and Karol Hausman and Brian Ichter and others},
  title = {{Code as Policies: Language Model Programs for Embodied Control}},
  journal = {arXiv preprint arXiv:2209.07753},
  year = {2022},
}

@article{loop_brohan2023rt2,
  author = {Anthony Brohan and Noah Brown and Justice Carbajal and Yevgen Chebotar and Xi Chen and Krzysztof Choromanski and others},
  title = {{RT-2: Vision-Language-Action Models Transfer Web Knowledge to Robotic Control}},
  journal = {arXiv preprint arXiv:2307.15818},
  year = {2023},
}

@article{loop_oxe2023,
  author = {{Open X-Embodiment Collaboration} and Abby O'Neill and Abdul Rehman and Abhinav Gupta and Abhiram Maddukuri and Abhishek Gupta and others},
  title = {{Open X-Embodiment: Robotic Learning Datasets and RT-X Models}},
  journal = {arXiv preprint arXiv:2310.08864},
  year = {2023},
}

@article{loop_octo2024,
  author = {{Octo Model Team} and Dibya Ghosh and Homer Walke and Karl Pertsch and Kevin Black and Oier Mees and others},
  title = {{Octo: An Open-Source Generalist Robot Policy}},
  journal = {arXiv preprint arXiv:2405.12213},
  year = {2024},
}

@article{loop_khazatsky2024droid,
  author = {Alexander Khazatsky and Karl Pertsch and Suraj Nair and Ashwin Balakrishna and Sudeep Dasari and Siddharth Karamcheti and others},
  title = {{DROID: A Large-Scale In-The-Wild Robot Manipulation Dataset}},
  journal = {arXiv preprint arXiv:2403.12945},
  year = {2024},
}

@article{loop_bousmalis2023robocat,
  author = {Konstantinos Bousmalis and Giulia Vezzani and Dushyant Rao and Coline Devin and Alex X. Lee and Maria Bauza and others},
  title = {{RoboCat: A Self-Improving Generalist Agent for Robotic Manipulation}},
  journal = {Transactions on Machine Learning Research},
  year = {2023},
}

@inproceedings{loop_ma2024eureka,
  author = {Yecheng Jason Ma and William Liang and Guanzhi Wang and De-An Huang and Osbert Bastani and Dinesh Jayaraman and others},
  title = {{Eureka: Human-Level Reward Design via Coding Large Language Models}},
  booktitle = {ICLR},
  year = {2024},
}

@article{loop_driess2023palme,
  author = {Danny Driess and Fei Xia and Mehdi S. M. Sajjadi and Corey Lynch and Aakanksha Chowdhery and Brian Ichter and others},
  title = {{PaLM-E: An Embodied Multimodal Language Model}},
  journal = {arXiv preprint arXiv:2303.03378},
  year = {2023},
}

@article{loop_brohan2022rt1,
  author = {Brohan, Anthony and Brown, Noah and Carbajal, Justice and Chebotar, Yevgen and Dabis, Joseph and Finn, Chelsea and others},
  title = {{RT-1: Robotics Transformer for Real-World Control at Scale}},
  journal = {arXiv preprint arXiv:2212.06817},
  year = {2022},
}

@inproceedings{loop_wang2024gensim,
  author = {Lirui Wang and Yiyang Ling and Zhecheng Yuan and Mohit Shridhar and Chen Bao and Yuzhe Qin and others},
  title = {{GenSim: Generating Robotic Simulation Tasks via Large Language Models}},
  booktitle = {ICLR},
  year = {2024},
}

@inproceedings{loop_wang2024robogen,
  author = {Yufei Wang and Zhou Xian and Feng Chen and Tsun-Hsuan Wang and Yian Wang and Katerina Fragkiadaki and others},
  title = {{RoboGen: Towards Unleashing Infinite Data for Automated Robot Learning via Generative Simulation}},
  booktitle = {ICML},
  year = {2024},
}

@inproceedings{loop_mandlekar2021demonstrations,
  author = {Ajay Mandlekar and Danfei Xu and Josiah Wong and Soroush Nasiriany and Chen Wang and Rohun Kulkarni and others},
  title = {{What Matters in Learning from Offline Human Demonstrations for Robot Manipulation}},
  booktitle = {CoRL},
  year = {2021},
}

@article{loop_fu2024mobilealoha,
  author = {Zipeng Fu and Tony Z. Zhao and Chelsea Finn},
  title = {{Mobile ALOHA: Learning Bimanual Mobile Manipulation with Low-Cost Whole-Body Teleoperation}},
  journal = {arXiv preprint arXiv:2401.02117},
  year = {2024},
}

@article{loop_ahn2024autort,
  author = {Michael Ahn and Debidatta Dwibedi and Chelsea Finn and Montse Gonzalez Arenas and Keerthana Gopalakrishnan and Karol Hausman and others},
  title = {{AutoRT: Embodied Foundation Models for Large Scale Orchestration of Robotic Agents}},
  journal = {arXiv preprint arXiv:2401.12963},
  year = {2024},
}

@article{loop_zhang2025dgm,
  author = {Jenny Zhang and Shengran Hu and Cong Lu and Robert Lange and Jeff Clune},
  title = {{Darwin Godel Machine: Open-Ended Evolution of Self-Improving Agents}},
  journal = {arXiv preprint arXiv:2505.22954},
  year = {2025},
}

@inproceedings{loop_huang2024selfcorrect,
  author = {Jie Huang and Xinyun Chen and Swaroop Mishra and Huaixiu Steven Zheng and Adams Wei Yu and Xinying Song and others},
  title = {{Large Language Models Cannot Self-Correct Reasoning Yet}},
  booktitle = {ICLR},
  year = {2024},
}

@inproceedings{loop_liu2023reflect,
  author = {Zeyi Liu and Arpit Bahety and Shuran Song},
  title = {{REFLECT: Summarizing Robot Experiences for Failure Explanation and Correction}},
  booktitle = {CoRL},
  year = {2023},
}

@article{loop_kober2013rl,
  author = {Kober, Jens and Bagnell, J. Andrew and Peters, Jan},
  title = {{Reinforcement learning in robotics: A survey}},
  journal = {The International Journal of Robotics Research},
  year = {2013},
}

@inproceedings{li2023behavior1k,
  title     = {BEHAVIOR-1K: A Benchmark for Embodied AI with 1,000 Everyday Activities and Realistic Simulation},
  author    = {Li, Chengshu and Zhang, Ruohan and Wong, Josiah and Gokmen, Cem and Srivastava, Sanjana and Mart{\'i}n-Mart{\'i}n, Roberto and others},
  booktitle = {CoRL},
  year      = {2023}}

@inproceedings{tao2025maniskill3,
  title     = {Demonstrating GPU Parallelized Robot Simulation and Rendering for Generalizable Embodied AI with ManiSkill3},
  author    = {Tao, Stone and Xiang, Fanbo and Shukla, Arth and Qin, Yuzhe and Hinrichsen, Xander and Yuan, Xiaodi and others},
  booktitle = {RSS},
  year      = {2025},
}

@inproceedings{zakka2025mujocoplayground,
  title     = {Demonstrating MuJoCo Playground},
  author    = {Zakka, Kevin and Tabanpour, Baruch and Liao, Qiayuan and Haiderbhai, Mustafa and Holt, Samuel and Luo, Jing Yuan and others},
  booktitle = {RSS},
  year      = {2025},
}

@inproceedings{geng2025roboverse,
  title     = {RoboVerse: A Unified Platform, Benchmark and Dataset for Scalable and Generalizable Robot Learning},
  author    = {Geng, Haoran and Wang, Feishi and Wei, Songlin and Li, Yuyang and Wang, Bangjun and An, Boshi and others},
  booktitle = {RSS},
  year      = {2025},
}

@inproceedings{li2025taccel,
  title     = {Taccel: Scaling Up Vision-based Tactile Robotics via High-performance GPU Simulation},
  author    = {Li, Yuyang and Du, Wenxin and Yu, Chang and Li, Puhao and Zhao, Zihang and Liu, Tengyu and others},
  booktitle = {NeurIPS},
  year      = {2025},
}

@article{mittal2025isaaclab,
  title         = {Isaac Lab: A GPU-Accelerated Simulation Framework for Multi-Modal Robot Learning},
  author        = {Mittal, Mayank and Roth, Pascal and Tigue, James and Richard, Antoine and Zhang, Octi and Du, Peter and others},
  journal       = {arXiv preprint arXiv:2511.04831},
  year          = {2025},
}

@inproceedings{zhou2025persistentwm,
  title     = {Learning 3D Persistent Embodied World Models},
  author    = {Zhou, Siyuan and Du, Yilun and Yang, Yuncong and Han, Lei and Chen, Peihao and Yeung, Dit-Yan and others},
  booktitle = {NeurIPS},
  year      = {2025},
}

@inproceedings{cao2026dlolab,
  title     = {DLO-Lab: Benchmarking Deformable Linear Object Manipulations with Differentiable Physics},
  author    = {Cao, Junyi and Wang, Yian and Xiong, Ziyan and Lin, Chunru and Chen, Zhehuan and Gan, Chuang},
  booktitle = {ICML},
  year      = {2026}}

@inproceedings{huang2026soma,
  title     = {SoMA: A Real-to-Sim Neural Simulator for Robotic Soft-Body Manipulation},
  author    = {Huang, Mu and Wang, Hui and Ren, Kerui and Xu, Linning and Zhou, Yunsong and Yu, Mulin and others},
  booktitle = {ICML},
  year      = {2026}}

@inproceedings{zhu2026flatlab,
  title     = {FlatLab: A Unified Methodology Framework and Simulation-Based Benchmark for Robotic Manipulation of Flat Objects},
  author    = {Zhu, Xingyu and Han, Wenshuo and Wang, Zhouyu and Wang, Yuran and Wu, Ruihai and Dong, Hao and others},
  booktitle = {ICML},
  year      = {2026}}

@inproceedings{nasiriany2026robocasa365,
  title     = {RoboCasa365: A Large-Scale Simulation Framework for Training and Benchmarking Generalist Robots},
  author    = {Nasiriany, Soroush and Nasiriany, Sep and Maddukuri, Abhiram and Zhu, Yuke},
  booktitle = {ICLR},
  year      = {2026}
}

@inproceedings{kim2026molmospaces,
  title     = {MolmoSpaces: Large-Scale Open Ecosystem for Robot Manipulation and Navigation},
  author    = {Kim, Yejin and Pumacay, Wilbert and Rayyan, Omar and Argus, Max and Han, Winson and Vanderbilt, Eli and others},
  booktitle = {RSS},
  year      = {2026},
}

@inproceedings{wang2026interactiveworld,
  title     = {Interactive World Simulator for Robot Policy Training and Evaluation},
  author    = {Wang, Yixuan and Syed, Rhythm and Wu, Fangyu and Zhang, Mengchao and Onol, Aykut and Barreiros, Jose and others},
  booktitle = {RSS},
  year      = {2026},
}

@inproceedings{yang2025embodiedbench,
  title     = {EmbodiedBench: Comprehensive Benchmarking Multi-modal Large Language Models for Vision-Driven Embodied Agents},
  author    = {Yang, Rui and Chen, Hanyang and Zhang, Junyu and Zhao, Mark and Qian, Cheng and Wang, Kangrui and others},
  booktitle = {ICML},
  year      = {2025},
}

@inproceedings{zhang2025vlabench,
  title     = {VLABench: A Large-Scale Benchmark for Language-Conditioned Robotics Manipulation with Long-Horizon Reasoning Tasks},
  author    = {Zhang, Shiduo and Xu, Zhe and Liu, Peiju and Yu, Xiaopeng and Li, Yuan and Gao, Qinghui and others},
  booktitle = {ICCV},
  year      = {2025}
}

@inproceedings{han2025robocerebra,
  title     = {RoboCerebra: A Large-scale Benchmark for Long-horizon Robotic Manipulation Evaluation},
  author    = {Han, Songhao and Qiu, Boxiang and Liao, Yue and Huang, Siyuan and Gao, Chen and Yan, Shuicheng and others},
  booktitle = {NeurIPS},
  year      = {2025},
}

@inproceedings{gundawar2025pacbench,
  title     = {PAC Bench: Do Foundation Models Understand Prerequisites for Executing Manipulation Policies?},
  author    = {Gundawar, Atharva and Sagar, Som and Senanayake, Ransalu},
  booktitle = {NeurIPS},
  year      = {2025},
}

@inproceedings{lin2025robotsmith,
  title     = {RobotSmith: Generative Robotic Tool Design for Acquisition of Complex Manipulation Skills},
  author    = {Lin, Chunru and Yuan, Haotian and Wang, Yian and Qiu, Xiaowen and Wang, Tsun-Hsuan Johnson and Guo, Minghao and others},
  booktitle = {NeurIPS},
  year      = {2025},
}

@inproceedings{wang2026enact,
  title     = {ENACT: Evaluating Embodied Cognition with World Modeling of Egocentric Interaction},
  author    = {Wang, Qineng and Huang, Wenlong and Zhou, Yu and Yin, Hang and Bao, Tianwei and Lyu, Jianwen and others},
  booktitle = {ICLR},
  year      = {2026}
}

@inproceedings{fu2026capx,
  title     = {CaP-X: A Framework for Benchmarking and Improving Coding Agents for Robot Manipulation},
  author    = {Fu, Letian and Yu, Justin and El-Refai, Karim and Kou, Ethan and Xue, Haoru and Huang, Huang and others},
  booktitle = {ICML},
  year      = {2026}}

@inproceedings{dai2026robomme,
  title     = {RoboMME: Benchmarking and Understanding Memory for Robotic Generalist Policies},
  author    = {Dai, Yinpei and Fu, Hongze and Lee, Jayjun and Liu, Yuejiang and Zhang, Haoran and Yang, Jianing and others},
  booktitle = {ICML},
  year      = {2026}}

@inproceedings{yan2026vlafix,
  title     = {Can VLMs Diagnose and Recover from VLA Manipulation Faults?},
  author    = {Yan, Bowen and Xiao, Jiahao and Liu, Kehui and Zhang, Jianbo and Zhang, Zicheng and Jia, Qi and others},
  booktitle = {ICML},
  year      = {2026}}

@inproceedings{kang2026movingout,
  title     = {Moving Out: Physically-grounded Human-AI Collaboration},
  author    = {Kang, Xuhui and Lee, Sung-Wook and Liu, Haolin and Wang, Yuyan and Kuo, Yen-Ling},
  booktitle = {ICML},
  year      = {2026}}

@inproceedings{he2026multiagent,
  title     = {A Closed-Loop Multi-Agent Framework for Robust Multi-Robot Manipulation},
  author    = {He, Yi-Xiang and Wei, Lan and Cen, Haoming and Jiang, Jian-Jian and Li, Zhuohao and Lu, Guanxing and others},
  booktitle = {RSS},
  year      = {2026},
}

@inproceedings{bai2026safelab,
  title     = {SafeLab: An Interactive High-Fidelity Benchmark for Embodied Safety in Scientific Robotics},
  author    = {Bai, Fengshuo and Li, Yufeng and Wu, Ruihai and Wang, Peishuo and Wang, Yuhan and Zhu, Bernie Hao and others},
  booktitle = {ICML},
  year      = {2026},
}

@inproceedings{balaji2026oopsieverse,
  title     = {OopsieVerse: A Safety Benchmark with Damage-Aware Simulation for Robot Manipulation},
  author    = {Balaji, Arnav and Bahety, Arpit and Ambatipudi, Sriniket and Lam, Daniel and Xu, Junhong and Mart{\'i}n-Mart{\'i}n, Roberto},
  booktitle = {RSS},
  year      = {2026},
}

@inproceedings{jangir2026robotarenainf,
  title     = {RobotArena $\infty$: Scalable Robot Benchmarking via Real-to-Sim Translation},
  author    = {Jangir, Yash and Zhang, Yidi and Yamazaki, Kashu and Zhang, Chenyu and Tu, Kuan-Hsun and Ke, Tsung-Wei and others},
  booktitle = {ICLR},
  year      = {2026}
}

@inproceedings{atreya2025roboarena,
  title     = {RoboArena: Distributed Real-World Evaluation of Generalist Robot Policies},
  author    = {Atreya, Pranav and Pertsch, Karl and Lee, Tony and Kim, Moo Jin and Jain, Arhan and Kuramshin, Artur and others},
  booktitle = {CoRL},
  year      = {2025},
}

@inproceedings{zhou2025autoeval,
  title     = {AutoEval: Autonomous Evaluation of Generalist Robot Manipulation Policies in the Real World},
  author    = {Zhou, Zhiyuan and Atreya, Pranav and Tan, You Liang and Pertsch, Karl and Levine, Sergey},
  booktitle = {CoRL},
  year      = {2025},
}

@inproceedings{li2025simpler,
  title     = {Evaluating Real-World Robot Manipulation Policies in Simulation},
  author    = {Li, Xuanlin and Hsu, Kyle and Gu, Jiayuan and Mees, Oier and Pertsch, Karl and Walke, Homer Rich and others},
  booktitle = {CoRL},
  year      = {2025},
}

@inproceedings{jain2026polaris,
  title     = {PolaRiS: Scalable Real-to-Sim Evaluations for Generalist Robot Policies},
  author    = {Jain, Arhan and Zhang, Mingtong and Arora, Kanav and Chen, William and Torne, Marcel and Irshad, Muhammad Zubair and others},
  booktitle = {RSS},
  year      = {2026},
}

@inproceedings{yang2026robolab,
  title     = {RoboLab: A High-Fidelity Simulation Benchmark for Analysis of Task Generalist Policies},
  author    = {Yang, Xuning and Dagli, Rishit and Zook, Alex and Hadfield, Hugo and Goyal, Ankit and Birchfield, Stan and others},
  booktitle = {RSS},
  year      = {2026},
}

@inproceedings{snyder2026beyondbinary,
  title     = {Beyond Binary Success: Sample-Efficient and Statistically Rigorous Robot Policy Comparison},
  author    = {Snyder, David and Badithela, Apurva and Matni, Nikolai and Pappas, George J. and Majumdar, Anirudha and Itkina, Masha and others},
  booktitle = {RSS},
  year      = {2026},
}

@article{ahn2022saycan,
  title={Do as i can, not as i say: Grounding language in robotic affordances},
  author={Michael Ahn and Anthony Brohan and Noah Brown and Yevgen Chebotar and Omar Cortes and Byron David and others},
  journal={arXiv preprint arXiv:2204.01691},
  year={2022}
}

@inproceedings{song2023llmplanner,
  title={Llm-planner: Few-shot grounded planning for embodied agents with large language models},
  author={Song, Chan Hee and Wu, Jiaman and Washington, Clayton and Sadler, Brian M and Chao, Wei-Lun and Su, Yu},
  booktitle={ICCV},
  year={2023}
}

@inproceedings{wu2024mldt,
  title={Mldt: Multi-level decomposition for complex long-horizon robotic task planning with open-source large language model},
  author={Wu, Yike and Zhang, Jiatao and Hu, Nan and Tang, Lanling and Qi, Guilin and Shao, Jun and others},
  booktitle={DASFAA},
  year={2024},
}

@inproceedings{prasad2024adapt,
  title={Adapt: As-needed decomposition and planning with language models},
  author={Prasad, Archiki and Koller, Alexander and Hartmann, Mareike and Clark, Peter and Sabharwal, Ashish and Bansal, Mohit and others},
  booktitle={Findings of the Association for Computational Linguistics: NAACL 2024},
  year={2024}
}

@inproceedings{liu2025delta,
  title={Delta: Decomposed efficient long-term robot task planning using large language models},
  author={Liu, Yuchen and Palmieri, Luigi and Koch, Sebastian and Georgievski, Ilche and Aiello, Marco},
  booktitle={ICRA},
  year={2025},
}

@article{ajay2023hip,
  title={Compositional foundation models for hierarchical planning},
  author={Ajay, Anurag and Han, Seungwook and Du, Yilun and Li, Shuang and Gupta, Abhi and Jaakkola, Tommi and others},
  journal={NeurIPS},
  year={2023}
}

@inproceedings{hu2024treeplanner,
  title={Tree-planner: Efficient close-loop task planning with large language models},
  author={Hu, Mengkang and Mu, Yao and Yu, Xinmiao and Ding, Mingyu and Wu, Shiguang and Shao, Wenqi and others},
  booktitle={ICLR},
  year={2024}
}

@inproceedings{du2023vlp,
  title={Video language planning},
  author={Du, Yilun and Yang, Mengjiao and Florence, Pete and Xia, Fei and Wahid, Ayzaan and Ichter, Brian and others},
  booktitle={ICLR},
  year={2024},
}

@article{lin2023text2motion,
  title={Text2motion: From natural language instructions to feasible plans},
  author={Lin, Kevin and Agia, Christopher and Migimatsu, Toki and Pavone, Marco and Bohg, Jeannette},
  journal={Autonomous Robots},
  year={2023}}

@article{li2024eai,
  title={Embodied agent interface: Benchmarking llms for embodied decision making},
  author={Manling Li and Shiyu Zhao and Qineng Wang and Kangrui Wang and Yu Zhou and Sanjana Srivastava and others},
  journal={NeurIPS},
  year={2024}
}

@article{singh2022progprompt,
  title={Progprompt: Generating situated robot task plans using large language models},
  author={Singh, Ishika and Blukis, Valts and Mousavian, Arsalan and Goyal, Ankit and Xu, Danfei and Tremblay, Jonathan and others},
  journal={arXiv preprint arXiv:2209.11302},
  year={2022}
}

@inproceedings{chen2024autotamp,
  title={Autotamp: Autoregressive task and motion planning with llms as translators and checkers},
  author={Chen, Yongchao and Arkin, Jacob and Dawson, Charles and Zhang, Yang and Roy, Nicholas and Fan, Chuchu},
  booktitle={ICRA},
  year={2024},
}

@article{huang2024rekep,
  title={Rekep: Spatio-temporal reasoning of relational keypoint constraints for robotic manipulation},
  author={Huang, Wenlong and Wang, Chen and Li, Yunzhu and Zhang, Ruohan and Fei-Fei, Li},
  journal={arXiv preprint arXiv:2409.01652},
  year={2024}
}

@article{huang2026roboharness,
  title={RoboHarness: Memory-Driven Orchestration of Heterogeneous Robot Policies for Long-Horizon Planning},
  author={Huang, Jinbang and Hu, Yuanzhao and Li, Zhiyuan and Qi, Ran and Xiao, Yixin and Zhang, Zhanguang and others},
  journal={arXiv preprint arXiv:2607.18060},
  year={2026}
}

@article{chen2026parallelworld,
  title={ParallelWorld: Test-Time Scaling for Embodied Reasoning},
  author={Chen, Min and Zhang, Shengjun and Li, Yuxin and Zhang, Zhang and Fei, Xin and Xia, Chong and others},
  journal={arXiv preprint arXiv:2608.22971},
  year={2026}
}

@article{liu2023llmp,
  title={Llm+ p: Empowering large language models with optimal planning proficiency},
  author={Liu, Bo and Jiang, Yuqian and Zhang, Xiaohan and Liu, Qiang and Zhang, Shiqi and Biswas, Joydeep and others},
  journal={arXiv preprint arXiv:2304.11477},
  year={2023}
}

@misc{ahmed2020causalworld,
  title = {{CausalWorld: A Robotic Manipulation Benchmark for Causal Structure and Transfer Learning}},
  author = {Ossama Ahmed and Frederik Tr{\"a}uble and Anirudh Goyal and Alexander Neitz and Yoshua Bengio and Bernhard Sch{\"o}lkopf and others},
  journal={arXiv preprint arXiv:2010.04296},
  year = {2020},
}

@misc{kelly2018hg,
  title = {{HG-DAgger: Interactive Imitation Learning with Human Experts}},
  author = {Michael Kelly and Chelsea Sidrane and Katherine Driggs-Campbell and Mykel J. Kochenderfer},
  journal={arXiv preprint arXiv:1810.02890},
  year = {2018},
}

@article{bohg2017interactive,
  title = {Interactive Perception: Leveraging Action in Perception and Perception in Action},
  author = {Bohg, Jeannette and Hausman, Karol and Sankaran, Bharath and Brock, Oliver and Kragic, Danica and Schaal, Stefan and others},
  journal = {IEEE Transactions on Robotics},
  year = {2017},
}

@inproceedings{houthooft2016vime,
  title = {{VIME}: Variational Information Maximizing Exploration},
  author = {Houthooft, Rein and Chen, Xi and Duan, Yan and Schulman, John and De Turck, Filip and Abbeel, Pieter},
  booktitle = {NeurIPS},
  year = {2016}
}

@inproceedings{sekar2020planning,
  title = {Planning to Explore via Self-Supervised World Models},
  author = {Sekar, Ramanan and Rybkin, Oleh and Daniilidis, Kostas and Abbeel, Pieter and Hafner, Danijar and Pathak, Deepak},
  booktitle = {ICML},
  year = {2020}
}

@inproceedings{agrawal2016learning,
  title = {Learning to Poke by Poking: Experiential Learning of Intuitive Physics},
  author = {Agrawal, Pulkit and Nair, Ashvin and Abbeel, Pieter and Malik, Jitendra and Levine, Sergey},
  booktitle = {NeurIPS},
  year = {2016}
}

@inproceedings{grauman2022ego4d,
  title = {{Ego4D}: Around the World in 3,000 Hours of Egocentric Video},
  author = {Kristen Grauman and Andrew Westbury and Eugene Byrne and Zachary Chavis and Antonino Furnari and Rohit Girdhar and others},
  booktitle = {CVPR},
  year = {2022}
}

@inproceedings{grauman2024egoexo4d,
  title = {{Ego-Exo4D}: Understanding Skilled Human Activity from First- and Third-Person Perspectives},
  author = {Kristen Grauman and Andrew Westbury and Lorenzo Torresani and Kris Kitani and Jitendra Malik and Triantafyllos Afouras and others},
  booktitle = {CVPR},
  year = {2024}
}

@inproceedings{bahl2023affordances,
  title = {Affordances from Human Videos as a Versatile Representation for Robotics},
  author = {Bahl, Shikhar and Mendonca, Russell and Chen, Lili and Jain, Unnat and Pathak, Deepak},
  booktitle = {CVPR},
  year = {2023}
}

@article{gu2023conceptgraphs,
  title = {{ConceptGraphs: Open-Vocabulary 3D Scene Graphs for Perception and Planning}},
  author = {Qiao Gu and Alihusein Kuwajerwala and Sacha Morin and Krishna Murthy Jatavallabhula and Bipasha Sen and Aditya Agarwal and others},
  journal = {arXiv preprint arXiv:2309.16650},
  year = {2023}
}

@inproceedings{florensa2018automatic,
  title = {Automatic Goal Generation for Reinforcement Learning Agents},
  author = {Florensa, Carlos and Held, David and Geng, Xinyang and Abbeel, Pieter},
  booktitle = {ICML},
  year = {2018}}

@article{portelas2019teacher,
  title = {Teacher Algorithms for Curriculum Learning of Deep {RL} in Continuously Parameterized Environments},
  author = {Portelas, R{\'e}my and Colas, C{\'e}dric and Hofmann, Katja and Oudeyer, Pierre-Yves},
  journal = {arXiv preprint arXiv:1910.07224},
  year = {2019}
}

@inproceedings{wang2024rlvlmf,
  title = {{RL-VLM-F}: Reinforcement Learning from Vision Language Foundation Model Feedback},
  author = {Wang, Yufei and Sun, Zhanyi and Zhang, Jesse and Xian, Zhou and Biyik, Erdem and Held, David and others},
  booktitle = {ICML},
  year = {2024}
}

@inproceedings{deer2024, author={Yue, Yang and Wang, Yulin and Kang, Bingyi and Han, Yizeng and Wang, Shenzhi and Song, Shiji and others}, title={{DeeR-VLA}: Dynamic inference of multimodal large language models for efficient robot execution}, booktitle={NeurIPS}, year={2024}}

@inproceedings{hirt2024, author={Zhang, Jianke and Guo, Yanjiang and Chen, Xiaoyu and Wang, Yen-Jen and Hu, Yucheng and Shi, Chengming and others}, title={{HiRT}: Enhancing robotic control with hierarchical robot transformers}, booktitle={CoRL}, year={2024}}

@inproceedings{fasa2026, author={Han, Yuchen and Wu, Jianhan and Qu, Xiaoyang and Kong, Lingwei and Li, Shiyi and Wang, Jianzong}, title={{FASA}: Feedback-aware sampling adaptation for efficient diffusion-based {VLA} models}, booktitle={NAS}, year={2026}}

@inproceedings{ecovla2026, author={Zhou, Ao and Dai, Bo and Yu, Le and Liu, Xingyu and Hao, Zeyu and Long, Lingkun and others}, title={{EcoVLA}: Energy-efficient device-edge co-inference for {VLA} models under real-time constraint}, booktitle={APPT}, year={2026}}

@inproceedings{comvla2026, author={Liu, Boliang and Poe, Wint Yi and Di, Jingyun and Trivisonno, Riccardo and Caire, Giuseppe}, title={{ComVLA}: Communication-aware split inference for {VLA} models in {6G}-connected robotics}, booktitle={IEEE GLOBECOM}, year={2026}}

@misc{vlaserving2026, author={Adamopoulos, Dionysios and Chanpaisit, Nattapol and Fakhri, Basel and Giannoula, Christina}, title={Efficient vision-language-action management and serving for robot factories}, year={2026}, journal={arXiv preprint arXiv:2609.12075}}

@misc{ulap2026, author={Cao, Deyu and Oi, Ryuji and Matsushima, Kosuke and Pan, Yuxuan and Wang, Ziheng and Fujiki, Daichi and others}, title={{VLA-ULAP}: Interleaving cloud {VLA} calls with ultra-lightweight local action prediction at the edge}, year={2026}, journal={arXiv preprint arXiv:2609.18663}}

@inproceedings{kwon2023vllm, author={Kwon, Woosuk and Li, Zhuohan and Zhuang, Siyuan and Sheng, Ying and Zheng, Lianmin and Yu, Cody Hao and others}, title={Efficient memory management for large language model serving with {PagedAttention}}, booktitle={ACM SOSP}, year={2023}}

@inproceedings{zheng2024sglang, author={Zheng, Lianmin and Yin, Liangsheng and Xie, Zhiqiang and Sun, Chuzhan and Huang, Jeff and Yu, Cody Hao and others}, title={{SGLang}: Efficient controlled generation with {RadixAttention}}, booktitle={NeurIPS}, year={2024}}

@misc{edgedlm2026, author={Li, Chenqi and Min, Minghui and Niyato, Dusit and Ni, Wei}, title={Diffusion language models for mobile edge agentic {AI}: Foundations, applications, and challenges},
  journal={arXiv preprint arXiv:2609.04778}, year={2026}}

@inproceedings{apisurvey2026, author={Li, Sheng and Li, Jing and Schijve, Felix and Hu, Jun and Barakova, Emilia}, title={Casting everything to online {API} services? {A} survey of integrating localized speech recognition models in robotics}, booktitle={ICSR}, year={2026}}

@inproceedings{kuffner2010cloud, author={Kuffner, James}, title={Cloud-enabled robots}, booktitle={IEEE-RAS Int'l Conf. on Humanoid Robots}, year={2010}}

@inproceedings{fogros2021, author={Chen, Kaiyuan and Liang, Yafei and Jha, Nikhil and Ichnowski, Jeffrey and Danielczuk, Michael and Gonzalez, Joseph and others}, title={{FogROS}: An adaptive framework for automating fog robotics deployment}, booktitle={CASE}, year={2021}}

@inproceedings{fog2019, author={Tanwani, Ajay Kumar and Mor, Nitesh and Kubiatowicz, John and Gonzalez, Joseph E. and Goldberg, Ken}, title={A Fog Robotics Approach to Deep Robot Learning: Application to Object Recognition and Grasp Planning in Surface Decluttering}, booktitle={ICRA}, year={2019}}

@misc{rosbagmcp2025, author={Fu, Lei and Salimpour, Sahar and Militano, Leonardo and Edelman, Harry and Peña Queralta, Jorge and Toffetti, Giovanni}, title={{ROSBag MCP} server: Analyzing robot data with {LLMs} for agentic embodied {AI} applications}, year={2025}, journal={arXiv preprint arXiv:2511.03497}}

@misc{drivemcp2026, author={Nadiri, Farzad and Cina, Mehdi and Rad, Ahmad B.}, title={{DriveMCP}: An agentic {AI} framework for advanced driver assistance system}, year={2026}, journal={arXiv preprint arXiv:2609.17247}}

@misc{roboos2025, author={Tan, Huajie and Hao, Xiaoshuai and Chi, Cheng and Lin, Minglan and Lyu, Yaoxu and Cao, Mingyu and others}, title={{RoboOS}: A hierarchical embodied framework for cross-embodiment and multi-agent collaboration}, year={2025}, journal={arXiv preprint arXiv:2505.03673}}

@misc{roboosnext2025, author={Tan, Huajie and Chi, Cheng and Chen, Xiansheng and Ji, Yuheng and Zhao, Zhongxia and Hao, Xiaoshuai and others}, title={{RoboOS-NeXT}: A unified memory-based framework for lifelong, scalable, and robust multi-robot collaboration},
  journal={arXiv preprint arXiv:2510.26536}, year={2025}}

@misc{nvidia2025groot, author={{NVIDIA}}, title={{GR00T N1}: An open foundation model for generalist humanoid robots}, year={2025}, journal={arXiv preprint arXiv:2503.14734}}

@misc{datafactory2026, author={Haddadin, Sami and Laptev, Ivan and Reid, Ian and Song, Dezhen and Stefanini, Cesare and Swikir, Abdalla and others}, title={The robot data factory}, year={2026}, journal={arXiv preprint arXiv:2609.16705}}

@article{bajcsy1988,
  title={Active perception},
  author={Bajcsy, Ruzena},
  journal={Proceedings of the IEEE},
  year={1988}}

@article{aloimonos1988, title={Active vision}, ISSN={1573-1405}, journal={International Journal of Computer Vision}, author={Aloimonos, John and Weiss, Isaac and Bandyopadhyay, Amit}, year={1988}, month=Jan,}

@inproceedings{clip,
  title={Learning transferable visual models from natural language supervision},
  author={Radford, Alec and Kim, Jong Wook and Hallacy, Chris and Ramesh, Aditya and Goh, Gabriel and Agarwal, Sandhini and others},
  booktitle={ICML},
  year={2021},
}

@article{vild,
  title={Open-vocabulary object detection via vision and language knowledge distillation},
  author={Gu, Xiuye and Lin, Tsung-Yi and Kuo, Weicheng and Cui, Yin},
  journal={arXiv preprint arXiv:2104.13921},
  year={2021}
}

@inproceedings{owlvit,
  title={Simple open-vocabulary object detection},
  author={Minderer, Matthias and Gritsenko, Alexey and Stone, Austin and Neumann, Maxim and Weissenborn, Dirk and Dosovitskiy, Alexey and others},
  booktitle={ECCV},
  year={2022},
}

@inproceedings{glip,
  title={Grounded language-image pre-training},
  author={Li, Liunian Harold and Zhang, Pengchuan and Zhang, Haotian and Yang, Jianwei and Li, Chunyuan and Zhong, Yiwu and others},
  booktitle={CVPR},
  year={2022},
}

@inproceedings{groundingdino,
  title={Grounding dino: Marrying dino with grounded pre-training for open-set object detection},
  author={Shilong Liu and Zhaoyang Zeng and Tianhe Ren and Feng Li and Hao Zhang and Jie Yang and others},
  booktitle={ECCV},
  year={2024},
}

@inproceedings{sam2,
  title={Sam 2: Segment anything in images and videos},
  author={Ravi, Nikhila and Gabeur, Valentin and Hu, Yuan-Ting and Hu, Ronghang and Ryali, Chaitanya and Ma, Tengyu and others},
  booktitle={ICLR},
  year={2025}
}

@misc{sam3,
      title={SAM 3: Segment Anything with Concepts}, 
      author={Nicolas Carion and Laura Gustafson and Yuan-Ting Hu and Shoubhik Debnath and Ronghang Hu and Didac Suris and others},
  journal={arXiv preprint arXiv:2511.16719},
      year={2026},
}

@article{aro,
  title={When and why vision-language models behave like bags-of-words, and what to do about it?},
  author={Yuksekgonul, Mert and Bianchi, Federico and Kalluri, Pratyusha and Jurafsky, Dan and Zou, James},
  journal={arXiv preprint arXiv:2210.01936},
  year={2022}
}

@inproceedings{mdetr,
  title={Mdetr-modulated detection for end-to-end multi-modal understanding},
  author={Kamath, Aishwarya and Singh, Mannat and LeCun, Yann and Synnaeve, Gabriel and Misra, Ishan and Carion, Nicolas},
  booktitle={ICCV},
  year={2021},
}

@article{kosmos2,
  title={Kosmos-2: Grounding multimodal large language models to the world},
  author={Peng, Zhiliang and Wang, Wenhui and Dong, Li and Hao, Yaru and Huang, Shaohan and Ma, Shuming and others},
  journal={arXiv preprint arXiv:2306.14824},
  year={2023}
}

@misc{som,
      title={Set-of-Mark Prompting Unleashes Extraordinary Visual Grounding in GPT-4V}, 
      author={Jianwei Yang and Hao Zhang and Feng Li and Xueyan Zou and Chunyuan Li and Jianfeng Gao},
  journal={arXiv preprint arXiv:2310.11441},
      year={2023},
}

@inproceedings{lisa,
  title={Lisa: Reasoning segmentation via large language model},
  author={Lai, Xin and Tian, Zhuotao and Chen, Yukang and Li, Yanwei and Yuan, Yuhui and Liu, Shu and others},
  booktitle={CVPR},
  year={2024},
}

@misc{myvlm,
      title={MyVLM: Personalizing VLMs for User-Specific Queries}, 
      author={Yuval Alaluf and Elad Richardson and Sergey Tulyakov and Kfir Aberman and Daniel Cohen-Or},
  journal={arXiv preprint arXiv:2403.14599},
      year={2024},
}

@misc{bringmycup,
      title={Bring My Cup! Personalizing Vision-Language-Action Models with Visual Attentive Prompting}, 
      author={Sangoh Lee and Sangwoo Mo and Wook-Shin Han},
  journal={arXiv preprint arXiv:2512.20014},
      year={2026},
}

@inproceedings{vlpart,
  title={Going denser with open-vocabulary part segmentation},
  author={Sun, Peize and Chen, Shoufa and Zhu, Chenchen and Xiao, Fanyi and Luo, Ping and Xie, Saining and others},
  booktitle={ICCV},
  year={2023},
}

@inproceedings{partslip,
  title={Partslip: Low-shot part segmentation for 3d point clouds via pretrained image-language models},
  author={Liu, Minghua and Zhu, Yinhao and Cai, Hong and Han, Shizhong and Ling, Zhan and Porikli, Fatih and others},
  booktitle={CVPR},
  year={2023},
}

@inproceedings{roboabc,
  title={Robo-abc: Affordance generalization beyond categories via semantic correspondence for robot manipulation},
  author={Ju, Yuanchen and Hu, Kaizhe and Zhang, Guowei and Zhang, Gu and Jiang, Mingrun and Xu, Huazhe},
  booktitle={ECCV},
  year={2024},
}

@misc{depthanything,
      title={Depth Anything: Unleashing the Power of Large-Scale Unlabeled Data}, 
      author={Lihe Yang and Bingyi Kang and Zilong Huang and Xiaogang Xu and Jiashi Feng and Hengshuang Zhao},
  journal={arXiv preprint arXiv:2401.10891},
      year={2024},
}

@misc{depthanythingv2,
      title={Depth Anything V2}, 
      author={Lihe Yang and Bingyi Kang and Zilong Huang and Zhen Zhao and Xiaogang Xu and Jiashi Feng and others},
  journal={arXiv preprint arXiv:2406.09414},
      year={2024},
}

@misc{metric3dv2,
      title={Metric3Dv2: A Versatile Monocular Geometric Foundation Model for Zero-shot Metric Depth and Surface Normal Estimation}, 
      author={Mu Hu and Wei Yin and Chi Zhang and Zhipeng Cai and Xiaoxiao Long and Kaixuan Wang and others},
  journal={arXiv preprint arXiv:2404.15506},
      year={2025},
}

@misc{unidepth,
      title={UniDepth: Universal Monocular Metric Depth Estimation}, 
      author={Luigi Piccinelli and Yung-Hsu Yang and Christos Sakaridis and Mattia Segu and Siyuan Li and Luc Van Gool and others},
  journal={arXiv preprint arXiv:2403.18913},
      year={2024},
}

@misc{foundationstereo,
      title={FoundationStereo: Zero-Shot Stereo Matching}, 
      author={Bowen Wen and Matthew Trepte and Joseph Aribido and Jan Kautz and Orazio Gallo and Stan Birchfield},
  journal={arXiv preprint arXiv:2501.09898},
      year={2025},
}

@misc{dust3r,
      title={DUSt3R: Geometric 3D Vision Made Easy}, 
      author={Shuzhe Wang and Vincent Leroy and Yohann Cabon and Boris Chidlovskii and Jerome Revaud},
  journal={arXiv preprint arXiv:2312.14132},
      year={2024},
}

@misc{mast3r,
      title={Grounding Image Matching in 3D with MASt3R}, 
      author={Vincent Leroy and Yohann Cabon and Jérôme Revaud},
  journal={arXiv preprint arXiv:2406.09756},
      year={2024},
}

@misc{vggt,
      title={VGGT: Visual Geometry Grounded Transformer}, 
      author={Jianyuan Wang and Minghao Chen and Nikita Karaev and Andrea Vedaldi and Christian Rupprecht and David Novotny},
  journal={arXiv preprint arXiv:2503.11651},
      year={2025},
}

@misc{mapanything,
      title={MapAnything: Universal Feed-Forward Metric 3D Reconstruction}, 
      author={Nikhil Keetha and Norman Müller and Johannes Schönberger and Lorenzo Porzi and Yuchen Zhang and Tobias Fischer and others},
  journal={arXiv preprint arXiv:2509.13414},
      year={2026},
}

@misc{robo3r,
      title={Robo3R: Enhancing Robotic Manipulation with Accurate Feed-Forward 3D Reconstruction}, 
      author={Sizhe Yang and Linning Xu and Hao Li and Juncheng Mu and Jia Zeng and Dahua Lin and others},
  journal={arXiv preprint arXiv:2602.10101},
      year={2026},
}

@misc{posecnn,
      title={PoseCNN: A Convolutional Neural Network for 6D Object Pose Estimation in Cluttered Scenes}, 
      author={Yu Xiang and Tanner Schmidt and Venkatraman Narayanan and Dieter Fox},
  journal={arXiv preprint arXiv:1711.00199},
      year={2018},
}

@misc{nocs,
      title={Normalized Object Coordinate Space for Category-Level 6D Object Pose and Size Estimation}, 
      author={He Wang and Srinath Sridhar and Jingwei Huang and Julien Valentin and Shuran Song and Leonidas J. Guibas},
  journal={arXiv preprint arXiv:1901.02970},
      year={2019},
}

@misc{megapose,
      title={MegaPose: 6D Pose Estimation of Novel Objects via Render \& Compare}, 
      author={Yann Labbé and Lucas Manuelli and Arsalan Mousavian and Stephen Tyree and Stan Birchfield and Jonathan Tremblay and others},
  journal={arXiv preprint arXiv:2212.06870},
      year={2022},
}

@misc{foundationpose,
      title={FoundationPose: Unified 6D Pose Estimation and Tracking of Novel Objects}, 
      author={Bowen Wen and Wei Yang and Jan Kautz and Stan Birchfield},
  journal={arXiv preprint arXiv:2312.08344},
      year={2024},
}

@misc{bundlesdf,
      title={BundleSDF: Neural 6-DoF Tracking and 3D Reconstruction of Unknown Objects}, 
      author={Bowen Wen and Jonathan Tremblay and Valts Blukis and Stephen Tyree and Thomas Muller and Alex Evans and others},
  journal={arXiv preprint arXiv:2303.14158},
      year={2023},
}

@misc{whatsup,
      title={What's "up" with vision-language models? Investigating their struggle with spatial reasoning}, 
      author={Amita Kamath and Jack Hessel and Kai-Wei Chang},
  journal={arXiv preprint arXiv:2310.19785},
      year={2023},
}

@misc{blink,
      title={BLINK: Multimodal Large Language Models Can See but Not Perceive}, 
      author={Xingyu Fu and Yushi Hu and Bangzheng Li and Yu Feng and Haoyu Wang and Xudong Lin and others},
  journal={arXiv preprint arXiv:2404.12390},
      year={2024},
}

@misc{mmsi,
      title={MMSI-Bench: A Benchmark for Multi-Image Spatial Intelligence}, 
      author={Sihan Yang and Runsen Xu and Yiman Xie and Sizhe Yang and Mo Li and Jingli Lin and others},
  journal={arXiv preprint arXiv:2505.23764},
      year={2026},
}

@misc{mvrobobench,
      title={Seeing Across Views: Benchmarking Spatial Reasoning of Vision-Language Models in Robotic Scenes}, 
      author={Zhiyuan Feng and Zhaolu Kang and Qijie Wang and Zhiying Du and Jiongrui Yan and Shubin Shi and others},
  journal={arXiv preprint arXiv:2510.19400},
      year={2026},
}

@misc{vsibench,
      title={Thinking in Space: How Multimodal Large Language Models See, Remember, and Recall Spaces}, 
      author={Jihan Yang and Shusheng Yang and Anjali W. Gupta and Rilyn Han and Li Fei-Fei and Saining Xie},
  journal={arXiv preprint arXiv:2412.14171},
      year={2025},
}

@inproceedings{spatialvlm,
  title={Spatialvlm: Endowing vision-language models with spatial reasoning capabilities},
  author={Chen, Boyuan and Xu, Zhuo and Kirmani, Sean and Ichter, Brain and Sadigh, Dorsa and Guibas, Leonidas and others},
  booktitle={CVPR},
  year={2024}
}

@misc{spatialrgpt,
      title={SpatialRGPT: Grounded Spatial Reasoning in Vision Language Models}, 
      author={An-Chieh Cheng and Hongxu Yin and Yang Fu and Qiushan Guo and Ruihan Yang and Jan Kautz and others},
  journal={arXiv preprint arXiv:2406.01584},
      year={2024},
}

@article{3dllm,
  title={3d-llm: Injecting the 3d world into large language models},
  author={Hong, Yining and Zhen, Haoyu and Chen, Peihao and Zheng, Shuhong and Du, Yilun and Chen, Zhenfang and others},
  journal={NeurIPS},
  year={2023}
}

@article{spatialmllm,
  title={Spatial-mllm: Boosting mllm capabilities in visual-based spatial intelligence},
  author={Wu, Diankun and Liu, Fangfu and Hung, Yi-Hsin and Duan, Yueqi},
  journal={NeurIPS},
  year={2026}
}

@misc{mindcube,
      title={MindCube: Spatial Mental Modeling from Limited Views}, 
      author={Qineng Wang and Baiqiao Yin and Pingyue Zhang and Jianshu Zhang and Kangrui Wang and Zihan Wang and others},
  journal={arXiv preprint arXiv:2506.21458},
      year={2026},
}

@misc{vadar,
      title={Visual Agentic AI for Spatial Reasoning with a Dynamic API}, 
      author={Damiano Marsili and Rohun Agrawal and Yisong Yue and Georgia Gkioxari},
  journal={arXiv preprint arXiv:2502.06787},
      year={2025},
}

@misc{spatialssrl,
      title={Spatial-SSRL: Enhancing Spatial Understanding via Self-Supervised Reinforcement Learning}, 
      author={Yuhong Liu and Beichen Zhang and Yuhang Zang and Yuhang Cao and Long Xing and Xiaoyi Dong and others},
  journal={arXiv preprint arXiv:2510.27606},
      year={2025},
}

@misc{spatialtree,
      title={SpatialTree: How Spatial Abilities Branch Out in MLLMs}, 
      author={Yuxi Xiao and Longfei Li and Shen Yan and Xinhang Liu and Sida Peng and Yunchao Wei and others},
  journal={arXiv preprint arXiv:2512.20617},
      year={2026},
}

@misc{comfort,
      title={Do Vision-Language Models Represent Space and How? Evaluating Spatial Frame of Reference Under Ambiguities}, 
      author={Zheyuan Zhang and Fengyuan Hu and Jayjun Lee and Freda Shi and Parisa Kordjamshidi and Joyce Chai and others},
  journal={arXiv preprint arXiv:2410.17385},
      year={2025},
}

@article{sqa3d,
  title={Sqa3d: Situated question answering in 3d scenes},
  author={Ma, Xiaojian and Yong, Silong and Zheng, Zilong and Li, Qing and Liang, Yitao and Zhu, Song-Chun and others},
  journal={arXiv preprint arXiv:2210.07474},
  year={2022}
}

@misc{sig3d,
      title={Situational Awareness Matters in 3D Vision Language Reasoning}, 
      author={Yunze Man and Liang-Yan Gui and Yu-Xiong Wang},
  journal={arXiv preprint arXiv:2406.07544},
      year={2024},
}

@inproceedings{robospatial,
  title={Robospatial: Teaching spatial understanding to 2d and 3d vision-language models for robotics},
  author={Song, Chan Hee and Blukis, Valts and Tremblay, Jonathan and Tyree, Stephen and Su, Yu and Birchfield, Stan},
  booktitle={CVPR},
  year={2025},
}

@misc{peract,
      title={Perceiver-Actor: A Multi-Task Transformer for Robotic Manipulation}, 
      author={Mohit Shridhar and Lucas Manuelli and Dieter Fox},
  journal={arXiv preprint arXiv:2209.05451},
      year={2022},
}

@inproceedings{vlmaps,
  title={Visual language maps for robot navigation},
  author={Huang, Chenguang and Mees, Oier and Zeng, Andy and Burgard, Wolfram},
  booktitle={ICRA},
  year={2023},
}

@article{conceptfusion,
  title={Conceptfusion: Open-set multimodal 3d mapping},
  author={Jatavallabhula, Krishna Murthy and Kuwajerwala, Alihusein and Gu, Qiao and Omama, Mohd and Chen, Tao and Maalouf, Alaa and others},
  journal={arXiv preprint arXiv:2302.07241},
  year={2023}
}

@inproceedings{lerf,
  title={Lerf: Language embedded radiance fields},
  author={Kerr, Justin and Kim, Chung Min and Goldberg, Ken and Kanazawa, Angjoo and Tancik, Matthew},
  booktitle={ICCV},
  year={2023},
}

@inproceedings{langsplat,
  title={Langsplat: 3d language gaussian splatting},
  author={Qin, Minghan and Li, Wanhua and Zhou, Jiawei and Wang, Haoqian and Pfister, Hanspeter},
  booktitle={CVPR},
  year={2024},
}

@article{hydra,
  title={Hydra: A real-time spatial perception system for 3D scene graph construction and optimization},
  author={Hughes, Nathan and Chang, Yun and Carlone, Luca},
  journal={arXiv preprint arXiv:2201.13360},
  year={2022}
}

@misc{physobjects,
      title={Physically Grounded Vision-Language Models for Robotic Manipulation}, 
      author={Jensen Gao and Bidipta Sarkar and Fei Xia and Ted Xiao and Jiajun Wu and Brian Ichter and others},
  journal={arXiv preprint arXiv:2309.02561},
      year={2024},
}

@misc{phys2real,
      title={Phys2Real: Fusing VLM Priors with Interactive Online Adaptation for Uncertainty-Aware Sim-to-Real Manipulation}, 
      author={Maggie Wang and Stephen Tian and Aiden Swann and Ola Shorinwa and Jiajun Wu and Mac Schwager},
  journal={arXiv preprint arXiv:2510.11689},
      year={2026},
}

@misc{physbench,
      title={PhysBench: Benchmarking and Enhancing Vision-Language Models for Physical World Understanding}, 
      author={Wei Chow and Jiageng Mao and Boyi Li and Daniel Seita and Vitor Guizilini and Yue Wang},
  journal={arXiv preprint arXiv:2501.16411},
      year={2025},
}

@article{gelsight, title={GelSight: High-Resolution Robot Tactile Sensors for Estimating Geometry and Force}, ISSN={1424-8220}, journal={Sensors}, author={Yuan, Wenzhen and Dong, Siyuan and Adelson, Edward}, year={2017}, month=Nov,}

@misc{octopi,
      title={Octopi: Object Property Reasoning with Large Tactile-Language Models}, 
      author={Samson Yu and Kelvin Lin and Anxing Xiao and Jiafei Duan and Harold Soh},
  journal={arXiv preprint arXiv:2405.02794},
      year={2024},
}

@misc{vlatouch,
      title={VLA-Touch: Enhancing Vision-Language-Action Models with Dual-Level Tactile Feedback}, 
      author={Jianxin Bi and Kevin Yuchen Ma and Ce Hao and Mike Zheng Shou and Harold Soh},
  journal={arXiv preprint arXiv:2507.17294},
      year={2025},
}

@misc{forcevla,
      title={ForceVLA: Enhancing VLA Models with a Force-aware MoE for Contact-rich Manipulation}, 
      author={Jiawen Yu and Hairuo Liu and Qiaojun Yu and Jieji Ren and Ce Hao and Haitong Ding and others},
  journal={arXiv preprint arXiv:2505.22159},
      year={2025},
}

@misc{tavla,
      title={TA-VLA: Elucidating the Design Space of Torque-aware Vision-Language-Action Models}, 
      author={Zongzheng Zhang and Haobo Xu and Zhuo Yang and Chenghao Yue and Zehao Lin and Huan-ang Gao and others},
  journal={arXiv preprint arXiv:2509.07962},
      year={2025},
}

@misc{where2act,
      title={Where2Act: From Pixels to Actions for Articulated 3D Objects}, 
      author={Kaichun Mo and Leonidas Guibas and Mustafa Mukadam and Abhinav Gupta and Shubham Tulsiani},
  journal={arXiv preprint arXiv:2101.02692},
      year={2021},
}

@misc{ditto,
      title={Ditto: Building Digital Twins of Articulated Objects from Interaction}, 
      author={Zhenyu Jiang and Cheng-Chun Hsu and Yuke Zhu},
  journal={arXiv preprint arXiv:2202.08227},
      year={2022},
}

@misc{interactionnet,
      title={Interaction Networks for Learning about Objects, Relations and Physics}, 
      author={Peter W. Battaglia and Razvan Pascanu and Matthew Lai and Danilo Rezende and Koray Kavukcuoglu},
  journal={arXiv preprint arXiv:1612.00222},
      year={2016},
}

@misc{dpinet,
      title={Learning Particle Dynamics for Manipulating Rigid Bodies, Deformable Objects, and Fluids}, 
      author={Yunzhu Li and Jiajun Wu and Russ Tedrake and Joshua B. Tenenbaum and Antonio Torralba},
  journal={arXiv preprint arXiv:1810.01566},
      year={2019},
}

@misc{adaptigraph,
      title={AdaptiGraph: Material-Adaptive Graph-Based Neural Dynamics for Robotic Manipulation}, 
      author={Kaifeng Zhang and Baoyu Li and Kris Hauser and Yunzhu Li},
  journal={arXiv preprint arXiv:2407.07889},
      year={2024},
}

@misc{physion,
      title={Physion: Evaluating Physical Prediction from Vision in Humans and Machines}, 
      author={Daniel M. Bear and Elias Wang and Damian Mrowca and Felix J. Binder and Hsiao-Yu Fish Tung and R. T. Pramod and others},
  journal={arXiv preprint arXiv:2106.08261},
      year={2022},
}

@misc{vjepa2,
      title={V-JEPA 2: Self-Supervised Video Models Enable Understanding, Prediction and Planning}, 
      author={Mido Assran and Adrien Bardes and David Fan and Quentin Garrido and Russell Howes and Mojtaba and others},
  journal={arXiv preprint arXiv:2506.09985},
      year={2025},
}

@misc{cotracker,
      title={CoTracker: It is Better to Track Together}, 
      author={Nikita Karaev and Ignacio Rocco and Benjamin Graham and Natalia Neverova and Andrea Vedaldi and Christian Rupprecht},
  journal={arXiv preprint arXiv:2307.07635},
      year={2024},
}

@misc{neuralfeels,
      title={Neural feels with neural fields: Visuo-tactile perception for in-hand manipulation}, 
      author={Sudharshan Suresh and Haozhi Qi and Tingfan Wu and Taosha Fan and Luis Pineda and Mike Lambeta and others},
  journal={arXiv preprint arXiv:2312.13469},
      year={2023},
}

@misc{vhop,
      title={V-HOP: Visuo-Haptic 6D Object Pose Tracking}, 
      author={Hongyu Li and Mingxi Jia and Tuluhan Akbulut and Yu Xiang and George Konidaris and Srinath Sridhar},
  journal={arXiv preprint arXiv:2502.17434},
      year={2025},
}

@misc{orbslam3,
      title={ORB-SLAM3: An Accurate Open-Source Library for Visual, Visual-Inertial and Multi-Map SLAM}, 
      author={Carlos Campos and Richard Elvira and Juan J. Gómez Rodríguez and José M. M. Montiel and Juan D. Tardós},
  journal={arXiv preprint arXiv:2007.11898},
      year={2021},
}

@inproceedings{onemap,
  title={One map to find them all: Real-time open-vocabulary mapping for zero-shot multi-object navigation},
  author={Busch, Finn Lukas and Homberger, Timon and Ortega-Peimbert, Jes{\'u}s and Yang, Quantao and Andersson, Olov},
  booktitle={ICRA},
  year={2025},
}

@misc{panopticmtsdf,
      title={Panoptic Multi-TSDFs: a Flexible Representation for Online Multi-resolution Volumetric Mapping and Long-term Dynamic Scene Consistency}, 
      author={Lukas Schmid and Jeffrey Delmerico and Johannes Schönberger and Juan Nieto and Marc Pollefeys and Roland Siegwart and others},
  journal={arXiv preprint arXiv:2109.10165},
      year={2022},
}

@article{khronos,
  title={Khronos: A unified approach for spatio-temporal metric-semantic slam in dynamic environments},
  author={Schmid, Lukas and Abate, Marcus and Chang, Yun and Carlone, Luca},
  journal={arXiv preprint arXiv:2402.13817},
  year={2024}
}

@article{dovsg,
  title={Dynamic open-vocabulary 3d scene graphs for long-term language-guided mobile manipulation},
  author={Yan, Zhijie and Li, Shufei and Wang, Zuoxu and Wu, Lixiu and Wang, Han and Zhu, Jun and others},
  journal={IEEE Robotics and Automation Letters},
  year={2025}}

@misc{dynamem,
      title={DynaMem: Online Dynamic Spatio-Semantic Memory for Open World Mobile Manipulation}, 
      author={Peiqi Liu and Zhanqiu Guo and Mohit Warke and Soumith Chintala and Chris Paxton and Nur Muhammad Mahi Shafiullah and others},
  journal={arXiv preprint arXiv:2411.04999},
      year={2025},
}

@misc{ltmem,
      title={LT-Mem: Volatility-Aware Spatio-Temporal Memory for Lifelong Scene Understanding}, 
      author={Yumin Lee and Hyoseok Ju and Giseop Kim},
  journal={arXiv preprint arXiv:2608.19059},
      year={2026},
}

@inproceedings{remembr,
  title={Remembr: Building and reasoning over long-horizon spatio-temporal memory for robot navigation},
  author={Anwar, Abrar and Welsh, John and Biswas, Joydeep and Pouya, Soha and Chang, Yan},
  booktitle={ICRA},
  year={2025},
}

@inproceedings{3dmem,
  title={3d-mem: 3d scene memory for embodied exploration and reasoning},
  author={Yang, Yuncong and Yang, Han and Zhou, Jiachen and Chen, Peihao and Zhang, Hongxin and Du, Yilun and others},
  booktitle={CVPR},
  year={2025},
}

@misc{embodiedrag,
      title={Embodied-RAG: General Non-parametric Embodied Memory for Retrieval and Generation}, 
      author={Quanting Xie and So Yeon Min and Pengliang Ji and Yue Yang and Tianyi Zhang and Kedi Xu and others},
  journal={arXiv preprint arXiv:2409.18313},
      year={2025},
}

@misc{cut3r,
      title={Continuous 3D Perception Model with Persistent State}, 
      author={Qianqian Wang and Yifei Zhang and Aleksander Holynski and Alexei A. Efros and Angjoo Kanazawa},
  journal={arXiv preprint arXiv:2501.12387},
      year={2025},
}

@misc{bajcsy2018,
      title={Revisiting Active Perception}, 
      author={Ruzena Bajcsy and Yiannis Aloimonos and John K. Tsotsos},
  journal={arXiv preprint arXiv:1603.02729},
      year={2016},
}

@inproceedings{connolly1985, year={1985}, title={The determination of next best views}, booktitle={ICRA}, author={Connolly, C.},}

@inproceedings{isler2016, title={An information gain formulation for active volumetric 3D reconstruction}, booktitle={ICRA}, author={Isler, Stefan and Sabzevari, Reza and Delmerico, Jeffrey and Scaramuzza, Davide}, year={2016}, month=May,}

@misc{lookaround,
      title={Learning to Look Around: Intelligently Exploring Unseen Environments for Unknown Tasks}, 
      author={Dinesh Jayaraman and Kristen Grauman},
  journal={arXiv preprint arXiv:1709.00507},
      year={2017},
}

@misc{semexp,
      title={Object Goal Navigation using Goal-Oriented Semantic Exploration}, 
      author={Devendra Singh Chaplot and Dhiraj Gandhi and Abhinav Gupta and Ruslan Salakhutdinov},
  journal={arXiv preprint arXiv:2007.00643},
      year={2020},
}

@misc{vlfm,
      title={VLFM: Vision-Language Frontier Maps for Zero-Shot Semantic Navigation}, 
      author={Naoki Yokoyama and Sehoon Ha and Dhruv Batra and Jiuguang Wang and Bernadette Bucher},
  journal={arXiv preprint arXiv:2312.03275},
      year={2023},
}

@inproceedings{openeqa, title={OpenEQA: Embodied Question Answering in the Era of Foundation Models}, booktitle={CVPR}, author={Majumdar, Arjun and Ajay, Anurag and Zhang, Xiaohan and Putta, Pranav and Yenamandra, Sriram and Henaff, Mikael and others}, year={2024},}

@misc{abstaineqa,
      title={When Robots Should Say "I Don't Know": Benchmarking Abstention in Embodied Question Answering}, 
      author={Tao Wu and Chuhao Zhou and Guangyu Zhao and Haozhi Cao and Yewen Pu and Jianfei Yang},
  journal={arXiv preprint arXiv:2512.04597},
      year={2025},
}

@misc{avaloha,
      title={Active Vision Might Be All You Need: Exploring Active Vision in Bimanual Robotic Manipulation}, 
      author={Ian Chuang and Andrew Lee and Dechen Gao and M-Mahdi Naddaf-Sh and Iman Soltani},
  journal={arXiv preprint arXiv:2409.17435},
      year={2025},
}

@misc{msbp,
      title={Map Space Belief Prediction for Manipulation-Enhanced Mapping}, 
      author={Joao Marcos Correia Marques and Nils Dengler and Tobias Zaenker and Jesper Mucke and Shenlong Wang and Maren Bennewitz and others},
  journal={arXiv preprint arXiv:2502.20606},
      year={2025},
}

@misc{roboexp,
      title={RoboEXP: Action-Conditioned Scene Graph via Interactive Exploration for Robotic Manipulation}, 
      author={Hanxiao Jiang and Binghao Huang and Ruihai Wu and Zhuoran Li and Shubham Garg and Hooshang Nayyeri and others},
  journal={arXiv preprint arXiv:2402.15487},
      year={2024},
}

@misc{ingress,
      title={Interactive Visual Grounding of Referring Expressions for Human-Robot Interaction}, 
      author={Mohit Shridhar and David Hsu},
  journal={arXiv preprint arXiv:1806.03831},
      year={2018},
}

@misc{invigorate,
      title={INVIGORATE: Interactive Visual Grounding and Grasping in Clutter}, 
      author={Hanbo Zhang and Yunfan Lu and Cunjun Yu and David Hsu and Xuguang Lan and Nanning Zheng},
  journal={arXiv preprint arXiv:2108.11092},
      year={2024},
}

@misc{codeasmonitor,
      title={Code-as-Monitor: Constraint-aware Visual Programming for Reactive and Proactive Robotic Failure Detection}, 
      author={Enshen Zhou and Qi Su and Cheng Chi and Zhizheng Zhang and Zhongyuan Wang and Tiejun Huang and others},
  journal={arXiv preprint arXiv:2412.04455},
      year={2025},
}

@misc{sentinel,
      title={Unpacking Failure Modes of Generative Policies: Runtime Monitoring of Consistency and Progress}, 
      author={Christopher Agia and Rohan Sinha and Jingyun Yang and Zi-ang Cao and Rika Antonova and Marco Pavone and others},
  journal={arXiv preprint arXiv:2410.04640},
      year={2024},
}

@misc{safe,
      title={SAFE: Multitask Failure Detection for Vision-Language-Action Models}, 
      author={Qiao Gu and Yuanliang Ju and Shengxiang Sun and Igor Gilitschenski and Haruki Nishimura and Masha Itkina and others},
  journal={arXiv preprint arXiv:2506.09937},
      year={2025},
}

@misc{3dvla,
      title={3D-VLA: A 3D Vision-Language-Action Generative World Model}, 
      author={Haoyu Zhen and Xiaowen Qiu and Peihao Chen and Jincheng Yang and Xin Yan and Yilun Du and others},
  journal={arXiv preprint arXiv:2403.09631},
      year={2024},
}

@misc{spatialvla,
      title={SpatialVLA: Exploring Spatial Representations for Visual-Language-Action Model}, 
      author={Delin Qu and Haoming Song and Qizhi Chen and Yuanqi Yao and Xinyi Ye and Yan Ding and others},
  journal={arXiv preprint arXiv:2501.15830},
      year={2025},
}

@inproceedings{fuse,
  title={Beyond sight: Finetuning generalist robot policies with heterogeneous sensors via language grounding},
  author={Jones, Joshua and Mees, Oier and Sferrazza, Carmelo and Stachowicz, Kyle and Abbeel, Pieter and Levine, Sergey},
  booktitle={ICRA},
  year={2025},
}

@misc{geminirobotics,
      title={Gemini Robotics: Bringing AI into the Physical World}, 
      author={Gemini Robotics Team and Saminda Abeyruwan and Joshua Ainslie and Jean-Baptiste Alayrac and Montserrat Gonzalez Arenas and Travis Armstrong and others},
  journal={arXiv preprint arXiv:2503.20020},
      year={2025},
}

@misc{geminirobotics15,
      title={Gemini Robotics 1.5: Pushing the Frontier of Generalist Robots with Advanced Embodied Reasoning, Thinking, and Motion Transfer}, 
      author={Gemini Robotics Team and Abbas Abdolmaleki and Saminda Abeyruwan and Joshua Ainslie and Jean-Baptiste Alayrac and Montserrat Gonzalez Arenas and others},
  journal={arXiv preprint arXiv:2510.03342},
      year={2025},
}

@misc{blindtest,
      title={Benchmark Designers Should "Train on the Test Set" to Expose Exploitable Non-Visual Shortcuts}, 
      author={Ellis Brown and Jihan Yang and Shusheng Yang and Rob Fergus and Saining Xie},
  journal={arXiv preprint arXiv:2511.04655},
      year={2026},
}

@inproceedings{wu2015galileo,
  author = {Wu, Jiajun and Yildirim, Ilker and Lim, Joseph J. and Freeman, William T. and Tenenbaum, Joshua B.},
  title = {{Galileo: Perceiving Physical Object Properties by Integrating a Physics Engine with Deep Learning}},
  booktitle = {NeurIPS},
  year = {2015}
}

@inproceedings{baradel2020cophy,
  author = {Baradel, Fabien and Neverova, Natalia and Mille, Julien and Mori, Greg and Wolf, Christian},
  title = {{CoPhy: Counterfactual Learning of Physical Dynamics}},
  booktitle = {ICLR},
  year = {2020}
}

@article{montesano2008affordances,
  author = {Montesano, Luis and Lopes, Manuel and Bernardino, Alexandre and Santos-Victor, Jos{\'e}},
  title = {{Learning Object Affordances: From Sensory--Motor Coordination to Imitation}},
  journal = {IEEE Transactions on Robotics},
  year = {2008},
}

@article{mandlekar2019roboturk,
  title = {{Scaling Robot Supervision to Hundreds of Hours with RoboTurk: Robotic Manipulation Dataset through Human Reasoning and Dexterity}},
  author = {Mandlekar, Ajay and Booher, Jonathan and Spero, Max and Tung, Albert and Gupta, Anchit and Zhu, Yuke and others},
  journal = {arXiv preprint arXiv:1911.04052},
  year = {2019},
}

@article{walke2023bridgedatav2,
  title = {{BridgeData V2: A Dataset for Robot Learning at Scale}},
  author = {Homer Walke and Kevin Black and Abraham Lee and Moo Jin Kim and Max Du and Chongyi Zheng and others},
  journal = {arXiv preprint arXiv:2308.12952},
  year = {2023},
}

@article{fang2023rh20t,
  title = {{RH20T: A Comprehensive Robotic Dataset for Learning Diverse Skills in One-Shot}},
  author = {Fang, Hao-Shu and Fang, Hongjie and Tang, Zhenyu and Liu, Jirong and Wang, Chenxi and Wang, Junbo and others},
  journal = {arXiv preprint arXiv:2307.00595},
  year = {2023},
}

@article{wu2024robomind,
  title = {{RoboMIND: Benchmark on Multi-embodiment Intelligence Normative Data for Robot Manipulation}},
  author = {Kun Wu and Chengkai Hou and Jiaming Liu and Zhengping Che and Xiaozhu Ju and Zhuqin Yang and others},
  journal = {arXiv preprint arXiv:2412.13877},
  year = {2024},
}

@article{chen2024rh20tp,
  title = {{RH20T-P: A Primitive-Level Robotic Dataset Towards Composable Generalization Agents}},
  author = {Zeren Chen and Zhelun Shi and Xiaoya Lu and Lehan He and Sucheng Qian and Zhenfei Yin and others},
  journal = {arXiv preprint arXiv:2403.19622},
  year = {2024},
}

@article{zawalski2024ecot,
  title = {{Robotic Control via Embodied Chain-of-Thought Reasoning}},
  author = {Zawalski, Micha{\l} and Chen, William and Pertsch, Karl and Mees, Oier and Finn, Chelsea and Levine, Sergey},
  journal = {arXiv preprint arXiv:2407.08693},
  year = {2024},
}

@article{ji2025robobrain,
  title = {{RoboBrain: A Unified Brain Model for Robotic Manipulation from Abstract to Concrete}},
  author = {Ji, Yuheng and Tan, Huajie and Shi, Jiayu and Hao, Xiaoshuai and Zhang, Yuan and Zhang, Hengyuan and others},
  journal = {arXiv preprint arXiv:2502.21257},
  year = {2025},
}

@article{sermanet2023robovqa,
  title = {{RoboVQA: Multimodal Long-Horizon Reasoning for Robotics}},
  author = {Pierre Sermanet and Tianli Ding and Jeffrey Zhao and Fei Xia and Debidatta Dwibedi and Keerthana Gopalakrishnan and others},
  journal = {arXiv preprint arXiv:2311.00899},
  year = {2023},
}

@article{duan2024aha,
  title = {{AHA: A Vision-Language-Model for Detecting and Reasoning Over Failures in Robotic Manipulation}},
  author = {Duan, Jiafei and Pumacay, Wilbert and Kumar, Nishanth and Wang, Yi Ru and Tian, Shulin and Yuan, Wentao and others},
  journal = {arXiv preprint arXiv:2410.00371},
  year = {2024},
}

@article{liu2023causaltriplet,
  title = {{Causal Triplet: An Open Challenge for Intervention-centric Causal Representation Learning}},
  author = {Liu, Yuejiang and Alahi, Alexandre and Russell, Chris and Horn, Max and Zietlow, Dominik and Sch{\"{o}}lkopf, Bernhard and others},
  journal = {arXiv preprint arXiv:2301.05169},
  year = {2023},
}
